\documentclass{article}

\usepackage{microtype}
\usepackage{graphicx}
\usepackage{subfigure}
\usepackage{booktabs} %

\usepackage{hyperref}

\usepackage[accepted]{icml2025}

\usepackage{amsmath}
\usepackage{amssymb}
\usepackage{mathtools}
\usepackage{amsthm}

\usepackage[capitalize,noabbrev]{cleveref}

\theoremstyle{plain}
\newtheorem{theorem}{Theorem}[section]
\newtheorem{proposition}[theorem]{Proposition}

\theoremstyle{definition}

\newtheorem{assumption}[theorem]{Assumption}
\theoremstyle{remark}
\newtheorem{remark}[theorem]{Remark}

\usepackage[textsize=tiny]{todonotes}

\usepackage{cuted} %
\usepackage[makeroom]{cancel} %
\usepackage{enumitem} %
\usepackage{bm}
\usepackage[toc,page,header]{appendix} %
\usepackage{minitoc}

\usepackage{hyperref}
\usepackage{caption}
\usepackage{float}

\newcommand{\defg}{\overset{\Delta}{=}}
\newcommand{\diffusionpath}{uncoupled affine Gaussian path}

\newcommand{\coupling}{\pi}

\icmltitlerunning{On the Guidance of Flow Matching}

\begin{document}

\twocolumn[
\icmltitle{On the Guidance of Flow Matching}

\icmlsetsymbol{equal}{*}

\begin{icmlauthorlist}
\icmlauthor{Ruiqi Feng}{wu}
\icmlauthor{Chenglei Yu$^*$}{wu}
\icmlauthor{Wenhao Deng$^*$}{wu}
\icmlauthor{Peiyan Hu}{cas}
\icmlauthor{Tailin Wu}{wu}
\end{icmlauthorlist}

\icmlaffiliation{wu}{Department of Artificial Intelligence, Westlake University, Hangzhou, China}
\icmlaffiliation{cas}{Academy of Mathematics and Systems Science, Chinese Academy of Sciences, Beijing, China}

\icmlcorrespondingauthor{Tailin Wu}{wutailin@westlake.edu.cn}

\icmlkeywords{Machine Learning, ICML}

\vskip 0.3in
]

\printAffiliationsAndNotice{\icmlEqualContribution} %

\begin{abstract}

Flow matching has shown state-of-the-art performance in various generative tasks, ranging from image generation to decision-making, where generation under energy guidance (abbreviated as guidance in the following) is pivotal. However, the guidance of flow matching is more general than and thus substantially different from that of its predecessor, diffusion models. Therefore, the challenge in guidance for general flow matching remains largely underexplored. In this paper, we propose the first framework of general guidance for flow matching. From this framework, we derive a family of guidance techniques that can be applied to general flow matching. These include a new training-free asymptotically exact guidance, novel training losses for training-based guidance, and two classes of approximate guidance that cover classical gradient guidance methods as special cases. We theoretically investigate these different methods to give a practical guideline for choosing suitable methods in different scenarios. Experiments on synthetic datasets, image inverse problems, and offline reinforcement learning demonstrate the effectiveness of our proposed guidance methods and verify the correctness of our flow matching guidance framework. Code to reproduce the experiments can be found at \url{https://github.com/AI4Science-WestlakeU/flow_guidance}.

\end{abstract}

\section{Introduction}
\label{sec:introduction}

Flow matching has emerged as a prominent class of generative models. It features the ability to use a vector field to transform samples from a source distribution into samples following a target distribution, thus realizing generative modeling \citep{lipman_flow_2023}. The probability distribution the samples follow during the flow is called the probability path. By designing the probability path in a large design space, flow matching has shown improved generative modeling fidelity as well as higher sampling efficiency in a variety of generative modeling tasks including image generation \citep{lipman_flow_2023}, decision-making \citep{zheng_guided_2023}, audio generation, and molecular structure design \citep{gat_discrete_2024,chen_flow_2024,ben-hamu_d-flow_2024}. Flow matching substantially extends diffusion models \citep{ho_denoising_2020,song_score-based_2021}. Most diffusion models leverage the score matching process \citep{song_generative_2019,song_sliced_2020,song_score-based_2021}, inherently limiting them to using the Gaussian distribution as the source distribution to construct a special probability path. Meanwhile, flow matching can learn the mapping between any source distribution and target distributions \citep{lipman_flow_2023,lipman_flow_2024,chen_flow_2024,gat_discrete_2024}.

Guiding flow matching models refers to steering the generated samples toward desired properties, e.g., sampling from a distribution weighted with some objective function \citep{lu_contrastive_nodate} or conditioned on class labels \citep{song_score-based_2021}\footnote{Mathematically, they are essentially equivalent (Section \ref{sec:general_g}).}. It is vital in many generative modeling applications \citep{song_loss-guided_2023,zheng_guided_2023}, but contrary to well-studied guidance in diffusion models \citep{song_loss-guided_2023,chung_diffusion_2024,dhariwal_diffusion_2021,song_pseudoinverse-guided_2022,zheng_ensemble_2024,lu_contrastive_nodate,dou_diffusion_2023,trippe_diffusion_2023}, the guidance of flow matching remains less investigated. Most existing guidance methods only apply to a subset of flow matching that assumes the source distribution to be Gaussian and the probability path to have a certain simple form \citep{lipman_flow_2024,zheng_guided_2023,pokle_training-free_2024,anonymous2025energyweighted,anonymous2025flow}. In these cases, it is allowed to simplify the guidance of flow matching to be essentially the same as diffusion model guidance, but flow matching's power of generating more flexible probability paths than diffusion models \citep{tong_improving_2024,chen_flow_2024,gat_discrete_2024} is restricted. 
There have been other controlled generation methods for flow matching, with a notable stream following the paradigm of optimizing some objective functions via differentiating through the sampling process \citep{ben-hamu_d-flow_2024,liu_flowgrad_2023,anonymous2025training}. However, their goal differs from our guidance of weighting the generated distribution.
Therefore, the guidance for flow-matching models remains unrevealed in the rest of the ample design space. 

To fill this gap, in this work, we start from a similar assumption as diffusion guidance and propose a general framework of flow matching guidance. From the perspective of this framework, we propose \emph{Monte Carlo-based training-free asymptotically exact guidance} for flow matching. We also propose different training losses for exact \emph{training-based guidance}, one of which covers existing losses as special cases \citep{lu_contrastive_nodate,anonymous2025energyweighted}. For \emph{approximate guidance methods}, we can theoretically derive from our framework many famous guidance methods that have appeared in the literature, including DPS \cite{chung_diffusion_2024}, $\Pi$GDM \cite{song_pseudoinverse-guided_2022}, LGD \citep{song_loss-guided_2023}, as well as their flow-matching extensions that are theoretically justified for general flow matching models. We demonstrate the effectiveness of our proposed method in both synthetic datasets and decision-making (offline RL) benchmarks. Furthermore, more extensive experiments are conducted on image inverse problems to provide an empirical comparison of different types of guidance methods for a guideline of choosing different methods.

We summarize our contributions as follows:
\textbf{(1)} We propose a theoretically justified unified framework to construct guidance for general flow matching, \emph{i.e.,} with arbitrary source distribution, coupling, and conditional paths.
\textbf{(2)} The framework inspires us to propose a family of new guidance methods, including Monte Carlo sampling-based asymptotically exact guidance and training-based exact guidance for flow matching.
\textbf{(3)} The framework can exactly cover multiple classical guidance methods in flow matching and diffusion models. Contrary to previous derivations relying on the flow to have a Gaussian source distribution, our derivation provides theoretical justification of these methods for general flow matching.
\textbf{(4)} Empirical comparisons between guidance methods are conducted in different tasks, providing insights into choosing appropriate guidance methods for different generative modeling tasks.

\section{Background}
\label{sec:background}

Let $\mathbb{R}^d$ denote the data space where the data samples $x_t\in \mathbb{R}^d$. Here the subscript $t\in[0,1]$ denotes inference time such that $p_1(x_1)$ is a target distribution we want to generate, and $p_0(x_0)$ is a base distribution that is easy to sample from. 
Flow-based generative models \citep{lipman_flow_2023,lipman_flow_2024} define a vector field
$v_t(x_t): [0,1] \times \mathbb{R}^d \to \mathbb{R}^d$ that generates a 
probability path $p_t(x_t):[0,1]\times\mathbb{R}^d\to \mathbb{R}_{>0}$  \emph{connecting} the tractable base distribution $p_0(x_0)$ and the target distribution $p_1(x_1)$.
By first sampling $x_0$ from $p_0(x_0)$ and then solving the ordinary differential equation (ODE) 
$\frac{d}{dt}x_t = v_t(x_t)$, one can generate clean samples $x_1\coloneqq x_t|_{t=1}$ that follow the target distribution $p_1(x_1)$.

An efficient way to learn the vector field $v_t(\cdot)$ by a model $v_\theta(\cdot, t)$ is to use flow matching 
\citep{lipman_flow_2023,lipman_flow_2024,tong_improving_2024}.
It works by first finding a conditional vector field $v_{t|z}(x_t|z)$ that generates
a \textit{conditional} probability path $p_t(x_t|z)$, where $z$ denotes sample pairs $(x_0,x_1)$
\footnote{The notation $z$ can also represent $x_1$ alone \citep{lipman_flow_2024}. Our analysis is in the general setting where $z=(x_0,x_1)$, but for ease of interpretation, one may consider $z$ as simply $x_1$.}.
The pairs (couplings) follow the probability distribution of $p(z)=\coupling(x_0,x_1)$\footnote{We use $\pi$ to denote the probability density of data couplings.}. 
We use $(x_0,x_1)$ and $z$ interchangeably throughout the paper.

It has been proved that the \textit{marginal} vector field
\vspace{-5pt}
\begin{equation}\nonumber
\vspace{-5pt}
    v_t(x_t) \coloneqq \int v_{t|z}(x_t|z) p(z|x_t) dz,
\end{equation}
where $p(z|x_t) = \frac{p_t(x_t|z)p(z)}{p(x_t)}$,
will generate the \textit{marginal} probability path $p_t(x_t) = \int p_t(x_t|z) p(z) dz$ \citep{lipman_flow_2023}. 
Thus, one only needs to fit the \textit{marginal} vector field
$v_t(x_t) = \int v_{t|z}(x_t|z) p_t(z|x_t) dz$
using a model $v_\theta (x_t,t)$.

It is intuitive to construct the loss 
\begin{equation}\nonumber
    \mathbb{E}_{t\sim\mathcal{U}(0,1),x_t\sim p(x_t)}
    \big[\|v_\theta (x_t,t) - \underbrace{v_t(x_t)}_\text{intractable}\|^2_2\big],
    \vspace{-5pt}
\end{equation}
which is, unfortunately, intractable. 
To cope with this problem, an equivalent conditional flow matching loss has been proposed
\citep{lipman_flow_2023,tong_improving_2024}:
\vspace{-2.5pt}
\begin{equation}\nonumber
    \mathbb{E}_{t\sim\mathcal{U}(0,1),z\sim p(z),x_t \sim p(x_t|z)} 
    \left[ \left\| v_\theta (x_t,t) - v_{t|z}(x_t|z) \right\|^2_2 \right],
\end{equation}
which is tractable and can be used to train $v_\theta$.

\section{Guidance Vector Fields}
\label{sec:method}

\begin{figure*}[t]
    \centering    \includegraphics[width=1\linewidth]{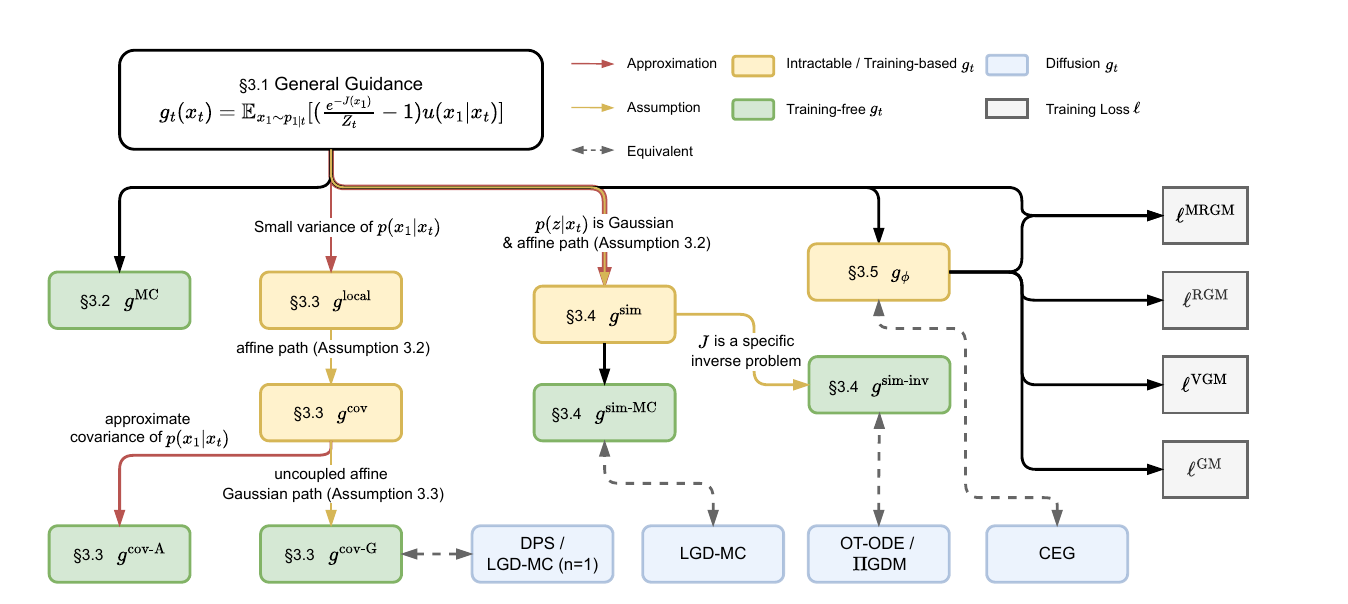}
    \caption{Overview of guidance methods in the paper. We start with a unified guidance expression and derive different guidance methods, including training-free and training-based methods, and cover many classical diffusion guidances.}
    \label{fig:fig1}
\end{figure*}

This section is organized as follows: Section \ref{sec:general_g} proposes the general flow matching guidance framework. In Section \ref{sec:g_mc}, we derive a new guidance method $g^{\text{MC}}$ based on Monte Carlo estimation, which is asymptotically exact. In Section \ref{sec:method_localized_posterior}, we derive a guidance $g_t^{\text{local}}$ proportional to the gradient of the energy function $J$ by approximating $g_t$ with Taylor expansion. Then, we introduce the affine path assumption (Assumption \ref{assumption:affine_path}) to obtain a tractable $g^{\text{cov}}\approx g_t^{\text{local}}$, and show that under the stronger \diffusionpath~assumption (Assumption \ref{assumption:uncoupled_affine_gaussian_path}),~$g^{\text{local}}$ covers classical diffusion guidance methods. Section \ref{sec:method_approximate_simple_posterior} introduces an alternative approximation that $p(z|x_t)$ is a Gaussian. Under this approximation and the affine path assumption (Assumption \ref{assumption:affine_path}), we derive a derivative-free guidance $g^{\text{sim-MC}}$, and a specialized guidance of $g^{\text{sim-A}}$ for inverse problems. Finally, Section \ref{sec:g_learned} provides different training losses for training-based exact guidance $g_\phi$. Figure \ref{fig:fig1} visualizes the above outline and the relations between these guidance methods.

\subsection{General Guidance Vector Fields}\label{sec:general_g}

Guided generative modeling aims to generate samples according to specific requirements, and energy-guided sampling is one of the primary approaches. Appendix \ref{app:posterior_sampling_and_energy_guided_sampling} shows that energy guidance is fundamentally equivalent to conditional sampling or posterior sampling. Accordingly, we use the notation of energy guidance throughout, with the theoretical results directly extending to these tasks. Given an energy function $J: \mathbb{R}^d\to \mathbb{R}$ and a pre-trained generative model for $p(x)$, energy-guided samples follow $x\sim p'(x) = \frac{1}{Z}p(x)e^{-J(x)}$, where $Z=\int p(x)e^{-J(x)}dx $ is the normalizing constant. Samples with lower energy values $J$ are more likely to be generated. Thus, the problem of guidance for flow-matching models becomes:

\textit{How can we alter the original vector field (VF) $v_t$ that generates $p(x)$ such that the new VF $v_t'$\footnotemark can generate samples from the new distribution $p'(x) = \frac{1}{Z}p(x)e^{-J(x)}$?}

\footnotetext{The prime symbol $'$ denotes probability distributions and vector fields corresponding to the \emph{new} distribution $p'(x)$.}

A natural choice that is commonly used in diffusion models \citep{dhariwal_diffusion_2021} is to add a guidance VF $g_t(x_t):[0,1] \times \mathbb{R}^d \to \mathbb{R}^d$
to the original VF $v_t(x_t)$, such that the new VF $v'_t(x_t) = g_t(x_t) + v_t(x_t)$ is a VF formed with the same flow matching hyperparameters but generates the new probability path arriving at the new distribution $p'(x)$. Therefore, 
\begin{equation}
    \nonumber g_t(x_t) = \int v'_{t|z}(x_t|z) p'(z|x_t) dz - \int v_{t|z}(x_t|z) p(z|x_t) dz,
\end{equation}
where $p(z|x_t) = \frac{p_t(x_t|z)p(z)}{p(x_t)}$ and $p'(z|x_t) = \frac{p'_t(x_t|z)p'(z)}{p'(x_t)}$.
Recall that new VF $v'_t(x_t)$ has the conditional probability path and conditional VF as that of the original VF, \emph{i.e.}, $v'_t(x_t|z)=v_t(x_t|z)$ and $p'_t(x_t|z) = p_t(x_t|z)$.
Then, we have the following theorem (proof in Appendix \ref{app:general_guidance}).
\begin{theorem}\label{theorem:general_guidance}
    Adding the guidance VF $g_t(x_t)$ to the original VF $v_t(x_t)$ will form VF $v'_t(x_t)$ that generates $p_t'(x_t) = \int p_t(x_t|z)p'(z)dz$, as long as $g_t(x_t)$ follows:
    \begin{align}\label{eq:general_guidance}
        &g_t(x_t) = \int  \left(\mathcal{P}\frac{e^{-J(x_1)}}{Z_t(x_t)} - 1\right) v_{t|z}(x_t|z) p(z|x_t) dz, \\
        \label{eq:general_guidance_z}
        &\text{where }Z_t(x_t) = \int \mathcal{P}e^{-J(x_1)} p(z|x_t)dz,
    \end{align}
    $\mathcal{P} = \frac{\coupling'(x_0|x_1)}{\coupling(x_0|x_1)}$ is the reverse coupling ratio, where $\coupling'(x_0|x_1)$ is the reverse data coupling for the new VF, i.e., the distribution of $x_0$ given $x_1$ sampled from the target distribution.
\end{theorem}

In this paper, we consider the case where $\mathcal{P}$ is or can be approximated as 1. $\mathcal{P}$ is exactly $1$ when the coupling is independent $\coupling(x_0,x_1)=p(x_0)p(x_1)$, which results in $\coupling'(x_0|x_1)=\coupling(x_0|x_1) = p(x_0)$. $\mathcal{P}=1$ is also reasonable in many practical flow matching methods with dependent couplings, such as mini-batch optimal transport (OT) conditional flow matching \citep{tong_improving_2024}, which we elaborate in Appendix \ref{app:p=1_approximation}. However, there may be cases where the VF of dependent couplings is significantly different from that of independent couplings. In these cases, the impact of $\mathcal{P}$ can be non-negligible, which we leave for future work.
Later, we will show this approximation allows us to cover many existing guidance techniques for flow matching or diffusion models.

Uncoupled affine Gaussian path flow matching, where $\coupling(x_0,x_1)=p(x_0)p(x_1)$, $v_{t|1}(x_t|x_1)=\alpha_t x_t + \beta_t x_1$, and $p(x_0)=\mathcal{N}(x_0;\mu,\Sigma)$, is known to be equivalent to diffusion models, with little difference in the noise schedule \citep{zheng_guided_2023,ma_sit_2024}. In this case, our general guidance for flow matching Eq. \eqref{eq:general_guidance} can be reduced to a commonly used guidance term in diffusion models:
$
    g_t(x_t)\propto \nabla_{x_t} \log Z_t(x_t)\nonumber
$
(proof in Appendix \ref{app:affine_gaussian_guidance_matching}).
Thus, most existing works that only consider \diffusionpath~flow matching essentially apply the same guidance techniques for diffusion models \citep{dhariwal_diffusion_2021,song_loss-guided_2023,chung_diffusion_2024}.

Next, we will explore more challenging scenarios of flow matching guidance, which are substantially different from diffusion guidance: either the coupling is dependent, or the source distribution is non-Gaussian. 

\subsection{Monte Carlo Estimation, $g_t^{\text{MC}}$}\label{sec:g_mc}

To start with, we discuss the Monte Carlo (MC) method to estimate the guidance $g_t(x_t)$ in Eq. \eqref{eq:general_guidance}, which is asymptotically exact while being training-free. The MC estimation of the integrals in Eq. \eqref{eq:general_guidance} and \eqref{eq:general_guidance_z} requires sampling from intractable $p(z|x_t)$, but they can be converted into:
\begin{align}\nonumber
&    \bm{g_t}^{\textbf{MC}}(x_t) \\\label{eq:g_MC}
& \defg \mathbb{E}_{x_1,x_0\sim p(z)}
\left[
(\frac{e^{-J(x_1)}}{Z_t} - 1) v_{t|z}(x_t|z) \frac{p_t(x_t|z)}{p_t(x_t)}
\right],
\\
&\nonumber    Z_t(x_t) = \mathbb{E}_{x_1,x_0\sim p(z)} 
\left[
e^{-J(x_1)} \frac{p_t(x_t|z)}{p_t(x_t)} 
\right],
\end{align}
where only $p_t(x_t)$ remains intractable. We can use the same MC samples to self-normalize the distribution $p_t(x_t|z)$, using $
    p_t(x_t) = \mathbb{E}_{x_1,x_0\sim p(z)}[p_t(x_t|z)]
$.
In the expressions above, $p_t(x_t|z)$ is usually designed to be a simple, known distribution \citep{tong_improving_2024} and $p(z)$ can be sampled either using the learned generative model or from the training distribution if accessible.
The above method can be understood as executing importance sampling to convert the expectation under the intractable distribution $\mathbb{E}_{z\sim p(z|x_t)}[\cdot]$ to that under a tractable distribution $\mathbb{E}_{z\sim p(z)}[\frac{p(x_t|z)}{p(x_t)}(\cdot)]$.
The pseudocode for computing $g^{\text{MC}}_t(x_t)$ can be found in Algorithm \ref{alg:mc_estimation_on_g}, and
a simplified version of $g^{\text{MC}}$ under the assumption of independent coupling is provided in Appendix \ref{app:independent_mc}.

It should be noted that this method is unbiased and applicable to \textit{any} source distribution.
This enables the guidance for flow matching with different source distributions such as uniform \citep{chen_flow_2024}, Gaussian process \citep{andrae2025continuous}, mixture of Gaussian \citep{hiranaka2025hero,papamakarios_masked_2018}, and others \citep{mathieu_riemannian_2020,stimper_resampling_2022}. This method can also be applied to flow matching with dependent couplings of $x_1,x_0$, \textit{e.g.}, optimal transport couplings \citep{tong_improving_2024} and rectified flow \citep{liu_flow_2022}. 

To our knowledge, $g^{\text{MC}}_t$ is the \emph{first} to provide an asymptotically exact training-free estimation of the guidance VF for flow matching whose source distribution is non-Gaussian. However, due to the high variance of MC given a limited number of samples, $g^{\text{MC}}$ is more suitable for tasks where the energy function $J$ varies gently and the data is relatively low-dimensional.

Since the efficiency of $g^{\text{MC}}$ is restricted by the high variance in the MC estimation, it has an unsatisfactory performance in high-dimensional generation tasks like image inverse problems. 
To improve the scalability of $g^{\text{MC}}$, there are many techniques that can be readily applied. For example, we can adopt importance sampling in Eq. \eqref{eq:g_MC} to reduce the variance of the MC estimation:
\begin{align}\nonumber
&\bm{g_t}^{\textbf{MC-IS}}(x_t) \\ 
\nonumber
& \defg \mathbb{E}_{x_1,x_0\sim \tilde{p}(z)} 
\left[\frac{{p}(z)}{\tilde{p}(z)}
(\frac{e^{-J(x_1)}}{Z_t} - 1)  \frac{p_t(x_t|z)}{p_t(x_t)}
v_t (x_t|z) 
\right],    
\end{align}
\begin{align}
\nonumber & Z_t(x_t) 
= \mathbb{E}_{x_1,x_0\sim \tilde{p}(z)} 
\left[\frac{{p}(z)}{\tilde{p}(z)}
e^{-J(x_1)} \frac{p_t(x_t|z)}{p_t(x_t)} 
\right],
\end{align}
where $\tilde p(z)$ is an alternative distribution that can be sampled from.
To enhance the scalability of $g^{\text{MC-IS}}$, we need to select a $\tilde{p}$ such that $\frac{e^{-J(x_1)}}{\tilde{p}(z)}$ has lower variance, \textit{i.e.}, when $e^{-J(x_1)}$ is large, $\tilde{p}(z)$ is also large, and vice versa. This can be achieved by sampling using another guided VF, such as those to be proposed subsequently, which generates samples close to (but not exactly) $\frac{1}{Z}p(x_1)e^{-J(x_1)}$. The probability density ratio $\frac{p(z)}{\tilde p(z)}$ can be estimated using, for example, the Hutchinson trace estimator to preserve scalability \citep{song_score-based_2021}. It should be noted that this estimation is still unbiased.\footnote{We leave empirical investigation into $g^{\text{MC-IS}}$ to future work.}

\subsection{Localized $p(x_1|x)$}\label{sec:method_localized_posterior}

Many practical guidance methods rely on approximations \citep{song_loss-guided_2023,pokle_training-free_2024}, so contrary to the unbiased MC estimation of $g_t$, we investigate approximate (and thus biased) guidance methods in this subsection.
We start from an intuitive assumption that the probability mass of $p(x_1|x_t)$ is centered around its mean. Following this, it is natural to approximate the integrals in Eq.~\eqref{eq:general_guidance} with the Taylor expansion that captures local behaviors around the mean. Thus, Eq. \eqref{eq:general_guidance} can be simplified and becomes tractable. 

First, we approximate the normalizing constant $Z_t(x_t)$ in Eq. \eqref{eq:general_guidance_z} as (Appendix \ref{app:localized_approximation}):
\begin{align}
Z_t(x_t) = \int p(z|x_t) e^{-J(x_1)} dz  
\approx e^{-J(\hat{x}_1)}\label{eq:approx_1order_z_uncoupled_approximation}
\end{align}
where $\hat{x}_1 \coloneqq \mathbb{E}_{x_0,x_1\sim p(z|x_t)}[x_1]$. 
Likewise, Eq. \eqref{eq:general_guidance} can be approximated as $g_t(x_t) \approx g_t^{\text{local}}$, and:
\begin{align}
    &\bm{g_t}^{\textbf{local}}(x_t)\nonumber
\label{eq:local_approx_guidance_initial}    \defg -\mathbb{E}_{z \sim p(z|x_t)}
    \left[
     (x_1 - \hat{x}_1)v_{1|t}(x_t|z)
    \right] \nabla_{\hat{x}_1} J(\hat{x}_1),
\end{align}
where $\hat{x}_1$ is defined as above. The approximation error $\|\delta g\|_2^2$ can be bounded: $\|\delta g\|^2_2 \le |\lambda_h \sigma_1 d / e^{-J(\hat{x}_1)}|^2(C_1+C_2)$ (Appendix \ref{app:localized_approximation}), where $\lambda_h$ is the maximum eigenvalue of the Hessian of $e^{-J(x)}$, $\sigma_1$ is the $L_2$ norm of the covariance matrix of the distribution $p(x_1|x_t)$, $d$ is the dimensionality of the data, and $C_1,C_2$ are constants dependent on the variance of the original conditional VF and the guided VF. 

The error bound gives insights on the approximation quality of $g^{\text{local}}_t$ (detailed discussion in Appendix \ref{app:localized_approximation}.):
\begin{itemize}
    \item The error is small when $J$ is smooth, in which case the Hessian of $e^{-J(x)}$ will approach zero. This corresponds to the mild guidance, where approximation-based $g^{\text{local}}$ works well.
    \item The error is small when $\sigma_1$ is small, \emph{i.e.} the covariance matrix $\Sigma_{11}$ has a small Frobenius norm. This is the case when the flow time $t\rightarrow 1$ (and $\sigma_t = 0$), where $x_t$ predicts $x_1$ well.
\end{itemize}

The gradient in $g_t^{\text{local}}$ is a natural outcome of the Taylor expansion near the mean of $p(x_1|x_t)$. $g_t^{\text{local}}$ is not only applicable to more general flow matching but also originates differently from diffusion guidance \citep{dhariwal_diffusion_2021} where the gradient naturally emerges from the score function $\nabla_{x_t}\log p(x_t)$. Moreover, the error bound we provide here is more practical compared to those previously proposed for diffusion guidance, \textit{e.g.}, the Jensen gap in diffusion posterior sampling (DPS) \citep{chung_diffusion_2024}, which only bounds the error in $\mathbb{E}[e^{-J(x_1)}]$, but it is $\nabla_{x_t}\log Z_t$ that is the guidance VF, whose error is not bounded. 

In order to obtain $\hat{x}_1$, we need the following assumption:
\begin{assumption}\label{assumption:affine_path}
    \textbf{The affine path assumption.} We assume the conditional probability path to be affine, \textit{i.e.}, $x_t = \alpha_t x_1 + \beta_t x_0 + \sigma_t\varepsilon$, where $\varepsilon$ is a random noise and $\sigma_t$, $\dot\sigma_t$ are both sufficiently small.\footnotemark
\end{assumption}
\footnotetext{This choice is a widely used one \cite{lipman_flow_2023,tong_improving_2024}. With a small random noise $\sigma_t \varepsilon$, $x_t$ under conditional VF flows is almost exactly from $x_0$ to $x_1$. }

Note that this assumption does not prevent the samples from having dependent coupling: $\coupling(x_0|x_1) \neq p(x_0)$.
Under Assumption \ref{assumption:affine_path}, we can use the $x_1$-parameterization \citep{lipman_flow_2024} to express $\hat{x}_1$ with the VF $v_t$ that is learned by the model $v_\theta$ (Appendix \ref{appendix:x1_parameterization}):
\vspace{-5pt}
\begin{equation}\label{eq:local_approx_guidance_affine_path_estimate_x1}
    \hat{x}_1 \approx -\frac{\dot\beta_t}{\dot \alpha_t \beta_t - \dot \beta_t \alpha_t} x_t + \frac{\beta_t}{\dot \alpha_t \beta_t - \dot \beta_t \alpha_t} v_t.
\end{equation}
With the commonly chosen schedule $\alpha_t = t, \beta_t = 1 - t$ \citep{lipman_flow_2023,tong_improving_2024}, $\hat{x}_1 = x_t + (1 - t) v_t$ coincides with the 1-step generated $x_1$ under the Euler scheme.
Also under this affine path assumption, $g^{\text{local}}_t$ can be expressed with the covariance matrix of $p(x_1|x_t)$, $\Sigma_{1|t}$, and the gradient $\nabla_{\hat{x}_1}J(\hat{x}_1)$ (proof in Appendix \ref{appendix:affine_path_cov_local_approx_guidance}): 
\begin{equation}\label{eq:g_cov}
    \bm{g_t^{\textbf{cov}}} \defg -\underbrace{\frac{\dot\alpha_t\beta_t - \dot\beta_t\alpha_t}{\beta_t}}_{\text{schedule}} \Sigma_{1|t} \nabla_{\hat{x}_1}J(\hat{x}_1).
    \vspace{-5pt}
\end{equation}
Intuitively, the guidance is the gradient of estimated $J$ preconditioned with the covariance matrix of $p(z|x_t)$. The preconditioning squeezes the guidance vector into the $p(z|x_t)$ manifold. Next, we discuss different ways to obtain this covariance term, resulting in $g^{\text{cov-A}}$, $g^{\text{cov-L}}$, and $g^{\text{cov-G}}$.

The simplest way is to approximate the covariance matrix with a manually set schedule $\lambda'_t I$, resulting in $g_t^{\text{cov-A}}$. This allows us to tune the guidance's schedule, a common practice in diffusion model guidance \citep{song_loss-guided_2023,song_pseudoinverse-guided_2022}.
\begin{equation}\nonumber
    \bm{g_t}^{\textbf{cov-A}} \defg -\lambda^{\text{cov-A}}_t \nabla_{\hat{x}_1}J(\hat{x}_1),
\end{equation}
where the original schedule in $g^{\text{cov}}_t$ is already included in the hyperparameter $\lambda^{\text{cov-A}}_t$. Since $p(x_1|x_t)$ is localized when $t\rightarrow 1$, a general guideline is to set $\lambda^{\text{cov-A}}_t$ decaying.
$g_t^{\text{cov-A}}$ only assumes affine path (Assumption \ref{assumption:affine_path}), and is thus theoretically justified for mini-batch optimal transport conditional flow matching \citep{tong_improving_2024} for which few existing guidance methods have been proposed.
Besides, $g_t^{\text{cov-A}}$ is efficient as its computation involves no extra number of function evaluations (NFE) than unguided sampling. 

Alternatively, we can use Proposition \ref{proposition:matching_anything} to train a model to fit the actual covariance matrix of $p(x_1|x_t)$ and acquire $g_t^{\text{cov-L}}$. Note that this covariance matrix is determined by the original distribution and is agnostic to the energy function $J$. The original flow matching essentially learns $\mathbb{E}_{x_1\sim p(x_1|x_t)}[x_1]$, but to achieve better approximate guided generation, more detailed information of the distribution $p(x_1|x_t)$, its covariance, also needs to be learned.

The important special case where flow matching is a diffusion model with a new schedule is formalized with:
\begin{assumption}
    \label{assumption:uncoupled_affine_gaussian_path}
    \textbf{The \diffusionpath~assumption}.
    In addition to Assumption \ref{assumption:affine_path}, the source distribution $p(x_0)$ is a standard Gaussian and the coupling is independent $\coupling(x_0,x_1) = p(x_0)p(x_1)$.
\end{assumption}
We will demonstrate that under Assumption \ref{assumption:uncoupled_affine_gaussian_path}, $g_t^{\text{cov}}$ simplifies into $g_t^{\text{cov-G}}$ (Eq. (\ref{eq:g_cov_G})), which covers classical guidance methods in diffusion models.
Specifically, using the second-order Tweedie's formula, we can express the covariance matrix $\Sigma_{1|t}$ in Eq. \eqref{eq:g_cov} in terms of the Hessian of the log-probability, $\nabla_{x_t} \nabla_{x_t} \log p(x_t)$ \citep{rozet_learning_2024,boys_tweedie_2024,ye_tfg_2024}. Then, under Assumption \ref{assumption:uncoupled_affine_gaussian_path}, $\nabla_{x_t} \log p(x_t)$ depends affinely on the VF $v_t$ in Eq. \eqref{eq:local_approx_guidance_affine_path_estimate_x1} which depends affinely on $\hat x_1$. Therefore, the covariance matrix $\Sigma_{1|t}$ can be expressed with the derivative of the Jacobian matrix $\frac{\partial \hat x_1}{\partial x_t}$.
We refer to this relationship as the Jacobian trick (proof in Appendix \ref{appendix:jacobian_trick}):
\begin{proposition} \textbf{The Jacobian trick}. Under Assumption \ref{assumption:uncoupled_affine_gaussian_path}, the inverse covariance matrix of $p(x_1|x_t)$, $\Sigma_{1|t}$, depends affinely on the Jacobian of the VF $\frac{\partial v_t}{\partial x_t}$, and is proportional to the Jacobian $\frac{\partial \hat{x}_1}{\partial x_t}$:
\vspace{-5pt}
\begin{equation}\nonumber
     \Sigma_{1|t}= \frac{\beta_t^2}{\alpha_t(\dot\alpha_t\beta_t - \dot\beta_t\alpha_t)} (-\dot\beta_t+ \beta_t\frac{\partial v_t}{\partial x_t} )
     = \frac{\beta_t^2}{\alpha_t} \frac{\partial \hat{x}_1}{\partial x_t}.
     \vspace{-10pt}
\end{equation}\label{proposition:jacobian_trick}
\end{proposition}
Inserting back to Eq. \eqref{eq:g_cov}, we have:
\begin{align}\label{eq:g_cov_G}
    \bm{g_t}^{\textbf{cov-G}}\defg&\lambda^{\text{cov-G}}_t\nabla_{x_t}J(\hat{x}_1),
\end{align}
where $\lambda^{\text{cov-G}}_t = -{\beta_t(\dot\alpha_t\beta_t - \dot\beta_t\alpha_t)}/{\alpha_t}$ is the schedule.

$g_t^{\text{cov-G}}$ covers classical diffusion model guidance including loss-guided diffusion\footnote{We refer to the simplest version of LGD-MC($n=1$) as LGD here. LGD-MC with $n>1$ will be covered in Section \ref{sec:method_approximate_simple_posterior}.} (LGD) \citep{song_loss-guided_2023} and diffusion posterior sampling (DPS) \citep{chung_diffusion_2024}. In diffusion models, the guidance can be expressed as $\nabla_{x_t} \log \mathbb{E}_{z\sim p(z|x_t)}[e^{-J(x_1)}]$, and DPS approximates this by neglecting the gap in the Jensen inequality \citep{chung_diffusion_2024} to move the expectation into $e^{J(\cdot)}$, and LGD approximates the expectation with a point estimation. Both methods arrive at $-\nabla_{x_t}J(\hat{x}_1)$, which is covered by our Eq. \eqref{eq:g_cov_G}.

\begin{figure*}[htb]
    \centering
    \includegraphics[width=1\linewidth]{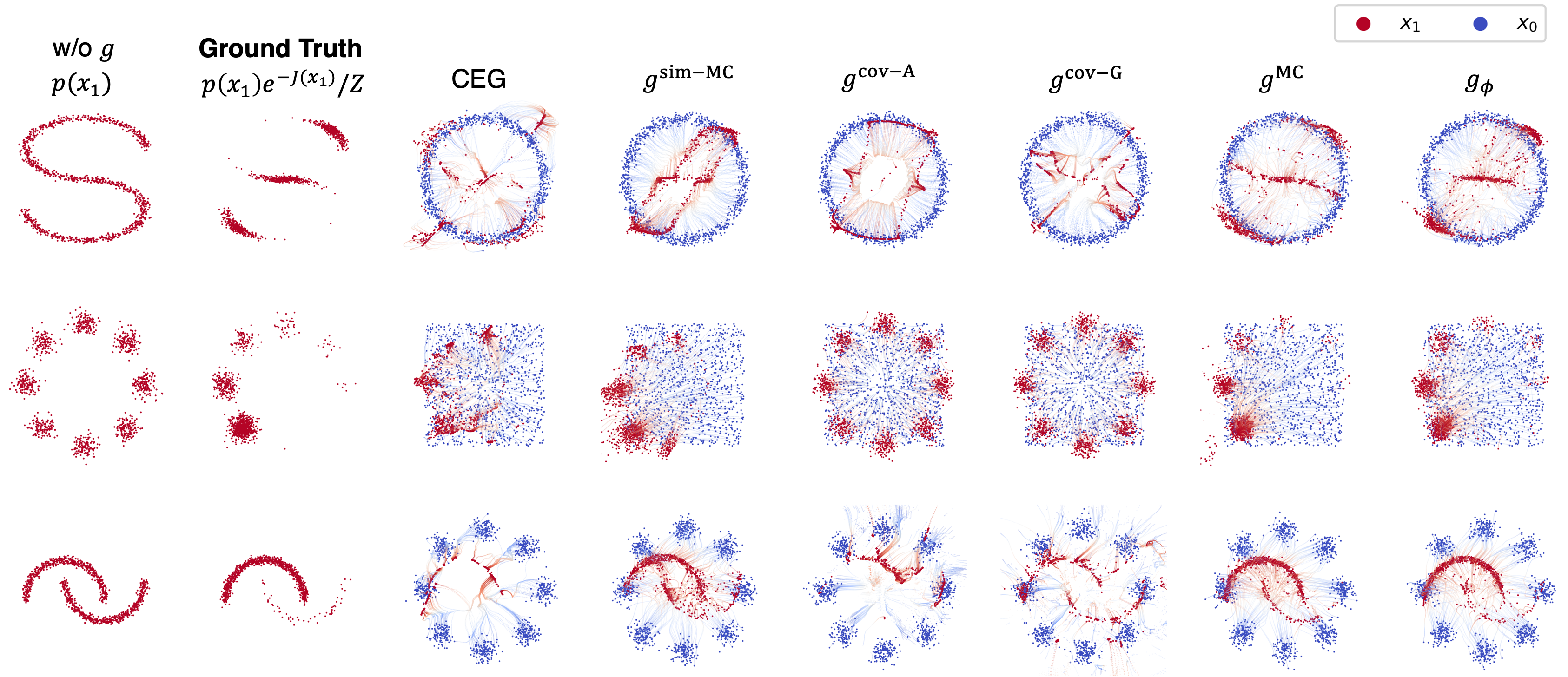}
    \caption{Results of the synthetic dataset with different source (blue) and target (red) distributions. We visualize the start/end points and the flow trajectories. $g^{\text{MC}}$ and $g_\phi$ yield the best guidance across different settings while diffusion guidance fails.}
    \label{fig:toy}
\end{figure*}

\subsection{Gaussian Approximation of $p(z|x_t)$}\label{sec:method_approximate_simple_posterior}

Instead of approximating guidance by expanding it near its mean, we can alternatively approximate
 $p(z|x_t)$ with a Gaussian distribution $\tilde p(z|x_t)=\mathcal{N}(z;\mu_t(x_t),\Sigma_t(x_t))$, and get $g_t\approx g_t^{\text{sim}}$.
To minimize the KL divergence between the approximate distribution and the original distribution, we need to choose $\mu_t(x_t)$ and $\Sigma_t(x_t)$ to that of the actual distribution $p(z|x_t)$ \citep{rozet_learning_2024,boys_tweedie_2024}. Under Assumption \ref{assumption:affine_path}, $\hat{x}_1$ can be estimated using Eq. \eqref{eq:local_approx_guidance_affine_path_estimate_x1}, $\hat{x}_0$ can be similarly estimated (Appendix \ref{appendix:x1_parameterization}), and $\Sigma_t(x_t)$ can be either set as a hyperparameter as in $g_t^{\text{cov-A}}$ or computed as in $g_t^{\text{cov-G}}$ in \diffusionpath~flow matching.

Now that if $J$ has an unknown expression, we can use Monte Carlo (MC) sampling to estimate $g_t^{\text{sim}}\approx g_t^{\text{sim-MC}}$, since we can sample from $\tilde{p}(z|x_t)$ which is a Gaussian distribution:
\begin{equation}
    \bm{g_t}^{\textbf{sim-MC}} \defg \sum_i^N \left(\frac{e^{-J(x_1^i)}}{\tilde{Z}_t} - 1\right) v_{t|z}(x_t|z^i),
\end{equation}
where $\tilde{Z}_t\coloneqq \frac{1}{N} \sum_i^N e^{-J(x_1^i)} $ is an estimated normalizing constant. This shares the spirit of LGD-MC \citep{song_loss-guided_2023}, which uses MC to estimate $Z_t$ and then computes diffusion guidance $\nabla_{x_t}\log Z_t$.

In some guided generation tasks like inverse problems where $J$ is known analytically, we can derive analytical expressions of $g_t^{\text{sim}}$. For example, when the measurement involves a known degradation operator $H$ applied to $x$ and then adding Gaussian noise with scale $\sigma_y$ to $Hx$, we have $J(x_1) \propto \frac{1}{2\sigma_y^2}\|y - Hx\|_2^2$. Thus we can derive $g_t^{\text{sim}}$ under the affine path assumption (Assumption \ref{assumption:affine_path}), and propose a practical approximation $g_t^{\text{sim-inv-A}}$ (Appendix \ref{appendix:approximate_simple_posterior_pigdm_like}):
\vspace{-5pt}
\begin{equation}
\vspace{-5pt}
    \bm{g_t}^{\textbf{sim-inv-A}} \defg -\lambda_t \Big{(}\frac{\sigma_y^2}{r_t^2} + H^TH\Big{)}^{-1}
    H^T\left(y - {H} \hat{x}_1\right),
\end{equation}
where $\lambda_t$ and $r_t$ are both hyperparameters. $g_t^{\text{sim-inv-A}}$ extends $\Pi$GDM \cite{song_pseudoinverse-guided_2022} to affine~flow matching but requires a further approximation $\frac{\partial \hat{x}_1}{\partial x_t}\approx I$, which is accurate when $t\rightarrow 1$.

In \diffusionpath~(Assumption \ref{assumption:uncoupled_affine_gaussian_path}), $g_t^{\text{sim}}$ in our framework can exactly cover $\Pi$GDM and OT-ODE \citep{song_pseudoinverse-guided_2022,pokle_training-free_2024}
(Appendix \ref{appendix:approximate_simple_posterior_pigdm_like}). 
Note that our $g_t^{\text{sim}}$ is theoretically justified for dependent couplings, such as optimal transport conditional flow matching (OT-CFM) \citep{tong_improving_2024}.

\subsection{Training-based Guidance $g_\phi$}
\label{sec:g_learned}

Previously, we have discussed training-free guidance methods. In this subsection, we discuss how to train a neural network $g_\phi$ to fit guidance $g_t$.
To construct a tractable training loss, we extend the conditional loss in flow matching to arbitrary conditional variables in the following proposition (proof in Appendix \ref{app:proposition_match_anything}).
\begin{proposition}\label{proposition:matching_anything}
    Any marginal variable $f(x_t,t)\coloneqq \mathbb{E}_{z\sim p_t(z|x_t)}[f_{t|z}(x_t,z,t)],~z=(x_0,x_1)$ has an intractable marginal loss:
    \vspace{-5pt}
    \begin{equation}\nonumber
    \vspace{-5pt}
    \mathcal{L}_t=\mathbb{E}_{x_t\sim p(x_t)}\left[\left\|f_{{\theta}}(x_t,t) - \mathbb{E}_{z\sim p_t(z|x_t)}[f_{t|z}(x_t,z,t)]\right\|_2^2\right],
    \end{equation}
    whose gradient w.r.t. $\theta$ is identical to that of the tractable conditional loss:
    \vspace{-5pt}
    \begin{equation}\nonumber
    \mathcal{L}_{t|z}=\mathbb{E}_{x_t,z\sim p(x_t,z)}\left[\left\|f_{\theta}(x_t,t) - f_{t|z}(x_t,z,t)\right\|_2^2\right].
    \vspace{-5pt}
    \end{equation}
\end{proposition}
\paragraph{Guidance Matching.} Based on proposition \ref{proposition:matching_anything}, We can train $g_\phi$ by first learning a surrogate model $Z_{\phi_Z}(x_t,t) \approx Z_t $, and then train $g_\phi\approx g_t$ with the following guidance matching (GM) losses.
\begin{align}\label{eq:guidance_matching_loss_sum}
    \mathcal{L}_{\phi_Z,\phi} &= 
    \mathbb{E}_{t\sim\mathcal{U}(0,1),z\sim p(z),x_t \sim p(x_t|z)} [\ell_{\phi_Z} + \ell_{\phi}^{\text{GM}}],\\
    \nonumber
    \ell_{\phi}^{\text{GM}} &= 
    \left\|g_{\phi}(x_t,t) - \left(\frac{e^{-J(x_1)}}{Z_{\phi_Z,sg}(x_t,t)} - 1\right) v_{t|z}(x_t|z)\right\|_2^2,
    \\
    \label{eq:guidance_matching_loss_g_1}
    \ell_{\phi_Z} &= 
    \left\|Z_{\phi_Z}(x_t,t) - e^{-J(x_1)}\right\|_2^2,
\end{align}
where $sg$ denotes stopping the gradient in automatic differentiation.
We prove in Appendix \ref{app:guidance_matching} that the minimizer of $\mathcal{L}_{\phi_Z,\phi}$ is indeed $Z_{\phi_Z} = Z_t$ and $g_{\phi} = g_t$.

In fact, there are other methods to learn $Z_{\phi_Z}$. In the literature on diffusion models, \citet{lu_contrastive_nodate} proposed to use contrastive learning to obtain $Z_t(x_t)$ that can be extended to include the general flow matching path (Appendix \ref{appendix:other_ways_to_learn_z}). Alternatively, Monte Carlo estimation can be applied to obtain $Z_{\phi_Z}$ (Appendix \ref{appendix:other_ways_to_learn_z}).
As for learning $g_\phi$,
alternative to the loss $\ell_{\phi}^{\text{GM}}$ in Eq. \eqref{eq:guidance_matching_loss_g_1}, 
there are other losses $\ell^{\text{VGM}}$, $\ell^{\text{RGM}}$, and $\ell^{\text{MRGM}}$ which, when substituted into Eq. \eqref{eq:guidance_matching_loss_sum}, will produce the same minimizer. We provide detailed analysis and proof for all these losses in Appendix \ref{app:guidance_matching}.
Notably, $\ell^{\text{MRGM}}$ that we derive is identical to a newly proposed training-based flow matching guidance loss in \citet{anonymous2025energyweighted}. For \diffusionpath~flow matching (Assumption \ref{assumption:uncoupled_affine_gaussian_path}), since the guidance is $g_t\propto\nabla_{x_1}\log Z_t(x_t)$ (Appendix \ref{app:affine_gaussian_guidance_matching}), learning $Z_{\phi_Z}$ in Eq. \eqref{eq:guidance_matching_loss_g_1} is adequate, and \citet{lu_contrastive_nodate} also learns $Z_{\phi_Z}$.

These different training-based guidance opens the design space of classifier-guidance \citep{dhariwal_diffusion_2021} for flow matching. For diffusion models, one only needs to train a classifier on noisy input to produce the correct guidance, whereas for general flow matching, one needs the training loss proposed above to train a network to get the accurate guidance.

In summary, we have proposed guidance matching methods for training-based guidance in this subsection. We derive different losses for learning $Z_t$ and $g_t$, which all provide unbiased estimations of the gradient and can be utilized without specific assumptions on flow matching.

\section{Experiments}
\label{sec:experiments}

In the experiment, we benchmark different guidance methods for flow matching models in different tasks including synthetic datasets, generative decision-making, and image inverse problems. These tasks all fall into the category of energy guidance \citep{janner_planning_2022,lu_contrastive_nodate} or posterior sampling \cite{chung_diffusion_2024}. 
With these experiments, we aim to answer the following questions:
    \textbf{(1)} Can the proposed learning-based exact guidance method $g_\phi$ learn the correct guidance VF $g_t$ for general (including non-\diffusionpath) flow matching?
    \textbf{(2)} For the asymptotically exact MC guidance method, can it produce the correct guidance in non-Gaussian flow matching, and is it exact assuming a sufficiently large sample budget?
    \textbf{(3)} On the practical aspect, how do different types of guidance methods (approximate/exact, training-free/training-based) perform on more realistic tasks, and how do we choose the appropriate flow matching guidance in different tasks?

\begin{table*}[h!]
\caption{Results of the D4RL Locomotion experiments. Entries with $\ge 95\%$ score of the best results per task (excluding baselines) are highlighted in bold. The standard deviation is deferred to Appendix \ref{app:exp_rl} due to limited space.}
\resizebox{\textwidth}{!}{
\centering
\begin{tabular}{cccccccccccccccc}
\toprule
 & & \multicolumn{2}{c}{Baselines} & \multicolumn{6}{c}{OT-CFM} & \multicolumn{6}{c}{CFM} \\
\cmidrule(lr){3-4} \cmidrule(lr){5-10} \cmidrule(lr){11-16}
& & BC & Diffuser & w/o $g$ & $g^{\text{cov-A}}$ & $g^{\text{cov-G}}$ & $g^{\text{sim-MC}}$ & $g^{\text{MC}}$ & $g_\phi$ & w/o $g$ & $g^{\text{cov-A}}$ & $g^{\text{cov-G}}$ & $g^{\text{sim-MC}}$ & $g^{\text{MC}}$ & $g_\phi$ \\
\midrule
              & HalfCheetah & 55.2 & 88.9 & 61.9 & 64.8 & 73.2 & 78.1 & \textbf{86.4} & 70.2 & 46.4 & 63.4 & 68.5 & \textbf{83.5} & \textbf{87.7} & 81.5 \\
Medium-Expert & Hopper & 52.5 & 103.3 & 95.2 & 101.8 & \textbf{112.3} & \textbf{112.3} & \textbf{112.7} & 98.1 & 83.4 & 93.9 & \textbf{113.3} & 88.5 & \textbf{113.3} & 91.5 \\
              & Walker2d & 107.5 & 106.9 & 79.1 & 97.3 & \textbf{107.2} & 101.0 & \textbf{107.2} & 91.3 & 65.7 & 100.4 & \textbf{106.9} & \textbf{107.0} & \textbf{107.1} & \textbf{102.3} \\
\midrule
       & HalfCheetah & 42.6 & 42.8 & 34.7 & \textbf{42.2} & \textbf{42.9} & \textbf{43.1} & \textbf{43.1} & \textbf{43.4} & \textbf{41.8} & \textbf{43.6} & \textbf{43.3} & \textbf{43.8} & \textbf{43.8} & \textbf{43.8} \\
Medium & Hopper & 52.9 & 74.3 & 63.3 & 75.1 & \textbf{89.8} & 76.2 & 79.8 & 79.7 & 73.2 & 79.1 & 82.7 & 82.1 & \textbf{88.0} & 85.2 \\
       & Walker2d & 75.3 & 79.6 & 72.4 & \textbf{82.7} & \textbf{81.3} & \textbf{83.4} & \textbf{83.0} & \textbf{80.6} & 72.2 & \textbf{80.7} & \textbf{82.5} & \textbf{81.9} & \textbf{81.9} & 72.9 \\
\midrule
              & HalfCheetah & 36.6 & 37.7 & 25.6 & 31.7 & 36.1 & 36.8 & \textbf{40.0} & 35.5 & 22.2 & 33.4 & \textbf{39.3} & 37.9 & \textbf{40.6} & \textbf{39.1} \\
Medium-Replay & Hopper & 18.1 & 93.6 & 40.1 & 57.7 & 74.1 & 60.9 & \textbf{88.6} & 55.3 & 55.1 & 63.0 & 69.3 & 61.0 & 80.9 & 63.5 \\
              & Walker2d & 26.0 & 70.6 & 31.2 & 62.5 & 82.5 & 64.4 & \textbf{88.1} & 52.4 & 28.3 & 64.9 & 76.6 & 58.9 & 70.9 & 70.3 \\
\midrule
\multicolumn{2}{c}{Average} & 51.9 & 77.5 & 55.9 & 68.4 & \textbf{77.7} & 72.9 & \textbf{81.0} & 67.4 & 54.3 & 69.2 & 75.8 & 71.6 & \textbf{79.4} & 72.2 \\
\bottomrule
\end{tabular}
}
\label{tab:rl}
\end{table*}

\subsection{Synthetic Dataset}
We first compare different guidance methods on 2-dimensional synthetic datasets where the source distributions are different from Gaussian. 
The base flow matching model is trained to learn the flow with source distributions other than Gaussian. 
During inference, different guidance methods are applied to perform guided sampling with different objective functions $J$ for each dataset. All of the source distributions are non-Gaussian, so traditional diffusion guidance should not be applied. In Figure \ref{fig:toy}, we compare the performance of $g^{\text{MC}}$, $g_\phi$, an exact diffusion guidance called contrastive energy guidance (CEG) proposed by \citet{lu_contrastive_nodate}, and approximate guidance $g^{\text{cov-A}}$, $g^{\text{cov-G}}$, and $g^{\text{sim-MC}}$. The original target distributions (w/o $g_t$) and the $J$-weighted distributions (ground truth) are shown in the first and second columns. 
The details of the experiment are provided in Appendix \ref{app:exp}.

It can be seen from Figure \ref{fig:toy} that $g^{\text{MC}}$ and $g_{\phi}$ generated samples that almost exactly match the ground truth distribution and the performance is consistent across different datasets. 
Note that the generated samples maintain the correct data manifold instead of concentrating on some points as gradient-based approximate guidance methods do.
As has been mentioned in Section \ref{sec:method}, CEG is essentially $\nabla_{x_t}\log Z_t$ which is exact under the \diffusionpath~assumption \ref{assumption:uncoupled_affine_gaussian_path}. However, none of the flow matching paths we have here are \diffusionpath s, so exact diffusion guidance performs poorly compared to $g_\phi$ and $g^{\text{MC}}$, showing a largely distorted generated distribution.

Also, we investigated the asymptotic exactness of $g^{\text{MC}}$. To quantitatively see the increasing guidance precision as the sample size increases, we show the Wasserstein-2 distance between the guided generation distribution and the ground truth target distribution, estimated using $1000$ samples. As shown in Figure \ref{fig:asymptotic}, the error decreases as the number of samples for computing $g^{\text{MC}}$ ($N$ in Algorithm \ref{alg:mc_estimation_on_g}) increases from $5$ to $10^4$, and eventually approaches or surpasses the error of the learned generative model for the original distribution, which can be seen as an approximate lower-bound of the guided generation error.

\subsection{Planning}

\begin{table*}[htb]
\caption{Image inverse problem results on CelebA-HQ. The best and runner-up results are highlighted in bold and underlined.}
\resizebox{\textwidth}{!}{ %
    \centering
    \begin{tabular}{clcccccccccccc}
    \toprule
     & & \multicolumn{4}{c}{Inpainting-Center, $\sigma_y=0.05$} & \multicolumn{4}{c}{Super-Resolution $\times4$, $\sigma_y=0.05$} & \multicolumn{4}{c}{Gaussian Deblurring, $\sigma_y=0.05$} \\
    \cmidrule(lr){3-6} \cmidrule(lr){7-10} \cmidrule(lr){11-14}
     & & FID $\downarrow$ & LPIPS $\downarrow$ & PSNR $\uparrow$ & SSIM $\uparrow$ & FID $\downarrow$ & LPIPS $\downarrow$ & PSNR $\uparrow$ & SSIM $\uparrow$ & FID $\downarrow$ & LPIPS $\downarrow$ & PSNR $\uparrow$ & SSIM $\uparrow$ \\
    \midrule
    & $g^{\text{cov-A}}$ & \textbf{7.330} & 0.1904 & 25.70 & 0.8432 & 26.06 & 0.3016 & 26.64 & 0.7292 & 14.34 & 0.2968 & 24.46 & 0.6982 \\
    & $g^{\text{sim-A}}$ & 10.67 & \textbf{0.1716} & 25.42 & 0.8681 & 31.78 & 0.3717 & 23.88 & 0.6072 & \textbf{11.83} & 0.2837 & 24.54 & 0.7109 \\
    OT-CFM& $g^{\text{cov-G}}$ & 18.43 & 0.2390 & 26.96 & 0.8125 & 15.42 & 0.2533 & \textbf{27.40} & 0.7783 & 17.04 & 0.2873 & 24.96 & 0.7196 \\
    & $\Pi$GDM & 12.56 & \underline{0.1717} & \underline{27.74} & \textbf{0.8723} & \textbf{9.828} & \textbf{0.2322} & \underline{27.24} & \textbf{0.7891} & 12.95 & \textbf{0.2226} & \textbf{28.43} & \textbf{0.7952 }\\
    & $g^{\text{MC}}$ & 22.75 & 0.5589 & 8.67 & 0.3484 & 22.92 & 0.5596 & 8.59 & 0.3468 & 22.82 & 0.5596 & 8.64 & 0.3469 \\
    \midrule
    & $g^{\text{cov-A}}$ & \underline{7.678} & 0.1920 & 25.95 & 0.8414 & 31.76 & 0.3770 & 22.79 & 0.5899 & 16.09 & 0.3052 & 24.21 & 0.6825 \\
    & $g^{\text{sim-A}}$ & 11.54 & 0.1785 & \textbf{28.00} & 0.8686 & 33.02 & 0.3749 & 23.75 & 0.6015 & 13.37 & 0.2947 & 24.34 & 0.6926 \\
    CFM& $g^{\text{cov-G}}$ & 19.65 & 0.2377 & 27.03 & 0.8140 & 13.89 & 0.2461 & 27.15 & \underline{0.7864} & 16.89 & 0.2908 & 24.84 & 0.7112 \\
    & $\Pi$GDM & 15.27 & 0.1753 & 25.48 & \underline{0.8700} & \underline{10.52} & \underline{0.2435} & 26.96 & 0.7755 & \underline{12.60} & \underline{0.2244} & \underline{28.27} & \underline{0.7893} \\
    & $g^{\text{MC}}$ & 26.37 & 0.5615 & 9.06 & 0.3495 & 26.75 & 0.5492 & 10.05 & 0.3689 & 26.84 & 0.5494 & 10.04 & 0.3684 \\
    \bottomrule
    \end{tabular}
}
\label{tab:img_inv}
\end{table*}

We also conduct experiments on offline RL tasks where generative models have been used as planners \citep{janner_planning_2022,chen_flow_2024}. The planning process is realized through sampling from $\frac{1}{Z}p(x_1)e^{R(x_1)}$ \citep{levine_reinforcement_2018}, which aligns with the goal of our guidance by setting $J=-R$, and $R$ being the return.
We report experiment results on the Locomotion tasks in the D4RL dataset \citep{d4rl}, with experiment setting details and complete results provided in Appendix \ref{app:exp_rl}.

The average normalized scores across five seeded runs of each guidance method are reported in Table \ref{tab:rl} where score $=100$ corresponds to the scores of the expert. The conclusions for CFM and OT-CFM are consistent: for gradient-based methods, $g^{\text{cov-G}}$ is generally better than $g^{\text{cov-A}}$ with an average increase in score of $8.0$. $g^{\text{sim-MC}}$ has a performance between $g^{\text{cov-A}}$ and $g^{\text{cov-G}}$.
The improved performance of $g^{\text{cov-G}}$ comes at a higher computation cost of differentiation through the VF model, though. 
The MC-based guidance, although being gradient-free, outperforms the second-best method $g^{\text{cov-G}}$ by $3.5$ on average and is the best method in $7$ out of $9$ tasks.
For $g_\phi$, we report the result of the best losses $\ell$, but their performance is still relatively weak, falling behind the best by $10.4$ on average. We attribute this to the unstable joint training of two networks and provide the complete results in Appendix \ref{app:exp_rl}.

\begin{figure}[!htb]
    \centering
    \includegraphics[width=0.95\linewidth]{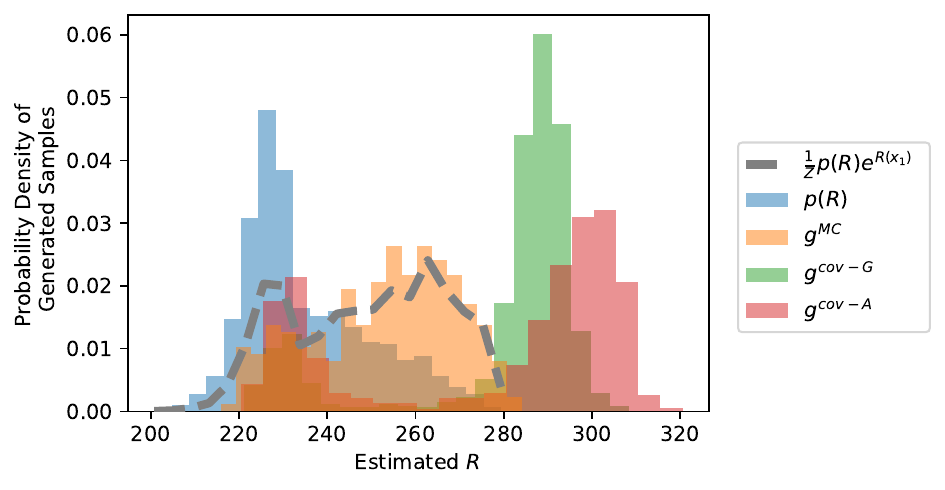}
    \caption{$R$ distribution of generated trajectories in Locomotion. $g^{\text{MC}}$ matches the target gray dashed line well.}
    \label{fig:rl_ablation_J_dist}
\end{figure}

To investigate the effectiveness of $g^{\text{MC}}$, we collect an ensemble of plans that are generated under guidance, compute the critic-predicted objective function value (estimated return $R$), and then plot the density distribution of the estimated return $R$. An ideal guidance will result in the $R$ distribution to be $p(R)e^{R}$ where $p(R)$ is the distribution generated without guidance (Appendix \ref{app:exp_rl}). As can be seen from Figure \ref{fig:rl_ablation_J_dist}, the samples generated with $g^{\text{MC}}$ have a density distribution that best matches the ground-truth target distribution. 

\subsection{Image Inverse Problems}

We conduct experiments on the image inverse problems on the CelebA-HQ (256$\times$256) dataset to reflect the guidance performances on higher dimensional generative tasks. We consider three different types of noised inverse problems: box inpainting, super-resolution, and Gaussian deburring, and compute the metrics FID, LPIPS, PSNR, and SSIM on 3000 samples from the test set. The details of the settings and result visualizations are included in Appendix \ref{app:exp_image}. 

The results demonstrate that $\Pi$GDM is generally better on all tasks, being the best in $8$ out of $12$ metrics. $g^{\text{cov-G}}$ has a similar but slightly worse performance than $\Pi$GDM, being the best or the runner-up in all $4$ metrics of the super-resolution task if ranking the results of CFM and OT-CFM separately and $3$ out of $4$ metrics in the deblurring task. $g^{\text{sim-A}}$ that does not involve the Jacobian shows remarkable performance in inpainting and deblurring, with a lower FID score than $\Pi$GDM by $1.5$ on average, and LPIPS, PSNR, and SSIM within $3\%$ relative difference compared to the best or the runner-up method in 5 out of 6 metrics, especially when considering the efficiency of $g^{\text{sim-A}}$ that no backpropagation through the model is needed. $g^{\text{cov-A}}$ is the worst excluding $g^{\text{MC}}$, being the second worst or the worst in 11 out of 12 metrics and ranking CFM and OT-CFM separately. It should be noted that $g^{\text{MC}}$ performs poorly here, as $J$ in the inverse problem is highly complex, thus requiring an infeasibly large sample budget to compute accurate $g^{\text{MC}}$. A more detailed explanation is deferred to Appendix \ref{app:exp_image}.

\section{Limitations and Future Works}

The major limitation of our work lies in the assumption that $\mathcal{P}\approx 1$. When the coupling is strong, the guidance VF no longer has the correct direction. 
In future works, we plan to address this problem by estimating $\mathcal{P}$, which enables guidance for flow matching models with exact coupling. 

In addition, we plan to improve the specific guidance methods. For example, $g^{\text{MC}}$ suffers from the low sample efficiency on high-dimensional datasets. Thus, it is worthwhile to comprehensively investigate and improve the effectiveness of $g_t^{\text{MC-IS}}$, or to explore other techniques (such as the control variable method) to further lower the variance.

Besides, guided VF $v'_t(x_t)$ can be chosen to have different properties, such as the straightness of the VF. In this way, there may be add-on VFs that facilitate the sampling efficiency, which we also leave for future work. 

\section{Conclusion}
In this work, we have proposed the first framework for general flow matching guidance, from which new MC-based guidance $g^{\text{MC}}$, many approximate guidance, and learned guidance $g_\phi$, are derived, all of them verified by experiments. Many classical guidance methods have been covered as special cases, and we provided a theoretical derivation for each guidance method. We believe this work will facilitate the application of flow matching by opening novel design spaces of its guidance methods.

\section*{Acknowledgements}
We thank Qianyi Chen, Tengfei Xu, Long Wei, Chuanrui Wang, Jiashu Pan, and Tao Zhang for the discussions and for providing feedback on our manuscript. 
We gratefully acknowledge the support of Westlake University Research Center for Industries of the Future and Westlake University Center for High-performance Computing. The content is solely the responsibility of the authors and does not necessarily represent the official views of the funding entities.

\section*{Impact Statement}
This paper presents work whose goal is to advance the field of Machine Learning. There are many potential societal consequences of our work, none of which we feel must be specifically highlighted here.

\bibliography{references}
\bibliographystyle{icml2025}

\newpage
\appendix
\onecolumn

\section{Complete Theoretical Derivations and Proofs}

Here, we provide the theoretical analysis that is deferred from the main text, including the following subsections:
Appendix \ref{app:posterior_sampling_and_energy_guided_sampling} includes proof of how energy-guided sampling from $p(x)e^{-J(x)} / Z$ is equivalent to conditional sampling from $p(x|y)$.
Appendix \ref{app:general_guidance} proves Theorem \ref{theorem:general_guidance} by showing $g_t+v_t$ is equal to $v'_t$ which generates the correctly guided probability path.
Appendix \ref{app:affine_gaussian_guidance_matching} Proof of that under uncoupled affine Gaussian paths, our general guidance $g_t$ is equivalent to the diffusion guidance $\nabla_{x_t} \log Z_t$.
Appendix \ref{app:proposition_match_anything} is proof of Proposition \ref{proposition:matching_anything}.
Appendix \ref{appendix:other_ways_to_learn_z} discusses other ways to obtain $Z_t$, including using contrastive learning and Monte Carlo estimation.
In appendix \ref{app:guidance_matching}, we propose three other training losses $\ell^{VGM}$, $\ell^{RGM}$, and $\ell^{MRGM}$ for $g_\phi$ and prove that the losses will produce the correct gradient.
Appendix \ref{app:independent_mc} includes a more detailed explanation of $g^{\text{MC}}$ and a variant of it under uncoupled paths.
Appendix \ref{app:localized_approximation} derives $g^{\text{local}}$ and proves its error bound.
Appendix \ref{appendix:x1_parameterization} shows how to estimate $\hat{x}_1$ using $x_t$ and $v_\theta(x_t,t)$ under the affine path assumption.
Appendix \ref{appendix:affine_path_cov_local_approx_guidance} proves that $g^{\text{local}}$ becomes $g^{\text{cov}}$ under affine paths.
Appendix \ref{appendix:jacobian_trick} includes the proof of the Jacobian trick (Proposition \ref{appendix:jacobian_trick}).
Appendix \ref{appendix:approximate_simple_posterior_pigdm_like} includes a proof of $g_t^{\text{sim-inv}}$ for the image inverse problem, how to derive $g^{\text{sum-inv-A}}$, and how to recover $\Pi$GDM under the uncoupled affine Gaussian path assumption.

\subsection{Energy Guided Sampling as Posterior Sampling}
\label{app:posterior_sampling_and_energy_guided_sampling}

There exists a $J(x)$ such that sampling from $\frac{1}{Z}p(x)e^{-J(x)}$ is equivalent to sampling from $p(x|y)$.

Simply take $J=-\log p(y|x)$. Plug it in to get
\begin{equation}
    \frac{1}{Z}p(x)e^{-J(x)}=\frac{p(x)e^{\log p(y|x)}}{\int p(x)e^{\log p(y|x)}dx}=\frac{p(x)e^{\log \frac{p(y|x)}{p(y)}}}{\int p(x)e^{\log \frac{p(y|x)}{p(y)}}dx}=p(x|y).
\end{equation}

This theorem can be interpreted as, when you have a classifier $p(y|x)$ and an energy guidance algorithm, you can directly use this algorithm to perform conditional generation from $p(y|x)$ by setting the energy $J(x)=p(y|x)$.

Similar approaches have been used in probability inference, reinforcement learning \citep{levine_reinforcement_2018}, and Diffuser \citep{janner_planning_2022} uses this to convert return-conditioned sampling into energy-guided sampling.

\subsection{General Guidance}
\label{app:general_guidance}
We prove Theorem \ref{theorem:general_guidance} here.
\begin{theorem}
    Adding the guidance VF $g_t(x_t)$ to the original VF $v_t(x_t)$ will form VF $v'_t(x_t)$ that generates $p_t'(x_t) = \int p_t(x_t|z)p'(z)dz$, as long as $g_t(x_t)$ follows:
    \begin{align}
        &g_t(x_t) = \int  \left({\mathcal{P}}\frac{e^{-J(x_1)}}{Z_t(x_t)} - 1\right) v_{t|z}(x_t|z) p(z|x_t) dz, \\
        &\text{where }Z_t(x_t) = \int\mathcal{P} e^{-J(x_1)} p(z|x_t)dz,
    \end{align}
    $\mathcal{P} = \frac{\coupling'(x_0|x_1)}{\coupling(x_0|x_1)}$ is the reverse coupling ratio, where $\coupling'(x_0|x_1)$ is the reverse data coupling for the new VF, i.e., the distribution of $x_0$ given $x_1$ sampled from the target distribution.
\end{theorem}

\textbf{Proof.}
We can subtract $v_t(x_t)$ from $v'_t(x_t)$ to construct $g_t(x_t)$:
\begin{align}
    g_t(x_t) & =  v'_t(x_t) - \int v_{t|z}(x_t|z) \frac{p_t(x_t|z)p(z)}{p_t(x_t)} dz.
\end{align}
$v'_t(x_t)$ generates $p_t(x_t) = \int p_t(x_t|z)p'(z)dz$ if
\begin{equation}
    v'_t(x_t) = \int v_{t|z}(x_t|z) \frac{p_t(x_t|z)p'(z)}{p'(x_t)} dz,
\end{equation}
where $p'(z) = \pi'(x_0|x_1) \frac{1}{Z}p(x_1)e^{-J(x_1)}$, which follows from conditional flow matching, \emph{i.e.}, a VF marginalizing a conditional VF will generate the corresponding marginal probability path \citep{lipman_flow_2023,tong_improving_2024}.
Then, we have a possible form of $g_t(x_t)$
\begin{align}
    \nonumber g_t(x_t) & = \int v_{t|z}(x_t|z) (\frac{p_t(x_t|z)p'(z)}{p'_t(x_t)} - \frac{p_t(x_t|z)p(z)}{p_t(x_t)}) dz \\
    \nonumber& = \int v_{t|z}(x_t|z) (\frac{p_t(x_t|z)\pi'(x_0|x_1)\frac{1}{Z}p(x_1)e^{-J(x_1)}}{p'_t(x_t)} - \frac{p_t(x_t|z)p(z)}{p_t(x_t)}) dz \\ 
    \nonumber& = \int v_{t|z}(x_t|z) (\mathcal{P}(z)\frac{p_t(x_t|z)p(z)\frac{1}{Z}e^{-J(x_1)}}{p'_t(x_t)} - \frac{p_t(x_t|z)p(z)}{p_t(x_t)}) dz \\ 
    & = \int v_{t|z}(x_t|z) \frac{p_t(x_t|z)p(z)}{p_t(x_t)} (\mathcal{P}(z)\frac{1}{Z}e^{-J(x_1)}\frac{p_t(x_t)}{p'_t(x_t)} - 1) dz, \label{eq:appendix:general_guidance_theorem_intermediate}
\end{align}
where $p_t(x_t) = \int p(x_t,z)dz$, $p'(x_t) = \int p'(x_t,z)dz$, and $\mathcal{P}(z) = \frac{\pi'(x_0|x_1)}{\pi(x_0|x_1)}$.
Since
\begin{equation}
    p(x_t)=\int p(x_t,z)dz = p(x_t) \int p(z|x_t)dz\label{eq:appendix:general_guidance_theorem_marginal_pt}
\end{equation}
and 
\begin{align}
     \nonumber p'(x_t)=\int p'(x_t,z)dz &= \int p'(x_t|z) p'(z)dz  \\
     \nonumber& = \int p(x_t|z) \pi'(x_0|x_1) \frac{1}{Z} p(x_1) e^{-J(x_1)}dz \\ 
     \nonumber& = \int \mathcal{P}(z)p(x_t|z) p(z) \frac{1}{Z} e^{-J(x_1)}dz \\ 
     \nonumber& = \int \mathcal{P}(z) p(x_t,z) \frac{1}{Z} e^{-J(x_1)}dz \\ 
     & = p(x_t) \int \mathcal{P}(z) p(z|x_t) \frac{1}{Z} e^{-J(x_1)}dz .\label{eq:appendix:general_guidance_theorem_marginal_ptprime}
\end{align}
Plugging Eq. \eqref{eq:appendix:general_guidance_theorem_marginal_pt} and Eq. \eqref{eq:appendix:general_guidance_theorem_marginal_ptprime} into Eq. \eqref{eq:appendix:general_guidance_theorem_intermediate}, we get 
\begin{align}
    \nonumber g_t(x_t) & = \int v_{t|z}(x_t|z) \frac{p_t(x_t|z)p(z)}{p_t(x_t)} (\mathcal{P}(z)\frac{1}{Z}e^{-J(x_1)}\frac{p_t(x_t)}{p'_t(x_t)} - 1) dz \\ 
    \nonumber& = \int v_{t|z}(x_t|z) \frac{p_t(x_t|z)p(z)}{p_t(x_t)} (\mathcal{P}(z)\cancel{\frac{1}{Z}}e^{-J(x_1)}\frac{\cancel{p(x_t)} \int p(z|x_t)dz}{\cancel{p(x_t)} \int \mathcal{P}(z) p(z|x_t) \cancel{\frac{1}{Z}} e^{-J(x_1)}dz} - 1) dz \\ 
    \nonumber& = \int v_{t|z}(x_t|z) \frac{p_t(x_t|z)p(z)}{p_t(x_t)} (\mathcal{P}(z)e^{-J(x_1)}\frac{1}{\int \mathcal{P}(z) p(z|x_t) e^{-J(x_1)}dz}\underbrace{\int p(z|x_t)dz}_{=1} - 1) dz \\ 
    & = \int v_{t|z}(x_t|z) \frac{p_t(x_t|z)p(z)}{p_t(x_t)} (\frac{\mathcal{P}(z)e^{-J(x_1)}}{\mathbb{E}_{z\sim p(z|x_t)}[\mathcal{P}(z)e^{-J(x_1)}]} - 1) dz. 
\end{align}
Finally, denote $Z_t = \mathbb{E}_{z\sim p(z|x_t)}[e^{-J(x_1)}]$ to complete the proof.

\textit{Remark.} The theorem states that $v'_t = g_t + v_t$ not only generates the desired terminal distribution $\frac{1}{Z}p(x_1)e^{-J(x_1)}$ at time $t=1$, but also generates a probability path $p'_t(x_t)$ that is similar to the original one. Specifically, their "noising process" $p(x_t|x_1)$, and the conditional vector fields $v(x_t|x_1)$, and the reverse coupling $p(x_0|x_1)$ are the same. These are all hyperparameters of flow matching, as one can choose an arbitrary conditional vector field satisfying boundary conditions and the conditional vector field uniquely determines the conditional probability path; the reverse coupling, given target (dataset) distribution $p(x_1)$ or $\frac{1}{Z}p(x_1)e^{-J(x_1)}$, composes the data coupling $p(x_0,x_1) = p(z)$ for flow matching training.

It should be noted that the $g_t$ and $v'_t$ we construct here is only one of infinitely many \citep{lipman_flow_2023} possible vector fields to generate $\frac{1}{Z}p(x)e^{-J(x_1)}$ at $t=1$. It remains an interesting question whether there exists better $v'_t$ that, for example, simplifies $g_t$ or improves the vector field by straightening the flow.

\subsection{The $\mathcal{P}=1$ Approximation}
\label{app:p=1_approximation}

\begin{table}[h]
\begin{tabular}{llllll}
\toprule
Flow time              & 0.05                  & 0.25                  & 0.5                   & 0.75                  & 0.95                  \\
\midrule
Relative L2            & $0.0382\pm0.0076$     & $0.0297\pm0.0043$     & $0.0271\pm0.0032$     & $0.0312\pm0.0033$     & $0.0717\pm 0.0096$    \\
\bottomrule
\end{tabular}
\caption{The relative L2 of the difference between the VF of OT-CFM and CFM at different time steps on the sampling trajectory.}
\label{tab:app_p=1_approximation}
\end{table}

From the empirical perspective, we found that $P=1$ is a valid choice for dependent couplings in realistic datasets. As we show in Table \ref{tab:app_p=1_approximation}, the VF of the OT-CFM (batch size 128) in and uncoupled CFM trained on CelebA 256$\times$256 have small relative error at all flow time steps, so their ideal guidance VFs are approximately the same, validating our approximation of $P=1$.

Next, we give a more intuitive understanding of the approximation.
Our framework allows us to choose any $\pi'(x_0|x_1)$ (and hence $P$) as long as the source distribution is consistent: $p(x_0) = \int\pi'(x_0|x_1)\frac{1}{Z}p(x_1)e^{-J(x_1)}dx_1$. In other words, the error induced by setting $P=1$ can be characterized by the deviation in either the source distribution or the vector field. 
In the former case, the guidance VF is essentially considered exact, which implies $\pi'(x_0|x_1) = \pi(x_0|x_1)$. Therefore, the error is induced by the fact that we should have sampled from $\int\pi'(x_0|x_1)\frac{1}{Z}p(x_1)e^{-J(x_1)}dx_1$, rather than the original $p(x_0)$. In the latter case, we actually assumed the source distribution to be unchanged, i.e., we need $\pi'(x_0|x_1)=p(x_0)$ to make the source distributions compatible automatically. In this case, the error is caused by approximating $\mathcal{P}=\frac{p(x_0)}{\pi(x_0|x_1)}$ with $1$.

We now discuss the practical effect of the approximation, i.e., when it is a good approximation and when it is not.

In the case of strongly dependent couplings, $P\approx 1$ still holds as long as $J$ varies slowly. This is demonstrated by the small deviation in the error of the source distribution (assuming $\pi'(x_0|x_1)=\pi(x_0|x_1)$) as we discussed above. If $J$ is always near its average value, the new source distribution $\int\pi(x_0|x_1)\frac{1}{Z}p(x_1)e^{-J(x_1)}dx_1$ is almost $\int\pi(x_0|x_1)p(x_1)dx_1 = p(x_0)$ that is the original source.

Nevertheless, when both the coupling is strong and $J$ varies intensively, a more complicated treatment is required for exact guidance. For example, we can try to sample from the new source distribution $\int\pi(x_0|x_1)\frac{1}{Z}p(x_1)e^{-J(x_1)}dx_1$. Although one may argue that this may be equally difficult as sampling from the target distribution $\frac{1}{Z}p(x_1)e^{-J(x_1)}$ exactly, it may be learned more easily as the target distribution is potentially smoothed after being convolved with the "kernel" $\pi(x_0|x_1)$. We leave this more detailed treatment of $\mathcal{P}$ to future work.

\subsection{Uncoupled Affine Gaussian Guidance}
\label{app:affine_gaussian_guidance_matching}

Here we prove that $\nabla_{x_t} \log Z_t(x_t)$ is proportional to the guidance $g_t$ in Eq. \eqref{eq:general_guidance}. 
Note that the term $\nabla_{x_t} \log Z_t(x_t)$ is widely used as guidance in the diffusion model literature \citep{dhariwal_diffusion_2021,ho_classifier-free_2022,song_score-based_2021,song_loss-guided_2023,song_pseudoinverse-guided_2022,janner_planning_2022,ajay_is_2023}. Therefore, we prove that our general flow matching guidance exactly covers the original diffusion guidance under the affine Gaussian path assumption, \emph{i.e.}, when flow matching falls back to the diffusion model. Our proof here also elucidates how the gradient $\nabla_{x_t}$ emerges from the original expression of the general guidance for flow matching in Eq. \eqref{eq:general_guidance} where there is no apparent gradient.

We restate Eq. \eqref{eq:general_guidance} here:
\begin{align}\label{eq:appendix_restate:general_guidance}
    g_t(x_t) & = \int (\frac{e^{-J(x_1)}}{Z_t(x_t)} - 1) v_{t|z}(x_t|z) \frac{p_t(x_t|z)p(z)}{p_t(x_t)} dz \\
    Z_t(x_t) & =  \int e^{-J(x_1)} \frac{p_t(x_t|z)p(z)}{p_t(x_t)} dz
\end{align}

Assuming the flow matching to be of uncoupled affine path, we have
\begin{equation}
    x_t = \sigma_t x_0 + \alpha_t x_1,
\end{equation}
where $\sigma_t$ and $\alpha_t$ are schedulers satisfying boundary conditions $\sigma_0 = \alpha_1 = 1$, $\sigma_1 = \alpha_0 = 0$.
Thus, 
\begin{align}
\nonumber
    v_{t|z}(x_t|z) & = \dot\sigma_t x_0 + \dot\alpha_t x_1 
    \\
    \label{eq:appendix:general_guidance_and_diffusion_guidance_affine_path}
    & =  \underbrace{\frac{\dot\sigma_t}{\sigma_t}}_{a_t} x_t + \underbrace{\frac{1}{\sigma_t} (\dot\alpha_t \sigma_t - \dot\sigma_t \alpha_t)}_{b_t} x_1 
\end{align}
where $\dot f_t \coloneqq \frac{d f}{d t}$ denotes derivative to time $t$, and we define $a_t\coloneqq \frac{\dot\sigma_t}{\sigma_t}$, $b_t\coloneqq \frac{1}{\sigma_t} (\dot\alpha_t \sigma_t - \dot\sigma_t \alpha_t)$.

First, we demonstrate a useful technique for the proof later.
Inserting Eq. \eqref{eq:appendix:general_guidance_and_diffusion_guidance_affine_path} into $g_t(x_t)$ in Eq. \eqref{eq:appendix_restate:general_guidance} and we have
\begin{align}\label{eq:appendix:general_guidance_expanded_v}
    g_t(x_t) = \int (\frac{e^{-J(x_1)}}{Z_t} - 1) (a_t x_t + b_t x_1) \frac{p_t(x_t|z)p(z)}{p_t(x_t)} dz.
\end{align}
Since $Z_t = \mathbb{E}_{z\sim p(z|x_t)}[e^{-J(x_1)}]$, 
\begin{equation}
    \int (\frac{e^{-J(x_1)}}{Z_t} - 1) a_t x_t \frac{p_t(x_t|z)p(z)}{p_t(x_t)} dz = 0.
    \label{eq:appendix:general_guidance_and_diffusion_guidance_integrate_to_zero}
\end{equation}
That is to say, $x_t$ inside the integral of Eq. \eqref{eq:appendix:general_guidance_and_diffusion_guidance_integrate_to_zero} will integrate to zero, and we can freely remove or add it to construct desired terms.

Recall the assumption of uncoupled Gaussian path, \emph{i.e.} $p(x_0,x_1) = p(x_0) p(x_1)$, $p(x_0) = \mathcal{N}(x_0;0,I)$. We can utilize the important fact that the conditional probability path for affine Gaussian path flows satisfies $x_t \sim \mathcal{N}(x_t;\alpha_t x_1, \sigma_t^2 I)$, which allows us to connect the conditional score to $x_1$
\begin{equation}\label{eq:appendix:general_guidance_and_diffusion_guidance_score_and_affine_path}
    \nabla_{x_t} \log p(x_t|x_1) = -\frac{x_t - \alpha_t x_1}{\sigma_t ^2}.
\end{equation}
Using Eq. \eqref{eq:appendix:general_guidance_and_diffusion_guidance_integrate_to_zero} and Eq. \eqref{eq:appendix:general_guidance_and_diffusion_guidance_score_and_affine_path}, Eq. \eqref{eq:appendix:general_guidance_expanded_v} can be further converted to 
\begin{align}\label{eq:appendix:general_guidance_and_diffusion_guidance_convert_conditional_v_to_score}
    g_t(x_t)& = \int (\frac{e^{-J(x_1)}}{Z_t} - 1) (b_t x_1 \underbrace{- \frac{b_t}{\alpha_t} x_t}_{\text{Eq. \eqref{eq:appendix:general_guidance_and_diffusion_guidance_integrate_to_zero}}}) \frac{p_t(x_t|x_1)p(x_1)}{p_t(x_t)} dx_1 \\ \nonumber
    & = \frac{b_t \sigma_t^2}{\alpha_t} \int (\frac{e^{-J(x_1)}}{Z_t} - 1) \nabla_{x_t} \log p(x_t|x_1) \frac{p_t(x_t|x_1)p(x_1)}{p_t(x_t)} dx_1 \\\nonumber
    & = \frac{b_t \sigma_t^2}{\alpha_t} \int (\frac{e^{-J(x_1)}}{Z_t} - 1) \left(\underbrace{\nabla_{x_t} \log p(x_t) + \nabla_{x_t} \log p(x_1 | x_t)}_{\text{Bayes' rule}}\right) \frac{p_t(x_t|x_1)p(x_1)}{p_t(x_t)} dx_1
    \\
    & = \frac{b_t \sigma_t^2}{\alpha_t} \int (\frac{e^{-J(x_1)}}{Z_t} - 1) \left(\underbrace{\cancel{\nabla_{x_t} \log p(x_t)}}_{\text{Integrates to zero as in Eq. \eqref{eq:appendix:general_guidance_and_diffusion_guidance_integrate_to_zero}}} + \nabla_{x_t} \log p(x_1 | x_t)\right) \frac{p_t(x_t|x_1)p(x_1)}{p_t(x_t)} dx_1 \\
    & = \frac{b_t \sigma_t^2}{\alpha_t} \int (\frac{e^{-J(x_1)}}{Z_t} - 1)  \nabla_{x_t} \log p(x_1 | x_t) \frac{p_t(x_t|x_1)p(x_1)}{p_t(x_t)} dx_1. \label{eq:appendix:general_guidance_and_diffusion_guidance_samplified_score}
\end{align}
Notice in Eq. \eqref{eq:appendix:general_guidance_and_diffusion_guidance_samplified_score} that 
\begin{align}\label{eq:appendix:general_guidance_and_diffusion_guidance_canceling_expectation_of_score}
    \nonumber
    &\int \nabla_{x_t} \log p(x_1 | x_t) \frac{p_t(x_t|x_1)p(x_1)}{p_t(x_t)} dx_1 \\
    \nonumber= & \int \underbrace{p(x_1|x_t)  \nabla_{x_t} \log p(x_1 | x_t)}_{\text{Since } f \nabla  \log f = \nabla f}  dx_1 \\
    \nonumber= & \int \nabla_{x_t} p(x_1 | x_t)  dx_1 \\
    \nonumber= &\, \nabla_{x_t} \int p(x_1 | x_t)  dx_1 \\
    = &\,0.
\end{align}
Therefore 
\begin{align}
    \nonumber g_t(x_t) & = -\frac{b_t \sigma_t^2}{\alpha_t} \int (\frac{e^{-J(x_1)}}{Z_t} \cancel{-1})  \nabla_{x_t} \log p(x_1 | x_t) \frac{p_t(x_t|x_1)p(x_1)}{p_t(x_t)} dx_1 \\
    \nonumber& = \frac{b_t \sigma_t^2}{\alpha_t} \int \frac{e^{-J(x_1)}}{Z_t(x_t)} \underbrace{p_t(x_1|x_t) \nabla_{x_t} \log p(x_1 | x_t)}_{\text{Using again $f \nabla \log f = \nabla f$}}  dx_1 \\
    \nonumber& = \frac{b_t \sigma_t^2}{\alpha_t} \frac{1}{Z_t(x_t)} \int e^{-J(x_1)} \nabla_{x_t} p(x_1 | x_t)  dx_1 \\
    \nonumber& = \frac{b_t \sigma_t^2}{\alpha_t} \frac{1}{Z_t(x_t)}\underbrace{\nabla_{x_t} \int e^{-J(x_1)} p(x_1 | x_t)  dx_1}_{\text{Absorb $e^{-J(x_1)}$ for it is independent of $x_1$, and exchange with integral}} \\
    \nonumber& = \frac{b_t \sigma_t^2}{\alpha_t} \frac{1}{Z_t(x_t)} \nabla_{x_t} \underbrace{Z_t(x_t)}_{\text{$Z_t$'s definition in Eq. \eqref{eq:general_guidance}}} \\
    & = \frac{b_t \sigma_t^2}{a_t} \nabla_{x_t} \log Z_t(x_t).
\end{align}

Another possible way to derive this is to first prove the vector field in affine Gaussian path flow matching to be affine to the marginal score, and we direct interested readers to \citep{zheng_guided_2023}. 

\begin{remark}
    The above derivation opens the possibility of using diffusion guidance into affine Gaussian path flow matching, \emph{i.e.}, by multiplying a scheduler $-\frac{b_t \sigma_t^2}{\alpha_t} = \frac{\sigma_t(\dot\alpha_t\sigma_t - \dot \sigma_t\alpha_t)}{\alpha_t} $ to the diffusion classifier guidance. The most common scheduler for flow matching is $\sigma_t = 1 - t, \alpha_t = t$ \citep{lipman_flow_2023,pokle_training-free_2024,zheng_guided_2023,tong_improving_2024,liu_flow_2022,lipman_flow_2024}. In this case, the guidance scheduler is $\frac{(1-t)}{t}$. It should be noted that this scheduler explodes near $t=0$, thus being unstable. The flow matching schedule $\sigma_t$ and $\alpha_t$ can be chosen as other ways to avoid this instability.
\end{remark} 

\begin{remark}
    Note that this guidance cannot be applied to coupled paths. Central to the proof is that in uncoupled affine Gaussian paths, we can convert the conditional vector field to the conditional score.
    If we could do this in coupled paths, we would require (1) $p_t(x|z)$ is Gaussian $\mathcal{N}(x;\mu_t,\sigma_t I)$, such that $\nabla_{x_t} \log p_t(x_t|z) \propto x_t - \mu_t$.  and (2) $v_{t|z}=\dot\mu_t + \dot \sigma_t (x_t - \mu_t) \propto \mu_t$, such that in Eq. \eqref{eq:appendix:general_guidance_and_diffusion_guidance_convert_conditional_v_to_score} the conditional vector field can be converted to the conditional score. Therefore, the following equation must hold \emph{for any $x_t,x_0,x_1$} 
    \begin{equation}
        \dot\mu_t + \dot \sigma_t (x_t - \mu_t) \propto \mu_t
    \end{equation}
    inside the integral of Eq. \eqref{eq:appendix:general_guidance_and_diffusion_guidance_convert_conditional_v_to_score}
    where $\mu_t = \xi_t x_0 + \eta_t x_1$. This equivalent to that
    \begin{equation}
        (\dot\xi_t - \dot \sigma_t \xi_t) x_0 + (\dot\eta_t - \dot \sigma_t \eta_t) x_1 +  \dot \sigma_t x_t \propto \xi_t x_0 + \eta_t x_1
    \end{equation}
    must hold \emph{for any $x_t,x_0,x_1$} inside the integral of Eq. \eqref{eq:appendix:general_guidance_and_diffusion_guidance_convert_conditional_v_to_score}. According to Eq. \eqref{eq:appendix:general_guidance_and_diffusion_guidance_integrate_to_zero}, $x_t$ terms will integrate to zero, thus
    \begin{equation}
        \eta_t(\dot\xi_t - \dot \sigma_t \xi) = \xi_t(\dot\eta_t - \dot \sigma_t \eta_t)
    \end{equation}
    which cannot hold\footnote{Unless $\xi_t=1$ which falls back to the uncoupled case.}: because of the boundary conditions $\xi_0 = \eta_1=1$ and $\xi_1=\eta_0 = 0$, $\exists t\in(0,1),\frac{d \log \xi_t}{ dt} \neq \frac{d \log \eta_t}{ dt}$. %
    It can be observed that the reason why this guidance does not apply to coupled paths is that $x_0,x_1$, and $x_t$ are all independent variables here, preventing us from canceling two of $x_0$ to avoid matching the schedulers' ratio.

\end{remark}

\subsection{Proof of Proposition \ref{proposition:matching_anything}}
\label{app:proposition_match_anything}
We prove proposition \ref{proposition:matching_anything} here.
\begin{proposition}
    Any \emph{marginal variable} $f(x_t,t)\coloneqq \mathbb{E}_{z\sim p_t(z|x_t)}[f_{t|z}(x_t,z,t)],~z=(x_0,x_1)$ has an intractable \emph{marginal loss}
    \begin{equation}\label{eq:app_proposition_matching_anything_marginal_loss}
        \mathcal{L}=\mathbb{E}_{x_t\sim p(x_t)}\left[\left\|f_{{\theta}}(x_t,t) - \mathbb{E}_{z\sim p_t(z|x_t)}[f_{t|z}(x_t,z,t)]\right\|_2^2\right],
    \end{equation}
    whose gradient is identical to the tractable \emph{conditional loss}
    \begin{equation}
        \mathcal{L}_{t|z}=\mathbb{E}_{x_t,z\sim p(x_t,z)}\left[\left\|f_{\theta}(x_t,t) - f_{t|z}(x_t,z,t)\right\|_2^2\right].
    \end{equation}
\end{proposition}

\textbf{Proof.} Expand and take gradient w.r.t. Eq. \eqref{eq:app_proposition_matching_anything_marginal_loss} to get
\begin{align}
    \nonumber\nabla_\theta \mathcal{L}_t
    &=\nabla_\theta \mathbb{E}_{x_t\sim p(x_t)}\left[\left\|f_{{\theta}}(x_t,t) - \mathbb{E}_{z\sim p_t(z|x_t)}[f(x_t,z,t)]\right\|_2^2\right] \\
    \nonumber&=\mathbb{E}_{x_t\sim p(x_t)}\left[\nabla_\theta \left\|f_{{\theta}}(x_t,t) - \mathbb{E}_{z\sim p_t(z|x_t)}[f(x_t,z,t)]\right\|_2^2\right] \\
    \nonumber&=\int \nabla_\theta p(x_t) \left\|f_{{\theta}}(x_t,t) - \mathbb{E}_{z\sim p_t(z|x_t)}[f(x_t,z,t)]\right\|_2^2 dx_t \\
    \nonumber&=\int \nabla_\theta p(x_t) \left(\|f_{{\theta}}(x_t,t)\|^2 - 2 \langle f_{{\theta}}(x_t,t),  \int p_t(z|x_t) dz f(x_t,z,t) \rangle\right)  dx_t \\ 
    \nonumber&=\int \nabla_\theta  p_t(z|x_t) p(x_t) \left(\|f_{{\theta}}(x_t,t)\|^2 - 2 \langle f_{{\theta}}(x_t,t),  f(x_t,z,t) \rangle\right)  dx_t dz \\ 
    \nonumber&=\int  p_t(z|x_t) p(x_t) \nabla_\theta  \left(\|f_{{\theta}}(x_t,t)\|^2 - 2 \langle f_{{\theta}}(x_t,t),  f(x_t,z,t) \rangle\right)  dx_t dz \\ 
    \nonumber&=\mathbb{E}_{z,x_t\sim p_t(z|x_t) p(x_t)} \left[ \nabla_\theta \left\|f_{{\theta}}(x_t,t) - f(x_t,z,t) \right\|^2 \right]\\ 
    &=\nabla_\theta \underbrace{\mathbb{E}_{z,x_t\sim p_t(z|x_t) p(x_t)} \left[ \left\|f_{{\theta}}(x_t,t) - f(x_t,z,t) \right\|^2 \right]}_{\mathcal{L}_{1|z}}.
\end{align}
Thus, the gradient of the \emph{marginal loss} $\mathcal{L}_{t}$ is identical to the gradient of the \emph{conditional loss} $\mathcal{L}_{t|z}$.

\subsection{Other Ways to Obtain $Z_{\phi_Z}$}
\label{appendix:other_ways_to_learn_z}

\citet{lu_contrastive_nodate} proposed to use contrastive learning to train $Z_{\phi_Z}$. The proof already applies to any uncoupled path, and we show that $Z_{\phi_Z}$ does not depend on the coupling
\begin{align}
    Z_t =& \mathbb{E}_{x_1\sim p(x_0,x_1|x_t)}[e^{-J(x_1)}]= \int e^{-J(x_1)}p(x_0|x_1,x_t)p(x_1|x_t) dx_0 dx_1 \\
    =& \int e^{-J(x_1)}p(x_1|x_t) dx_1 = \mathbb{E}_{x_1\sim p(x_1|x_t)}[e^{-J(x_1)}].
\end{align}
That is, instead of actually sampling from $p(x_0,x_1|x_t)$, sampling from $p(x_1|x_t)$ will result in the same $Z_t$. In the case of the coupled path, the marginalized distribution is identical to the uncoupled path case. Therefore, the contrastive learning method can be readily applied to train $Z_{\phi_Z}$.

Besides training-based $Z_{\phi_Z}$, we can also use Monte Carlo estimation to obtain $Z_t$. Notice that by using importance sampling, we have
\begin{equation}
     Z_t = \mathbb{E}_{x_1\sim p(x_0,x_1|x_t)}[e^{-J(x_1)}] = \mathbb{E}_{x_1\sim p(x_0,x_1)}\left[\frac{p(x_t|x_0,x_1)}{p(x_t)}e^{-J(x_1)}\right].
\end{equation}
As long as $p(x_t|x_0,x_1)$ is known (which is often the case \citep{lipman_flow_2023,tong_improving_2024}), we can estimate $Z_t$ by sampling $N$ pairs of $x_0^i,x_1^i$ from $p(x_0,x_1)$ and estimate
\begin{equation}
    \tilde{Z}_t = \sum_i^N \left(\frac{p(x_0^i,x_1^i|x_t)}{\sum_j^N p(x_0^j,x_1^j|x_t)}e^{-J(x_1^i)}\right).
\end{equation}
A similar technique is used in Section \ref{sec:g_mc}.

\subsection{Guidance Matching Losses}
\label{app:guidance_matching}
Here, we prove that the loss in guidance matching is correct and show there are three other equivalent training losses $\ell^{\text{VGM}},\ell^{\text{RGM}},\ell^{\text{MRGM}}$.
The expressions of different losses are summarized below, and their proof follows.

\paragraph{VF-added Guidance Matching (VGM) Loss.} By utilizing the learned VF $v_\theta(x_t, t)$ into Eq. \eqref{eq:guidance_matching_loss_g_1}, we have 
\begin{equation}
    \ell_\phi^{\text{VGM}} = \left\|g_{\phi}(x_t,t) + v_\theta(x_t,t) - \frac{e^{-J(x_1)}}{Z_{\phi_Z,sg}(x_t,t)}v_{t|z}(x_t|z)\right\|_2^2.
\end{equation}

\paragraph{Reweight Guidance Matching (RGM) Loss.} $\ell_\phi^{\text{VGM}}$ can be further shown equivalent to 
\begin{equation}\label{eq:guidance_matching_loss_g_3_loss}
    \ell_\phi^{\text{RGM}}=\frac{e^{-J(x_1)}}{Z_{\phi_Z,sg}(x_t,t)} \left\|g_{\phi}(x_t,t) + v_\theta(x_t,t) - v_{t|z}(x_t|z)\right\|_2^2.
\end{equation}
This training loss steers $g_\phi$ to where $e^{-J(x_1)}$ is larger by assigning a large loss to steer $g_t$ towards high $e^{-J(x_1)}$ regions.

\paragraph{Marginalized Reweight Guidance Matching (MRGM) Loss.} The above loss can be re-assigned a weight, which will result in the same optimal $g_{\phi_2}(x_t,t)$ as in Eq. \eqref{eq:guidance_matching_loss_g_1}. Specifically, by changing $Z_{\phi_Z,sg}(x_t,t)$ to its expectation under $p_t(x_t)$, we have the following equivalent loss
\begin{equation}\label{eq:guidance_matching_loss_g_4_loss}
    \ell_\phi^{\text{MRGM}}=\frac{e^{-J(x_1)}}{Z} \|g_{\phi_2}(x_t,t) + v_\theta(x_t,t) - v_{t|z}(x_t|z)\|_2^2,
\end{equation}
where $Z = \mathbb{E}_{x_1\sim p(x_1)}[e^{-J(x_1)}]$.
$\ell_\phi^{\text{MRGM}}$ is identical to a newly proposed fine-tuning loss in \citet{anonymous2025energyweighted}. It can also be derived via importance sampling in Eq. \ref{eq:guidance_matching_loss_sum} and
similar reweighting-based fine-tuning losses have been studied in the literature of diffusion models \citep{fan_dpok_2023}.

\paragraph{(1) Guidance Matching Loss $\ell^{\text{GM}}$}
By using proposition \ref{proposition:matching_anything}, the following conditional loss
\begin{equation}\label{eq:appendix:guidance_matching_loss_g_1}
    \mathcal{L}_{\phi}^{\text{GM}} = \mathbb{E}_{t\sim\mathcal{U}(0,1),z\sim p(z),x_t \sim p(x_t|z)}\left[\underbrace{\left\|g_{\phi}(x_t,t) - (\frac{e^{-J(x_1)}}{Z_{\phi_Z,sg}(x_t,t)} - 1) v_{t|z}(x_t|z)\right\|_2^2}_{=\ell^{\text{GM}}}\right]
\end{equation}
has a gradient that is equivalent to the marginal loss
\begin{equation}
    \mathbb{E}_{t\sim\mathcal{U}(0,1),z\sim p(z)x_t \sim p(x_t|z)}\left[\left\|g_{\phi}(x_t,t) - \underbrace{\mathbb{E}_{z\sim p(z|x_t)}\left[(\frac{e^{-J(x_1)}}{Z_{\phi_Z,sg}(x_t,t)} - 1) v_{t|z}(x_t|z)\right]}_{=g_t(x_t)}\right\|_2^2\right].
\end{equation}
Therefore, using the loss $\mathcal{L}_{\phi_Z}$ we can train $g_\phi$ to matching $g_t$. Recall that $\mathcal{L}$ in Eq. \eqref{eq:guidance_matching_loss_g_1} is identical
to $\mathcal{L}_{\phi_Z}$, and we proved the validity of the guidance matching training.

\paragraph{(2) VF-added Guidance Matching Loss $\ell^{\text{VGM}}$.} By replacing the learned VF $v_\theta(x_t, t)$ into Eq. \eqref{eq:appendix:guidance_matching_loss_g_1}, we show that  
\begin{equation}
\label{eq:appendix:guidance_matching_loss_g_2}
    \mathcal{L}_{\phi}^{\text{VGM}} = \mathbb{E}_{t\sim\mathcal{U}(0,1),z\sim p(z),x_t \sim p(x_t|z)}\left[\left\|g_{\phi}(x_t,t) + v_\theta(x_t,t) - \frac{e^{-J(x_1)}}{Z_{\phi_Z,sg}(x_t,t)}v_{t|z}(x_t|z)\right\|_2^2\right],
\end{equation}
has a gradient equal to that of $\mathcal{L}_{\phi}$ in Eq. \eqref{eq:appendix:guidance_matching_loss_g_1}.

Expand Eq. \eqref{eq:appendix:guidance_matching_loss_g_1} to get
\begin{align}
    \nonumber\mathcal{L}_{\phi}^{\text{GM}} & = \mathbb{E}_{t\sim\mathcal{U}(0,1),z\sim p(z),x_t \sim p(x_t|z)}\left[\left\|g_{\phi}(x_t,t) - (\frac{e^{-J(x_1)}}{Z_{\phi_Z,sg}(x_t,t)} - 1) v_{t|z}(x_t|z)\right\|_2^2\right] \\
    \nonumber & = \mathbb{E}_{t\sim\mathcal{U}(0,1),z\sim p(z),x_t \sim p(x_t|z)}[\|\underbrace{g_{\phi}(x_t,t)\|_2^2}_{\text{dependent on }\phi} + \|\frac{e^{-J(x_1)}}{Z_{\phi_Z,sg}(x_t,t)}v_{t|z}(x_t|z)\|_2^2 + \|v_{t|z}(x_t|z)\|_2^2 \\
    \label{eq:appendix:guidance_matching_loss_g1_expanded}& \underbrace{-2\langle g_{\phi}(x_t,t), \frac{e^{-J(x_1)}}{Z_{\phi_Z,sg}(x_t,t)}v_{t|z}(x_t|z)\rangle}_{\text{dependent on }\phi} - 2\langle\frac{e^{-J(x_1)}}{Z_{\phi_Z,sg}(x_t,t)}v_{t|z}(x_t|z), v_{t|z}(x_t|z)\rangle \underbrace{- 2\langle v_{t|z}(x_t|z), g_{\phi}(x_t,t)\rangle}_{\text{dependent on }\phi} ]. 
\end{align}
After taking gradient w.r.t. $\phi$, only the terms 
\begin{equation}
    \nabla_{\phi}\mathbb{E}_{t\sim\mathcal{U}(0,1),z\sim p(z),x_t \sim p(x_t|z)}
    \left[
    \|g_{\phi}(x_t,t)\|_2^2 - 2\langle g_{\phi}(x_t,t), \frac{e^{-J(x_1)}}{Z_{\phi_Z,sg}(x_t,t)}v_{t|z}(x_t|z)\rangle
    - 2\langle v_{t|z}(x_t|z), g_{\phi}(x_t,t)\rangle
    \right]
\end{equation}
survive. Therefore, by assuming a perfectly learned $v_\theta(x_t,t)$, \emph{i.e.}, 
\begin{equation}
    v_\theta(x_t,t) = \mathbb{E}_{z\sim p(z|x_t)}\left[ v_{t|z}(x_t|z) \right],
\end{equation}
we have 
\begin{align}
\nonumber&\mathbb{E}_{t\sim\mathcal{U}(0,1),z\sim p(z),x_t \sim p(x_t|z)}
\left[
    \langle v_{t|z}(x_t|z), g_{\phi}(x_t,t)\rangle
\right]  \\
\nonumber= &
\mathbb{E}_{t\sim\mathcal{U}(0,1),z\sim p(z|x_t),x_t \sim p(x_t)}
\left[
    \langle v_{t|z}(x_t|z), g_{\phi}(x_t,t)\rangle
\right] \\
\nonumber= & \mathbb{E}_{\tilde{z}\sim p(\tilde{z}|x_t)}\left[\mathbb{E}_{t\sim\mathcal{U}(0,1),z\sim p(z|x_t),x_t \sim p(x_t)}
\left[
    \langle v_{t|z}(x_t|z), g_{\phi}(x_t,t)\rangle
\right]
\right] \\
\nonumber= & \mathbb{E}_{\tilde{z}\sim p(\tilde{z}|x_t)}\left[\mathbb{E}_{t\sim\mathcal{U}(0,1),x_t \sim p(x_t)}
\left[
    \langle \mathbb{E}_{z\sim p(z|x_t)} [v_{t|z}(x_t|z)], g_{\phi}(x_t,t)\rangle
\right]
\right] \\
\nonumber= & \mathbb{E}_{z\sim p(z|x_t)}\left[\mathbb{E}_{t\sim\mathcal{U}(0,1),x_t \sim p(x_t)}
\left[
    \langle \mathbb{E}_{\tilde{z}\sim p(\tilde{z}|x_t)} [v_{t|z}(x_t|z)], g_{\phi}(x_t,t)\rangle
\right]
\right] \\
\label{eq:appendix:guidance_matching_loss_g_2_v_term}
= & \mathbb{E}_{t\sim\mathcal{U}(0,1),z\sim p(z|x_t),x_t \sim p(x_t)}
\left[
    \langle v_\theta(x_t,t), g_{\phi}(x_t,t)\rangle
\right], 
\end{align}
so by adding back terms that the gradient is agnostic to, we can see that the new loss  $\mathcal{L}_{\phi}^{(1)}$ in Eq. \eqref{eq:appendix:guidance_matching_loss_g_2} is equivalent to $\mathcal{L}_{\phi}$ in Eq. \eqref{eq:appendix:guidance_matching_loss_g_1}
\begin{align}
    \nonumber\nabla_{\phi}{\mathcal{L}}_{\phi}^{\text{GM}} & = \nabla_{\phi}\mathbb{E}_{t\sim\mathcal{U}(0,1),z\sim p(z),x_t \sim p(x_t|z)}\left[\left\|g_{\phi}(x_t,t) - (\frac{e^{-J(x_1)}}{Z_{\phi_Z,sg}(x_t,t)} - 1) v_{t|z}(x_t|z)\right\|_2^2\right] \\
    \nonumber& = \nabla_{\phi}\mathbb{E}_{t\sim\mathcal{U}(0,1),z\sim p(z),x_t \sim p(x_t|z)}[\|{g_{\phi}(x_t,t)\|_2^2} + \|\frac{e^{-J(x_1)}}{Z_{\phi_Z,sg}(x_t,t)}v_{t|z}(x_t|z)\|_2^2 + \underbrace{\|v_{\theta}(x_t,t)\|_2^2}_{\text{Vanishes after $\nabla_\phi$}} \\
    & {-2\langle g_{\phi}(x_t,t), \frac{e^{-J(x_1)}}{Z_{\phi_Z,sg}(x_t,t)}v_{t|z}(x_t|z)\rangle} - 2\langle\frac{e^{-J(x_1)}}{Z_{\phi_Z,sg}(x_t,t)}v_{t|z}(x_t|z), v_{t|z}(x_t|z)\rangle \underbrace{- 2\langle v_\theta(x_t,t), g_{\phi}(x_t,t)\rangle}_{\text{Changed $v_{t|z}$ to $v_\theta$ using Eq. \eqref{eq:appendix:guidance_matching_loss_g_2_v_term}}} ]\\
    \nonumber& =\nabla_{\phi} \mathbb{E}_{t\sim\mathcal{U}(0,1),z\sim p(z),x_t \sim p(x_t|z)}\left[\underbrace{\left\|g_{\phi}(x_t,t) + v_\theta(x_t,t) - \frac{e^{-J(x_1)}}{Z_{\phi_Z,sg}(x_t,t)}v_{t|z}(x_t|z)\right\|_2^2}_{\coloneqq\ell^{\text{VGM}}}\right]\\
    & = \nabla_{\phi} {\mathcal{L}}_{\phi}^{\text{VGM}}.
\end{align}

\paragraph{(3) Reweighted Guidance Matching Loss $\ell^{\text{RGM}}$.} Eq. Replacing $\ell^{\text{GM}}$ in \eqref{eq:appendix:guidance_matching_loss_g_1} with $\ell^{\text{VGM}}$: 
\begin{equation}\label{eq:appendix:guidance_matching_loss_g_3_loss}
    \frac{e^{-J(x_1)}}{Z_{\phi_Z,sg}(x_t,t)} \|g_{\phi}(x_t,t) + v_\theta(x_t,t) - v_{t|z}(x_t|z)\|_2^2,
\end{equation}
and the loss $\mathcal{L}^{\text{GM}}_\phi$ becomes $\mathcal{L}^{\text{VGM}}_\phi$, which are shown equivalent in the following.

Starting from Eq. \eqref{eq:appendix:guidance_matching_loss_g1_expanded}, we can extract $\frac{e^{-J(x_1)}}{Z_{\phi_Z}}$ from the three terms depended on $\phi$, and thus showing the resulting loss is indeed $\ell^{\text{RGM}}$.
Notice that because $Z_{\phi_Z} = \mathbb{E}_{z\sim p(z|x_t)}[e^{-J(z)}]$,
\begin{equation}
    \mathbb{E}_{z\sim p(z|x_t)}[\frac{e^{-J(z)}}{Z_{\phi_Z}(x_t)}f(x_t,t)] = f(x_t,t).
\end{equation}
Thus, we have:
\begin{align}
    \nonumber \mathcal{L}^{\text{GM}}_\phi& = \mathbb{E}_{t\sim\mathcal{U}(0,1),z\sim p(z),x_t \sim p(x_t|z)}[\|\underbrace{g_{\phi}(x_t,t)\|_2^2}_{\text{dependent on }\phi} + \|\frac{e^{-J(x_1)}}{Z_{\phi_Z,sg}(x_t,t)}v_{t|z}(x_t|z)\|_2^2 + \|v_{t|z}(x_t|z)\|_2^2 \\
    \nonumber& \underbrace{-2\langle g_{\phi}(x_t,t), \frac{e^{-J(x_1)}}{Z_{\phi_Z,sg}(x_t,t)}v_{t|z}(x_t|z)\rangle}_{\text{dependent on }\phi} - 2\langle\frac{e^{-J(x_1)}}{Z_{\phi_Z,sg}(x_t,t)}v_{t|z}(x_t|z), v_{t|z}(x_t|z)\rangle \underbrace{- 2\langle v_{t|z}(x_t|z), g_{\phi}(x_t,t)\rangle}_{\text{dependent on }\phi} ] \\
   \nonumber & =\mathbb{E}_{t\sim\mathcal{U}(0,1),z\sim p(z),x_t \sim p(x_t|z)}[\|\frac{e^{-J(x_1)}}{Z_{\phi_Z}(x_t)}{g_{\phi}(x_t,t)\|_2^2} + \|\frac{e^{-J(x_1)}}{Z_{\phi_Z,sg}(x_t,t)}v_{t|z}(x_t|z)\|_2^2 + \|v_{t|z}(x_t|z)\|_2^2 \\
    \nonumber& {-2\langle g_{\phi}(x_t,t), \frac{e^{-J(x_1)}}{Z_{\phi_Z,sg}(x_t,t)}v_{t|z}(x_t|z)\rangle} - 2\langle\frac{e^{-J(x_1)}}{Z_{\phi_Z,sg}(x_t,t)}v_{t|z}(x_t|z), v_{t|z}(x_t|z)\rangle - 2\frac{e^{-J(x_1)}}{Z_{\phi_Z}(x_t)}\langle v_{t|z}(x_t|z), g_{\phi}(x_t,t)\rangle ] \\
    & = \mathbb{E}_{t\sim\mathcal{U}(0,1),z\sim p(z),x_t \sim p(x_t|z)} \left[\underbrace{\frac{e^{-J(x_1)}}{Z_{\phi_Z,sg}(x_t,t)} \|g_{\phi}(x_t,t) + v_\theta(x_t,t) - v_{t|z}(x_t|z)\|_2^2}_{=\ell^{\text{RGM}}}\right]\\
    &=\mathcal{L}_\phi^{\text{RGM}}.
\end{align}
Where we used the conclusion of Eq. \eqref{eq:appendix:guidance_matching_loss_g_2_v_term}, and inserted terms that vanish after $\nabla_{\phi}$ to make $\ell^{\text{RGM}}$.

\paragraph{(3) Marginalized Reweighted Guidance Matching Loss $\ell^{\text{MRGM}}$.} The above loss Eq. \eqref{eq:appendix:guidance_matching_loss_g_3_loss} can be re-assigned a weight, which will result in the same optimal $g_{\phi}(x_t,t)$ as in Eq. \eqref{eq:guidance_matching_loss_g_1}. Specifically, by changing $\frac{1}{Z_{\phi_Z,sg}(x_t,t)}$ to $\frac{1}{\mathbb{E}_{x_t\sim p(x_t)}[Z_{\phi_Z,sg}(x_t,t)]}=\frac{1}{\int p(x_t)Z_{\phi_Z,sg}(x_t,t) dx_t}$, we have 
\begin{equation}\label{eq:appendix:guidance_matching_loss_g_4_loss_appendix}
    \frac{e^{-J(x_1)}}{\int e^{-J(x_1)} p(z)dz} \|g_{\phi}(x_t,t) + v_\theta(x_t,t) - v_{t|z}(x_t|z)\|_2^2.
\end{equation}
We only need to prove that
\begin{equation}
    \int p(x_t)Z_{t}(x_t) dx_t = \int e^{-J(x_1)} p(z)dz.
\end{equation}
Recall  that 
\begin{equation}
    Z_{t}(x_t) = \int p(z|x_t) e^{-J(x_1)} dz,
\end{equation}
so 
\begin{align}
    &\int p(x_t)Z_{t}(x_t) dx_t
    = \int  p(x_t)\int p(z|x_t) e^{-J(x_1)} dx_t dz \\
    =& \int p(z)e^{-J(x_1)}dz = Z.
\end{align}

Eq. \eqref{eq:guidance_matching_loss_g_4_loss} can also be derived by applying importance sampling $\mathbb{E}_{z\sim \frac{1}{Z}p(z)e^{-J(x_1)}}\left[\|\cdot\|_2^2 \right]=\mathbb{E}_{z\sim p(z)}\left[ \frac{e^{-J(x_1)}}{Z} \|\cdot\|_2^2  \right]$ to the flow matching objective of the new VF for the new target distribution $p'(x_t) = \frac{1}{Z} p(x_t) e^{-J(x_t)}$.

\paragraph{Discussions}
The losses have the same expected gradient, but their performance may differ. Among the four losses, $\ell_\phi^{\text{MRGM}}$ is the only one that does not require the auxiliary model $Z_{\phi_Z}$. However, $\ell_\phi^{\text{RGM}}$ assigns loss weight dependent on $x_t$. The weight is emphasized when the expectation of $e^{-J(x_1)}$ under $p(x_1|x_t)$ is small. Compared to these two losses, $\ell_\phi^{\text{GM}},\ell_\phi^{\text{VGM}}$ do not reweight the loss. The variance of $\ell_\phi^{\text{GM}}$ will be smaller if $J$ is smooth, while $\ell_\phi^{\text{VGM}}$ is better when $v_t$ is more complex.

\subsection{Algorithm Details and Variants of $g^{\text{MC}}$}\label{app:independent_mc}
The pseudocode for computing $g^{\text{MC}}$ is as follows.
\begin{algorithm}[h]
\caption{Monte Carlo estimation of the guidance $g_t(x_t)$}
\label{alg:mc_estimation_on_g}
\begin{algorithmic}[1]
\REQUIRE Current $t$, $x_t$, known $p_t(x_t|z)$.

\STATE Sample $z_i \sim p(z)$, where $i=1,2,...,N$ 
{\color{gray} // Recall $z_i = (x_{1}^i,x_{0}^i)$}
\STATE $\tilde{p}_t(x_t) \gets \frac{1}{N} \sum_i p_t(x_t|z_i)$
\STATE $\tilde{Z}_t(x_t) \gets \frac{1}{N} \sum_i e^{-J(x_{1}^i)} \frac{p_t(x_t|z_i)}{\tilde{p}(x_t)}$
\STATE ${g}^{\text{MC}}_t(x_t) \gets \frac{1}{N} \sum_i (\frac{e^{-J(x_{1}^i)}}{\tilde{Z}_t(x_t)} - 1) v_{t|z}(x_t|z_i) \frac{p_t(x_t|z_i)}{\tilde{p}_t(x_t)}$
\STATE \textbf{return} ${g}^{\text{MC}}_t(x_t)$
\end{algorithmic}
\end{algorithm}

\paragraph{Independent Couplings.}
Although we introduced the flow matching using the condition $z=(x_0,x_1)$, it can also be chosen as $x_0$ or $x_1$ \citep{lipman_flow_2024}. When we $z\coloneqq x_1$, Algorithm \ref{alg:mc_estimation_on_g} can be readily adopted for the $x_0$ condition. This way, the Monte Carlo estimation reduces the integration region dimensionality to half of the original one, thus becoming more efficient. 

In the case where $z=(x_0,x_1)$ and the data coupling is independent $\pi(x_0|x_1) = p(x_0)$, we show here that the MC estimation can be simplified to the $x_1$-conditioned case that is more efficient:
\begin{align}
&    \bm{g_t}^{\textbf{MC-$x_1$}}(x_t) \defg \mathbb{E}_{x_1\sim p(x_1)} 
\left[
(\frac{e^{-J(x_1)}}{Z_t} - 1) v_{t|x_1}(x_t|x_1) \frac{p_t(x_t|x_1)}{p_t(x_t)}
\right],
\\
& Z_t^{\text{MC-$x_1$}}(x_t) = \mathbb{E}_{x_1\sim p(x_1)} 
\left[
e^{-J(x_1)} \frac{p_t(x_t|x_1)}{p_t(x_t)} 
\right].
\end{align}
Obviously, as $Z_t = \int p(x_0,x_1|x_t) e^{-J(x_1)}dx_0dx_1 = \int p(x_0|x_1,x_t)p(x_1|x_t) e^{-J(x_1)}dx_0dx_1 $, integrating out $x_0$ gives $Z_t = Z^{\text{MC-$x_1$}}_t$. Therefore, to prove the above simplification, we only need to prove that:
\begin{equation}
    \mathbb{E}_{x_0,x_1\sim p(x_0,x_1|x_t)}\left[(\frac{e^{-J(x_1)}}{Z_t} - 1)  v(x_t|x_0,x_1)\right] = \mathbb{E}_{x_1\sim p(x_1|x_t)}\left[(\frac{e^{-J(x_1)}}{Z_t} - 1)  v(x_t|x_1)\right].
\end{equation}
The proof is simply integrating out $x_0$:
\begin{align}
    \nonumber&\int p(x_0,x_1|x_t) (\frac{e^{-J(x_1)}}{Z_t} - 1)  v(x_t|x_0,x_1) dx_0 dx_1 \\
    \nonumber=& \int \underbrace{\int p(x_0|x_1,x_t)v(x_t|x_0,x_1)dx_0}_{\coloneqq v(x_t|x_1)} (\frac{e^{-J(x_1)}}{Z_t} - 1) p(x_1|x_t) dx_1 \\
    =& \int v(x_t|x_1) (\frac{e^{-J(x_1)}}{Z_t} - 1) p(x_1|x_t) dx_1.
\end{align}
It should be noted that $v(x_t|x_1)$ is defined to be generally different from $v(x_t|x_0,x_1)$, and to do MC estimation via importance sampling, we need the forward probability path $p(x_t|x_1)$ to have a known density. The variance-reducing variant of $g^{\text{MC}}$ is summarized in Algorithm \ref{alg:mc_uncoupled}.

\begin{algorithm}[H]
\caption{Monte Carlo estimation of the guidance $g_t(x_t)$}
\label{alg:mc_uncoupled}
\begin{algorithmic}[1]
\REQUIRE Current $t$, $x_t$, known $p_t(x_t|x_1)$.

\STATE Sample $x_1^i \sim p(x_1)$, where $i=1,2,...,N$
\STATE $\tilde{p}_t(x_t) \gets \frac{1}{N} \sum_i p_t(x_t|x_1^i)$
\STATE $\tilde{Z}_t(x_t) \gets \frac{1}{N} \sum_i e^{-J(x_{1}^i)} \frac{p_t(x_t|x_1^i)}{\tilde{p}(x_t)}$
\STATE $\tilde{g}_t(x_t) \gets \frac{1}{N} \sum_i (\frac{e^{-J(x_{1}^i)}}{\tilde{Z}_t(x_t)} - 1) v_{t|z}(x_t|x_1^i) \frac{p_t(x_t|x_1^i)}{\tilde{p}_t(x_t)}$
\STATE \textbf{return} $\tilde{g}_t(x_t)$
\end{algorithmic}
\end{algorithm}

\subsection{Localized Approximation}
\label{app:localized_approximation}

To get $g^{\text{local}}$, we presume $p(z|x_t)$ is localized, and we can use a point estimation to approximate $Z_t$:
\begin{align}
Z_t(x_t) = \int p(z|x_t) e^{-J(x_1)} dz  
\approx e^{-J(\hat{x}_1)}
\end{align}
where $\hat{x}_1 \coloneqq \mathbb{E}_{x_0,x_1\sim p(z|x_t)}[x_1]$,
and then expanding $g_t$ to the first order
\begin{align}\nonumber
    g_t(x_t) \approx g_t(x_t)^{\text{local}} &= \mathbb{E}_{z \sim p(z|x_t)} 
    \left[
    (\frac{e^{-J(x_1)}}{e^{-J(\hat{x}_1)}} - 1)v_{t|z}(x_t|z)
    \right] \\
    \nonumber &\approx \mathbb{E}_{z \sim p(z|x_t)} \left[
    (\frac{e^{-J(\hat{x}_1)}(1 - \nabla_{\hat{x}_1} J(\hat{x}_1) (x_1 - \hat{x}_1))}{e^{-J(\hat{x}_1)}} - 1)v_{t|z}(x_t|z)
    \right] \\
    &= -\mathbb{E}_{x_1 \sim p(x_1|x_t)}
    \left[
     (x_1 - \hat{x}_1)v_{t|z}(x_t|z)
    \right] \nabla_{\hat{x}_1} J(\hat{x}_1).
\end{align}

To quantify the approximation error, we have
\begin{align}
    \nonumber \|\delta g\|^2&\coloneqq \|g_t - g^{\text{local}}\|^2_2 \\
    \nonumber&=\bigg{\|}\mathbb{E}_{z \sim p(z|x_t)} 
    \left[
    (\frac{e^{-J(x_1)}}{Z_t(x_t)} - 1)v_{t|z}(x_t|z)
    \right] \\
    \nonumber&- \mathbb{E}_{z \sim p(z|x_t)} 
    \left[
    (\frac{e^{-J(\hat{x}_1)}(1 - \nabla_{\hat{x}_1} J(\hat{x}_1) (x_1 - \hat{x}_1))}{e^{-J(\hat{x}_1)}} - 1)v_{t|z}(x_t|z)
    \right] \bigg{\|}_2^2\\
    &=\left\|\mathbb{E}_{z \sim p(z|x_t)} 
    \left[
    (\frac{e^{-J(x_1)}}{Z_t(x_t)}-\frac{e^{-J(\hat{x}_1)}(1 - \nabla_{\hat{x}_1} J(\hat{x}_1) (x_1 - \hat{x}_1))}{e^{-J(\hat{x}_1)}})v_{t|z}(x_t|z)
    \right] \right\|_2^2
\end{align}
where 
\begin{equation}
    Z_t(x_t) = \mathbb{E}_{z\sim p(z|x_t)} [e^{-J(x_1)}].
\end{equation}
We start by computing the error bound of approximating $Z_t$ with $e^{-J(\hat{x}_1)}$. Using Taylor expansion and the Taylor Remainder Theorem \footnote{The notations here neglect the order of vector/matrix products, but this does not matter as all of them will be scaled using the triangle inequality.}, 
\begin{align}\nonumber
    \left\|Z_t(x_t)-e^{-J(\hat{x}_1)}\right\|^2_2 &= \left\|\mathbb{E}_{z\sim p(z|x_t)}[\sum_{k=2} \frac{1}{k!}D_{x_1}^k e^{-J(x)}\big{|}_{x_1=\hat{x}_1} (x_1 - \hat{x}_1)^k] \right\|_2^2 \\
    &\le \mathbb{E}_{z\sim p(z|x_t)}
    \left
    [\left\|\frac{1}{2} (x_1 - \hat{x}_1)^T \underbrace{\nabla_{x_1} \nabla_{x_1} e^{-J(x)}\big{|}_{x_1=\hat{x}_1 + t (x_1 - \hat{x}_1)}}_{\coloneqq h^{(J)}_t} (x_1 - \hat{x}_1)
    \right\|_2^2
    \right] ,
\end{align}
where $t\in[0, 1]$.

If we set the L2 norm of the covariance matrix $\mathbb{E}_{z\sim p(z|x_t)}[ (x_1 - \hat{x}_1)(x_1 - \hat{x}_1)^T]$ as $\sigma_1$, and the eigenvalue with the largest absolute value of $\max_{t,x_1}|h^{(J)}_t|$ to be $\lambda_{h}$, we can show that  
\begin{align}
    \nonumber \left\|Z_t(x_t)-e^{-J(\hat{x}_1)}\right\|_2 \le &\mathbb{E}_{z\sim p(z|x_t)}[(x_1 - \hat{x}_1)^T h^{(J)}_t(x_1 - \hat{x}_1)]\\
    \nonumber \le &\mathbb{E}_{z\sim p(z|x_t)}[(x_1 - \hat{x}_1)^T \lambda_h(x_1 - \hat{x}_1)]\\
    \nonumber \le &\lambda_h\mathbb{E}_{z\sim p(z|x_t)}[ (x_1 - \hat{x}_1)^T(x_1 - \hat{x}_1)]\\
    \nonumber \le& \lambda_h \mathbf{tr}\Sigma_{11} \\
    \le&\lambda_h \sigma_1 d,
\end{align}
where $\Sigma_{11}$ is the covariance matrix of $p(x_1|x_t)$, $d$ is the dimensionality of $x \in \mathbb{R}^d$. The last inequality follows from $\mathbf{tr}A = \sum_i^n \lambda_i \le n \max_i\lambda_i $, and the L2 norm of a matrix is its largest singular value, \emph{i.e.}, for the covariance matrix, that is the largest eigenvalue.

Then, 
\begin{align}
    \nonumber \delta g &= \mathbb{E}_{z \sim p(z|x_t)} 
    \left[
    \left(\frac{e^{-J(x_1)}}{Z_t(x_t)}-\frac{e^{-J(\hat{x}_1)}(1 - \nabla_{\hat{x}_1} J(\hat{x}_1) (x_1 - \hat{x}_1))}{e^{-J(\hat{x}_1)}}\right)v_{t|z}
    \right] \\
    & = \mathbb{E}_{z \sim p(z|x_t)} \left[ 
    \left(\underbrace{\frac{e^{-J({x}_1)}}{Z_t(x_t)} - \frac{e^{-J(x_1)}}{e^{-J(\hat{x}_1)}}}_{\text{Using the error bound of } Z_t} + \frac{e^{-J(x_1)}}{e^{-J(\hat{x}_1)}} -\frac{e^{-J(\hat{x}_1)}(1 - \nabla_{\hat{x}_1} J(\hat{x}_1) (x_1 - \hat{x}_1))}{e^{-J(\hat{x}_1)}}\right)v_{t|z}
    \right].
\end{align}
Therefore,
\begin{align}
    \nonumber \|\delta g\|_2^2& \le  \left\|-\frac{\lambda_h \sigma_1 d}{Z_t(x_t)e^{-J(\hat{x}_1)}} {\mathbb{E}_{z \sim p(z|x_t)} [e^{-J(x_1)}v_{t|z}]}\right\|_2^2 \\
    \nonumber &+ \left\|\mathbb{E}_{z \sim p(z|x_t)} \left[ 
    \frac{e^{-J(x_1)} - e^{-J(\hat{x}_1)}(1 - \nabla_{\hat{x}_1} J(\hat{x}_1) (x_1 - \hat{x}_1))}{e^{-J(\hat{x}_1)}} v_{t|z}
    \right] \right\|_2^2\\
    \nonumber & =  \bigg{\|} -\frac{\lambda_h \sigma_1 d}{Z_t(x_t)e^{-J(\hat{x}_1)}} {\mathbb{E}_{z \sim p(z|x_t)} [e^{-J(x_1)}v_{t|z}]}\bigg{\|}_2^2 \\
    & + \bigg{\|}\mathbb{E}_{z \sim p(z|x_t)} \left[ 
    \frac{e^{-J(\hat{x}_1)}(1 - \nabla_{\hat{x}_1} J(\hat{x}_1) (x_1 - \hat{x}_1)) + R_2 - e^{-J(\hat{x}_1)}(1 - \nabla_{\hat{x}_1} J(\hat{x}_1) (x_1 - \hat{x}_1))}{e^{-J(\hat{x}_1)}} v_{t|z}
    \right]\bigg{\|}_2^2.
\end{align}
By using the Taylor Remainder Theorem again, we have 
\begin{equation}
R_2 = \frac{1}{2}(x_1 - \hat{x}_1)^T \underbrace{\nabla_{\xi}\nabla_{\xi} e^{-J(\xi)}\big{|}_{\xi=\hat{x}_1 + t (x_1 - \hat{x}_1)}}_{= h^{(J)}_t}(x_1 - \hat{x}_1).
\end{equation}
Thus,
\begin{align}
    \nonumber \|\delta g\|_2^2\le&\left\|\lambda_h\sigma_1 d  \frac{\mathbb{E}[e^{-J(x_1)}v(x_t|z)]}{\mathbb{E}[e^{-J(x_1)}]e^{-J(\hat{x}_1)}} \right\|_2^2 + \bigg{\|}\mathbb{E}_{z \sim p(z|x_t)}\left[\frac{1}{2e^{-J(\hat{x}_1)}}(x_1 - \hat{x}_1)^T h_t^{(J)}(x_1 - \hat{x}_1) 
    v_{t|z}
    \right]\bigg{\|}_2^2\\
    \nonumber \le&\left\| \lambda_h\sigma_1 d \frac{\mathbb{E}[e^{-J(x_1)}v(x_t|z)]}{\mathbb{E}[e^{-J(x_1)}]e^{-J(\hat{x}_1)}} \right\|_2^2 \\
    \nonumber +&
        \left[
            \mathbb{E}_{z \sim p(z|x_t)} 
            \left\|
            \frac{1}{2e^{-J(\hat{x}_1)}}(x_1 - \hat{x}_1)^T h_t^{(J)}(x_1 - \hat{x}_1)
            \right\|_2^2
        \right]
        \left[
            \mathbb{E}_{z \sim p(z|x_t)} 
            \left\|
            v(x_t|z)
            \right\|_2^2
        \right]\\
    \le &  \left\| \lambda_h\sigma_1 d \frac{\mathbb{E}[e^{-J(x_1)}v(x_t|z)]}{\mathbb{E}[e^{-J(x_1)}]e^{-J(\hat{x}_1)}} \right\|_2^2 + \left\|\frac{\lambda_h \sigma_1 d}{2e^{-J(\hat{x}_1)}} \right\|^2_2 \mathbb{E}\left[\|v(x_t|z)\|_2^2\right].
\end{align}

Then we have
\begin{align}
    \nonumber \|\delta g\|^2_2 &\le  \left\| \lambda_h\sigma_1 d \frac{\mathbb{E}[e^{-J(x_1)}v(x_t|z)]}{\mathbb{E}[e^{-J(x_1)}]e^{-J(\mathbb{E}[x_1])}} \right\|_2^2
    + \left\|\frac{\lambda_h \sigma_1 d}{2e^{-J(\mathbb{E}[x_1])}} \right\|^2_2 \mathbb{E}\left[\|v(x_t|z)\|_2^2\right] \\
    & = \left|\frac{\lambda_h \sigma_1 d}{e^{-J(\mathbb{E}[x_1])}}\right|^2
    \left( 
    \underbrace{\left\|  \frac{\mathbb{E}[e^{-J(x_1)}v(x_t|z)]}{\mathbb{E}[e^{-J(x_1)}]} \right\|_2^2}_{C_1} + \underbrace{\frac{1}{4}\mathbb{E}\left[\|v(x_t|z)\|_2^2\right]}_{C_2}
    \right),
\end{align}
where we omit $z\sim p(z|x_t)$ in $\mathbb{E}_{z\sim p(z|x_t)}[\cdot]$ and simplify the notation to $\mathbb{E}[\cdot]$.
Therefore, the approximation error of $g^{\text{local}}$ is bounded by $(\lambda_h \sigma_1 d)^2 (C_1 + C_2)/|e^{-J(\mathbb{E}[x_1])}|$, where $\lambda_h$ is the largest eigenvalue of $h_t^{(J)}$, the Hessian matrix of the objective function $e^{-J}$, $\sigma_1$ is the L2 norm of the covariance matrix, $d$ is the sample dimensionality, $C_1$ is a constant that has to do with the norm of the new VF, and $C_2$ is the variance of the original \emph{conditional} VF. Some intuitions can be emphasized:
\begin{enumerate}
    \item The error is small when $J$ is smooth, in which case the Hessian of $e^{-J}$ will approach zero. This corresponds to the mild guidance, where approximation-based $g^{\text{local}}$ works well.
    \item The error is small when $\sigma_1$ is small, \emph{i.e.} the covariance matrix $\Sigma_{11}$ has a small Frobenius norm. This is the case when the flow time $t\rightarrow 1$ (and $\sigma_t = 0$), where $x_t$ predicts $x_1$ well.
    \item The magnitude of $|e^{-J(\mathbb{E}[x_1])}|$ reflects how well how well the \emph{estimated generated sample} $\mathbb{E}[{x}_1]$ matches the objective $J$ given the current $x_t$. If $\mathbb{E}[{x}_1]$ lies inside the region where $J$ is small, \emph{i.e.}, $\mathbb{E}[{x}_1]$ is a good sample, then the approximate guidance will be more accurate as the optimization is conducted locally, and gradient can reflect the landscape well. If $J(\mathbb{E}[x_1])$ is large, the gradient is almost randomly exploring the sample space, producing a larger approximation error.
    \item The cases where $C_1$ and $C_2$ are small are not necessarily those where the guidance is of better accuracy. Because of the small norm of the VF, the error in the guidance VF will likely cause a larger deviation due to increased relative error.
\end{enumerate}

\subsection{Estimation of $\hat{x}_1$}
\label{appendix:x1_parameterization}
Under the \emph{affine path assumption} (Assumption \ref{assumption:affine_path}), we can estimate the expectation of $x_1$ under the distribution $p(z|x_t)$. This is a well-known trick \citep{lipman_flow_2024,pokle_training-free_2024}, but our analysis includes the dependent coupling case.

Since the flow matching model learns
\begin{equation}
    v_{\theta}(x_t,t)\approx v_t(x_t) = \mathbb{E}_{z\sim p(z|x_t)}\left[
    v(x_t|z)
    \right],
\end{equation}
using the \emph{affine path assumption} ($x_t = \alpha_t x_1 + \beta_t x_0 + \sigma_t\dot\sigma_t\varepsilon$), 
\begin{equation}
    v(x_t|z) = \frac{d}{dt}x_t  = (\dot\alpha x_1 + \dot\beta_t x_0 + \dot\sigma_t\varepsilon),
\end{equation}
so
\begin{equation}\label{eq:appendix:x1_parameterization_v}
    v_t(x_t) = \mathbb{E}_{z\sim p(z|x_t)}\left[
    \dot\alpha x_1 + \dot\beta_t x_0 +\dot \sigma_t\varepsilon
    \right].
\end{equation}

Meanwhile, taking the expectation of $x_t$ under $p(z|x_t)$ yields
\begin{align}\label{eq:appendix:x1_parameterization_x}
    \underbrace{\mathbb{E}_{z\sim p(z|x_t)}[x_t] = x_t}_{\text{because} \int z p(z|x_t) dz = 1} = \mathbb{E}_{z\sim p(z|x_t)}[\alpha_t x_1 + \beta_t x_0 + \sigma_t\varepsilon].
\end{align}

Then, by using Eq. \ref{eq:appendix:x1_parameterization_v} and \ref{eq:appendix:x1_parameterization_x}, we can eliminate either $\hat{x}_0$ or $\hat{x}_1$ in each other's expression:
\begin{align}
    \hat{x}_0 \coloneqq \mathbb{E}_{z\sim p(z|x_t)}[x_0] & = 
    \frac{\dot\alpha_t x_t - \alpha_t v_t(x_t)}{\beta_t\dot\alpha_t - \dot\beta_t\alpha_t}
    +
    \underbrace{\mathbb{E}_{z\sim p(z|x_t)}\left[
    \frac{-\dot\alpha_t\sigma_t\varepsilon + \alpha_t\dot\sigma_t\varepsilon}{\beta_t\dot\alpha_t - \dot\beta_t\alpha_t}
    \right]}_{\coloneqq \zeta^0_t}
    \\
    \hat{x}_1 \coloneqq \mathbb{E}_{z\sim p(z|x_t)}[x_1] & = 
    \frac{-\dot\beta_t x_t + \beta_t v_t(x_t)}{\beta_t\dot\alpha_t - \dot\beta_t\alpha_t} + \underbrace{\mathbb{E}_{z\sim p(z|x_t)}\left[
    \frac{\dot\beta_t\sigma_t\varepsilon - \beta_t\dot\sigma_t\varepsilon}{\beta_t\dot\alpha_t - \dot\beta_t\alpha_t}
    \right]}_{\coloneqq \zeta^1_t}.
\end{align}
It should be noted that we have assumed that $\sigma_t$ is small and thus $\zeta_t^0$ and $\zeta_t^1$ are also small in the \emph{affine path assumption} (Assumption \ref{assumption:affine_path}):
\begin{align}
    \nonumber \zeta_t^0 &= \frac{-\dot\alpha_t\sigma_t + \alpha_t\dot\sigma_t}{\beta_t\dot\alpha_t - \dot\beta_t\alpha_t}\mathbb{E}_{z\sim p(z|x_t)}[\varepsilon] \\
    \nonumber &=\frac{-\dot\alpha_t\sigma_t + \alpha_t\dot\sigma_t}{\beta_t\dot\alpha_t - \dot\beta_t\alpha_t}
    \int \frac{p(x_t|x_0,x_1)\pi(x_0,x_1)}{p(x_t)}\varepsilon dx_0dx_1 \\
    &=\frac{-\dot\alpha_t\sigma_t + \alpha_t\dot\sigma_t}{\beta_t\dot\alpha_t - \dot\beta_t\alpha_t}\int\frac{1}{p(x_t)}
    \mathbb{E}_{\varepsilon\sim p_{\varepsilon}(\varepsilon)} \left[
    \pi(x_0(x_t,x_1,\varepsilon),x_1)\varepsilon \right] dx_1.
\end{align}
Since $\pi(x_0(x_t,x_1,\varepsilon),x_1)$ is a probability distribution that is assumed to be bounded, we denote $\max_{\varepsilon}\|\pi(x_0(x_t,x_1,\varepsilon),x_1)\|\le \mathcal{M}(x_1,x_t)$, and thus 
\begin{equation}
    \lim_{\sigma_t\rightarrow 0, \dot\sigma_t\rightarrow 0}\zeta_t^0
    \le\lim_{\sigma_t\rightarrow 0, \dot\sigma_t\rightarrow 0}\left|\frac{-\dot\alpha_t\sigma_t + \alpha_t\dot\sigma_t}{\beta_t\dot\alpha_t - \dot\beta_t\alpha_t}\right|\cdot \int 
    \left\|
    \frac{1}{p(x_t)}
    \right\|_2^2
    \cdot
    \left\|
    \mathbb{E}_{\varepsilon\sim p_{\varepsilon}(\varepsilon)}
    \|\varepsilon\|_2
    \mathcal{M}(x_1,x_t)
    \right\|_2^2dx_1 = 0.
\end{equation}
Since everything in the integral is independent of $\varepsilon$, $x_0$, or $\sigma_t$, as $\sigma_t\rightarrow 0$ $\zeta^0$ simply converges to zero. A similar approach can prove that $\lim_{\sigma_t\rightarrow 0, \dot\sigma_t\rightarrow 0}\zeta_2^1$ is also zero.

Next, we explain why we specifically care about the case where the small $\sigma_t$ assumption holds.
In independent coupling flow matching, $\sigma_t\varepsilon$ is exactly zero since we can use two of $x_t,x_0$, and $x_1$ to express the third one. In dependent coupling flow matching, this assumption also holds for famous methods such as optimal transport conditional flow matching or Schrodinger Bridge conditional flow matching \cite{tong_improving_2024}, where $\varepsilon\sim\mathcal{N}(0,I)$ and $\sigma_t$ is set as a small constant. Therefore, the assumption that $\sigma_t\varepsilon$ is small in \emph{affine path assumption} is general and applies to many existing flow matching methods. Hence, by approximating $\zeta_t^0$ and $\zeta_t^1$ as zero, we have the final estimation of $\hat{x}_1$
\begin{align}
    &\hat{x}_0 \approx 
    \frac{\dot\alpha_t x_t - \alpha_t v_\theta(x_t,t)}{\beta_t\dot\alpha_t - \dot\beta_t\alpha_t} \\
    &\label{eq:appendix:local_approx_guidance_affine_path_estimate_x1}\hat{x}_1 \approx 
    \frac{-\dot\beta_t x_t + \beta_t v_t(x_t)}{\beta_t\dot\alpha_t - \dot\beta_t\alpha_t},
\end{align}
where Eq. \eqref{eq:appendix:local_approx_guidance_affine_path_estimate_x1} is just Eq. \eqref{eq:local_approx_guidance_affine_path_estimate_x1}. Note the approximations become exact under Assumption \ref{assumption:uncoupled_affine_gaussian_path}.

\subsection{Proof of $g^{\text{cov}}$}
\label{appendix:affine_path_cov_local_approx_guidance}
Here, we prove that under the affine path assumption (Assumption \ref{assumption:affine_path}), Eq. \eqref{eq:g_cov}
\begin{equation}\label{eq:restate_g_cov}
    g_t^{\text{local}}\approx g_t^{\text{cov}} = -\underbrace{\frac{\dot\alpha_t\beta_t - \dot\beta_t\alpha_t}{\beta_t}}_{\text{schedule}} \Sigma_{1|t} \nabla_{\hat{x}_1}J(\hat{x}_1),
\end{equation}
where 
\begin{equation}
    g_t^{\text{local}} = -\mathbb{E}_{z \sim p(z|x_t)}
    \left[
     (x_1 - \hat{x}_1)v_{1|t}(x_t|z)
    \right] \nabla_{\hat{x}_1} J(\hat{x}_1).
\end{equation}

Under the affine path $x_t = \alpha_t x_1 + \beta_t x_0 + \sigma_t \varepsilon$ the conditional vector field $v_{1|t}$ follows
\begin{equation}
    v_{1|t}(x_t) = \dot \alpha_t x_1 + \dot\beta_t x_0 + \dot\sigma_t \varepsilon.
\end{equation}
Plugging this into the definition of $g^{\text{local}}$ and we get
\begin{align}
    \nonumber g_t^{\text{local}} = & -\mathbb{E}_{z \sim p(z|x_t)}[\underbrace{(x_1 - \hat{x}_1)(\dot \alpha_t x_1 + \dot\beta_t x_0 + \dot\sigma_t\varepsilon}_{\text{substitute } x_0 \text{ with } x_1,~ \sigma_t\varepsilon, \text{ and } x_t})] \nabla_{\hat{x}_1} J(\hat{x}_1)\\
    \nonumber =& -\mathbb{E}_{z \sim p(z|x_t)}[(x_1 - \hat{x}_1)(\dot \alpha_t x_1 + \frac{\dot\beta_t}{\beta_t} (x_t - \alpha_t x_1 - \sigma_t \varepsilon) + \dot\sigma_t\varepsilon)] \nabla_{\hat{x}_1} J(\hat{x}_1)\\
    \nonumber =& -\mathbb{E}_{z \sim p(z|x_t)}\left[
    (x_1 - \hat{x}_1)\left(
    \left(\frac{\beta_t\dot \alpha_t - \alpha_t\dot\beta_t}{\beta_t}\right) x_1 +  \cancel{x_t}  + (\dot\sigma_t - \sigma_t)\varepsilon
    \right)
    \right] 
    \nabla_{\hat{x}_1} J(\hat{x}_1)\\
    = & -{\frac{\dot\alpha_t\beta_t - \dot\beta_t\alpha_t}{\beta_t}} \Sigma_{1|t} \nabla_{\hat{x}_1}J(\hat{x}_1)
    + \underbrace{(\sigma_t - \dot\sigma_t)\mathbb{E}_{z \sim p(z|x_t)}[(x_1-\hat{x}_1)\varepsilon]\nabla_{\hat{x}_1}J(\hat{x}_1)}_{\coloneqq\Upsilon,~\lim_{\sigma_t\rightarrow 0, \dot\sigma_t \rightarrow 0}\|\Upsilon\|_2^2=0 },
\end{align}
where the $x_t$ term is canceled out because $\int p(z|x_t) (x_1 - \mathbb{E}_{z\sim p(z|x_t)}[x_1])dz = \int p(z|x_t) x_1 dz -\mathbb{E}_{z\sim p(z|x_t)}[x_1] = 0$, $\Sigma_{1|t}\coloneqq \mathbb{E}_{z \sim p(z|x_t)}\left[
(x_1 - \hat{x}_1)(x_1 - \hat{x}_1)\right]$, and the residual term that characterizes the approximation error (denoted as $\|\Upsilon\|^2_2$) in Eq. \eqref{eq:g_cov} (restated in Eq. \eqref{eq:restate_g_cov}) is
\begin{align}
    \nonumber \Upsilon = &(\sigma_t - \dot\sigma_t)\mathbb{E}_{z \sim p(z|x_t)}[(x_1-\hat{x}_1)\varepsilon]\nabla_{\hat{x}_1}J(\hat{x}_1) \\
    \nonumber =&(\sigma_t - \dot\sigma_t)\nabla_{\hat{x}_1}J(\hat{x}_1)\int \frac{p(x_t|z)p(z)}{p(x_t)}(x_1-\hat{x}_1)\varepsilon dx_1dx_0 \\
    \nonumber =&(\sigma_t - \dot\sigma_t)\nabla_{\hat{x}_1}J(\hat{x}_1)\int \frac{p(\sigma_t\varepsilon)\pi(x_0|x_1)p(x_1)}{p(x_t)}(x_1-\hat{x}_1)\frac{1}{\sigma_t}(x_t - \alpha_t x_1 - \beta_t x_0) dx_1dx_0 \\
    \nonumber =&(\sigma_t - \dot\sigma_t)\nabla_{\hat{x}_1}J(\hat{x}_1)\int \frac{p(x_1)}{p(x_t)}(x_1-\hat{x}_1) dx_1 \int p(\sigma_t\varepsilon)\pi(x_0|x_1) \varepsilon dx_0 \\
    =&(\sigma_t - \dot\sigma_t)\nabla_{\hat{x}_1}J(\hat{x}_1)\int \frac{p(x_1)}{p(x_t)}(x_1-\hat{x}_1) 
    \mathbb{E}_{\varepsilon\sim p_\varepsilon(\varepsilon)}
    \left[{\pi\left(\frac{1}{\beta_t}(x_t - \alpha_t x_1 - \sigma_t \varepsilon)\mid x_1\right)} \varepsilon \right]dx_1 ,
\end{align}
where $p_\varepsilon(\varepsilon)$ is the marginal distribution of $\varepsilon$.
Suppose $\|\pi\left(\frac{1}{\beta_t}(x_t - \alpha_t x_1 - \sigma_t \varepsilon)\mid x_1\right)\|^2_2 = \|\pi\left(x_0 \mid x_1\right)\|^2_2\le \mathcal{M}(x_1,x_t)$ (which is a function independent of $\varepsilon$, then
\begin{align}
    \nonumber &\|\Upsilon\|^2_2 \\
    \nonumber \le& \left\|(\sigma_t - \dot\sigma_t)\nabla_{\hat{x}_1}J(\hat{x}_1)\right\|_2^2\cdot \left\|\int\frac{p(x_1)}{p(x_t)}(x_1-\hat{x}_1)\mathbb{E}_{\varepsilon\sim p_\varepsilon(\varepsilon)}
    \left[{\pi\left(\frac{1}{\beta_t}(x_t - \alpha_t x_1 - \sigma_t \varepsilon)\mid x_1\right)} \varepsilon \right]dx_1\right\|^2_2 \\
    \nonumber \le& |(\sigma_t - \dot\sigma_t)|\underbrace{\left\|\nabla_{\hat{x}_1}J(\hat{x}_1)\right\|_2^2}_{\coloneqq \mathcal{G}}\cdot \int
    \underbrace{\left\|\frac{p(x_1)}{p(x_t)}(x_1-\hat{x}_1)\right\|^2_2}_{\coloneqq\mathcal{Q}}
    \cdot
    \underbrace{\left\|\mathbb{E}_{\varepsilon\sim p_\varepsilon(\varepsilon)}
    \left[{\pi\left(\frac{1}{\beta_t}(x_t - \alpha_t x_1 - \sigma_t \varepsilon)\mid x_1\right)} \varepsilon \right]\right\|^2_2}_{\le \left(\mathcal{M}\mathbb{E}_{\varepsilon\sim p_\varepsilon(\varepsilon)}[\|\varepsilon\|_2]\right)^2 \le \mathcal{M}^2 \text{Var}_{p_\varepsilon}}dx_1 \\
    \le& |(\sigma_t - \dot\sigma_t)|\mathcal{G}(x_t)\int \mathcal{Q}(x_1,x_t))\mathcal{M}^2(x_1,x_t)\text{Var}_{p_\varepsilon} dx_1,
\end{align}
all of which are independent on $x_0$ or $\sigma_t$. Thus, 
\begin{equation}
    \lim_{\sigma\rightarrow 0,\sigma_t\rightarrow 0}\|\Upsilon\|^2_2 = 0.
\end{equation}

\subsection{The Jacobian Trick}\label{appendix:jacobian_trick}
We prove the Jacobian Trick here. 
\begin{proposition} The Jacobian trick. Under Assumption \ref{assumption:uncoupled_affine_gaussian_path}, the inverse covariance matrix of $p(x_1|x_t)$, $\Sigma_{1|t}$, is affine to the Jacobian of the VF $\frac{\partial v_t}{\partial x_t}$, and is proportional to the Jacobian $\frac{\partial \hat{x}_1}{\partial x_t}$:
\begin{align}\nonumber
     \Sigma_{1|t}= \frac{\beta_t^2}{\alpha_t(\dot\alpha_t\beta_t - \dot\beta_t\alpha_t)} (-\dot\beta_t+ \beta_t\frac{\partial v_t}{\partial x_t} )
     = \frac{\beta_t^2}{\alpha_t} \frac{\partial \hat{x}_1}{\partial x_t}.
\end{align}
\end{proposition}

\textbf{Proof.}

To begin with, we prove $\Sigma_{1|t}=\frac{\beta_t^2}{\alpha_t} \frac{\partial \hat{x}_1}{\partial x_t}$. A similar conclusion has been proved in \citet{ye_tfg_2024}. We generalize their proof to affine Gaussian path flow matching:

Recall from Eq. \eqref{eq:appendix:local_approx_guidance_affine_path_estimate_x1} that
\begin{equation}
    \hat{x}_1 = -\frac{\dot\beta_t}{\dot \alpha_t \beta_t - \dot \beta_t \alpha_t} x_t + \frac{\beta_t}{\dot \alpha_t \beta_t - \dot \beta_t \alpha_t} v_t
\end{equation}
and \begin{equation}
    \hat{x}_0 = \frac{\dot\alpha_t}{\dot \alpha_t \beta_t - \dot \beta_t \alpha_t} x_t - \frac{\alpha_t}{\dot \alpha_t \beta_t - \dot \beta_t \alpha_t} v_t.
\end{equation}
So we have the Jacobian trick
\begin{equation}
    \frac{\partial \hat{x}_1}{\partial x_t} = -\frac{\dot\beta_t}{\dot \alpha_t \beta_t - \dot \beta_t \alpha_t} + \frac{\beta_t}{\dot \alpha_t \beta_t - \dot \beta_t \alpha_t} \frac{\partial v_t(x_t)}{\partial x_t},
\end{equation}
and because the VF is associated with the score
\begin{equation}\label{eq:appendix:score_and_vector_field}
    v_t(x_t) = \frac{\beta_t(\dot\alpha_t\beta_t - \dot \beta_t\alpha_t)}{\alpha_t} \nabla_{x_t}\log p_t(x_t) + \frac{\dot\alpha_t}{\alpha_t}x_t.
\end{equation}

Next, we try to prove 
\begin{equation}\label{eq:appendix:score_derivative_and_covariance}
    \nabla_{x_t}\nabla_{x_t}\log p_t(x_t) = -\frac{1}{\beta_t^2} + \frac{\alpha_t^2}{\beta_t^4} \Sigma_{x_1x_1},
\end{equation}
which allows us to connect the derivative of the score $\nabla^2_{x_t}\log p(x_t)$\footnotemark with the covariance matrix $\Sigma_{1|t}$.
\footnotetext{We use $\nabla \nabla$ and $\nabla^2$ interchangeably, with a little abuse of notation. It should not cause confusion since the size of the terms in the equations must match.}

\begin{align}
    \nonumber \nabla_{x_t}^2\log p(x_t) = 
    \nonumber & \frac{\nabla_{x_t}^2 p(x_t) }{p(x_t)} - \nabla_{x_t}\log p(x_t) \nabla_{x_t}\log p(x_t)\\
    \nonumber =& \frac{1}{p(x_t)} \int p(x_1)
    \underbrace{\nabla_{x_t}^2 p(x_t|x_1)}_{\text{using }\nabla^2 p = p\nabla^2 \log p + p(\nabla \log p)^2}
    dx_1
    - \nabla_{x_t}\log p(x_t) \nabla_{x_t}\log p(x_t) \\
    \nonumber =& \frac{1}{p(x_t)} \int p(x_1) (p(x_t|x_1)\nabla_{x_t}^2\log p(x_t|x_1) + p(x_t|x_1)\nabla_{x_t}\log p(x_t|x_1)\nabla_{x_t}\log p(x_t|x_1)) dx_1\\
    \nonumber &
    - \nabla_{x_t}\log p(x_t) \nabla_{x_t}\log p(x_t)\\
    \nonumber =&\mathbb{E}_{x_1\sim p(x_1|x_t)}\left[
        \nabla_{x_t}^2\log p(x_t|x_1)+ \nabla_{x_t}\log p(x_t|x_1)\nabla_{x_t}\log p(x_t|x_1)
    \right]
    - \nabla_{x_t}\log p(x_t) \nabla_{x_t}\log p(x_t) \\
    \nonumber =&\mathbb{E}_{x_1\sim p(x_1|x_t)}\left[
        -\frac{1}{\beta_t^2}
        +\left(
            \frac{x_t - \alpha_t x_1}{\beta_t^2}
        \right)^2
    \right]
    -\left(
        \frac{x_t - \alpha_t \mathbb{E}_{x_1\sim p(x_1|x_t)}[x_1]}{\beta_t^2}
    \right)^2 \\
    \nonumber =&-\frac{1}{\beta_t^2} + \frac{\alpha_t^2}{\beta_t^4} 
    \left(\mathbb{E}[x_1 x_1^T] - \mathbb{E}[x_1]\mathbb{E}[x_1]^T\right) \\
    =& -\frac{1}{\beta_t^2} + \frac{\alpha_t^2}{\beta_t^4} \Sigma_{x_1x_1}.
\end{align}

Then by combining Eq. \eqref{eq:appendix:score_and_vector_field} and \eqref{eq:appendix:score_derivative_and_covariance} we have
\begin{align}
    \nonumber \frac{\partial \hat{x}_1}{\partial x_t} &=  -\frac{\dot\beta_t}{\dot \alpha_t \beta_t - \dot \beta_t \alpha_t} + \frac{\beta_t}{\dot \alpha_t \beta_t - \dot \beta_t \alpha_t} 
    \left(
    \frac{\beta_t(\dot\alpha_t\beta_t - \dot \beta_t\alpha_t)}{\alpha_t} \nabla_{x_t}\nabla_{x_t}\log p_t(x_t) + \frac{\dot\alpha_t}{\alpha_t}
    \right) \\
    \nonumber & = -\frac{\dot\beta_t}{\dot \alpha_t \beta_t - \dot \beta_t \alpha_t} + \frac{\beta_t}{\dot \alpha_t \beta_t - \dot \beta_t \alpha_t} 
    \left(
    \frac{\beta_t(\dot\alpha_t\beta_t - \dot \beta_t\alpha_t)}{\alpha_t} (-\frac{1}{\beta_t^2} + \frac{\alpha_t^2}{\beta_t^4} \Sigma_{x_1x_1}) +\frac{\dot\alpha_t}{\alpha_t}
    \right) \\
    &=\frac{\alpha_t}{\beta_t^2}\Sigma_{x_1x_1}.
\end{align}

Inserting back Eq. \eqref{eq:appendix:local_approx_guidance_affine_path_estimate_x1} and we prove 
\begin{equation}
    \Sigma_{1|t}= \frac{\beta_t^2}{\alpha_t(\dot\alpha_t\beta_t - \dot\beta_t\alpha_t)} (-\dot\beta_t+ \beta_t\frac{\partial v_t}{\partial x_t} ).
\end{equation}

\subsection{Proof for $g_t^{\text{sim-inv}}$}\label{appendix:approximate_simple_posterior_pigdm_like}
We begin with
\begin{align}
    &g^{\text{sim-inv}}_t(x_t) = \int \left(\frac{e^{-J(x_1)}}{\tilde{Z}_t} - 1\right) v_{t|z}(x_t|z) \tilde{p}(z|x_t) dz, \\
    &\text{where }\tilde{Z}_t = \int e^{-J(x_1)} \tilde{p}(z|x_t)dz,
\end{align}
and approximate $p(x_1|x_t)$ with $\mathcal{N}(x_1;\hat{x}_1, \Sigma_t)$ where $\hat{x}_1\coloneqq \mathbb{E}_{z\sim p(z|x_t)}[x_1]$ and $\Sigma_t$ is already known.

With Assumption \ref{assumption:affine_path} and assuming $e^{-J(x_1)} = \mathcal{N}(Hx_1;y, \sigma_y I)$, we have 
\begin{align}
    \tilde{Z}_t &= \int e^{-J(x_1)} \tilde{p}(z|x_t)dz \nonumber\\
    &=\int e^{-\frac{1}{2\sigma_y^2}\|y-Hx_1\|^2_2 - \frac{1}{2}(z-\hat{z})^T\Sigma_t^{-1}(z-\hat{z})} dz
\end{align}
where $\hat{z} = (\hat{x}_0, \hat{x}_1)$ is the expectation of $z$ under $p(z|x_t)$. Note $H$ operates on $x_1$ only, and we pad the blocks related to $x_0$ with zero in $H$.

Then, by inserting the Gaussian approximation
\begin{align}
    g^{\text{sim-inv}}_t(x_t) &\approx \int \left(\frac{e^{-J(x_1)}}{\tilde{Z}_t} - 1\right) (\dot\alpha_t x_1 + \dot \beta_t x_0) \underbrace{\tilde{p}(z|x_t)}_{\text{Gaussian}} dz \\
    & = \int \underbrace{\frac{1}{\tilde{Z}_t}\exp\left(-\frac{1}{2\sigma_y^2}\|y-Hx_1\|^2_2 - \frac{1}{2}(z-\hat{z})^T\Sigma_t^{-1}(z-\hat{z})\right)}_{\coloneqq \tilde{\tilde{p}}(z|x_t)} (\dot\alpha_t x_1 + \dot \beta_t x_0)  dz - v_t(x_t).\label{eq:appendix:gaussian_approx_affine_integral_of_g}
\end{align}

\begin{remark}
    Note that $\Sigma_t^{-1}$ couples $x_0$ and $x_1$. This is a fundamental feature of dependent couplings $\pi(x_0,x_1)$. However, it may seem tempting to further assume that $\Sigma_t^{-1}$ is diagonal or even a scalar. It should be noted that this assumption completely discards the dependency of $x_0$ and $x_1$ in the coupling, and thus, we try to avoid that in the dependent coupling case.
\end{remark}

For clarity, we need to express $\Sigma_t^{-1}$ with 
\begin{equation}
    \Sigma_t^{-1} \overset{\Delta}{=} \begin{pmatrix}
        \Xi_{00} & \Xi_{01} \\
        \Xi_{10} & \Xi_{11}
    \end{pmatrix}.
\end{equation}

Then, the distribution $\exp\left(-\frac{1}{2\sigma_y^2}\|y-Hx_1\|^2_2 - \frac{1}{2}(z-\hat{z})^T\Sigma_t^{-1}(z-\hat{z})\right)$ is still a Gaussian, and to estimate the expectation of $z=(x_0,x_1)$ we need to simply the probability density of this Guassian into a standard form.
\begin{align}
    \nonumber {\tilde{Z}_t}\tilde{\tilde{p}}(z|x_t) = &\exp\left(-\frac{1}{2\sigma_y^2}\|y-Hx_1\|^2_2 - \frac{1}{2}(z-\hat{z})^T\Sigma_t^{-1}(z-\hat{z})\right) \\
    \nonumber =&\exp\left(-\frac{1}{2\sigma_y^2}
    \left(
        \|y\|^2- 2\langle y, Hx_1 \rangle + \|Hx_1\|^2_2
    \right)
    -
    \frac{1}{2}z^T\Sigma_t^{-1}z - \frac{1}{2}\hat{z}^T{\Sigma_t^{-1}}\hat{z} + \underbrace{\langle z,\Sigma_t^{-1} \hat{z} \rangle}_{\text{since } \Sigma_t^{-1} = {\Sigma_t^{-1}}^T}\right) \\
    =&\exp\left(-\frac{1}{2}\left(
        z^T\left(\frac{H^TH}{\sigma_y^2} + \Sigma_t^{-1}\right)z
        -2\langle
            z, \frac{H^T}{\sigma_y^2}y + \Sigma_t^{-1}\hat{z}
        \rangle
        +\left(
            \frac{1}{2\sigma_y^2}\|y\|^2 + \frac{1}{2}\hat{z}^T{\Sigma_t^{-1}}\hat{z}
        \right)
        \right)
    \right),
\end{align}
It is obvious that the mean of this Gaussian is
\begin{equation}
    \mu = \Bigg{(}\underbrace{\frac{H^TH}{\sigma_y^2} + \Sigma_t^{-1}}_{\coloneqq P}\Bigg{)}^{-1}
    \left(
    \frac{H^T}{\sigma_y^2}y + \Sigma_t^{-1}\hat{z}
    \right),
\end{equation}
where we can find $P$'s blocks to be 
\begin{equation}
    P = \begin{pmatrix}
        \Xi_{00} & \Xi_{01} \\
        \Xi_{10} & \Xi_{11} + \frac{H^TH}{\sigma_y^2}
    \end{pmatrix}.
\end{equation}

Then, by computing $\mu = (\hat{\hat{x}}_0,\hat{\hat{x}}_1)$ we can compute $g_t^{\text{sim-inv}} + v_t$, where $\hat{\hat{x}}_0,\hat{\hat{x}}_1$ are the $x_0$ and $x_1$ term in the integral of Eq. \eqref{eq:appendix:gaussian_approx_affine_integral_of_g}, because
\begin{equation}
     g_t^{\text{sim-inv}} + v_t = \mathbb{E}_{z\sim\tilde{\tilde{p}}(z|x_t)} [\dot \alpha_t x_1 + \dot\beta x_0].
\end{equation}

To simplify, insert back $v_t$ to get
\begin{align}
     \nonumber g_t^{\text{sim-inv}} &= \mathbb{E}_{z\sim\tilde{\tilde{p}}(z|x_t)} [\dot \alpha_t x_1 + \dot\beta x_0] - \mathbb{E}_{z\sim \tilde{p}(z|x_t)}[\dot \alpha_t x_1 + \dot\beta x_0] \\
     \nonumber &=
     \begin{pmatrix}
     \dot\beta_t I & \dot\alpha_t I
     \end{pmatrix}
     P^{-1}
     \left(
    \begin{pmatrix}0 \\ \frac{H^T}{\sigma_y^2}y\end{pmatrix}
        +
     \left(
        \begin{pmatrix}
            \Xi_{00} & \Xi_{01} \\
            \Xi_{10} & \Xi_{11}
        \end{pmatrix}
        -
        P
    \right)
     \begin{pmatrix}\hat{x}_0 \\ \hat{x}_1 \end{pmatrix}
     \right)\\
     \nonumber &=
     \begin{pmatrix}
     \dot\beta_t I & \dot\alpha_t I
     \end{pmatrix}
     P^{-1}\left(
    \begin{pmatrix}0 \\ \frac{H^T}{\sigma_y^2}y\end{pmatrix}
        +
    \begin{pmatrix}
            0 & 0 \\
            0 & -\frac{H^TH}{\sigma_y^2}
    \end{pmatrix}
    \begin{pmatrix}\hat{x}_0 \\ \hat{x}_1 \end{pmatrix}
    \right) \\
    & = 
    \begin{pmatrix}
     \dot\beta_t I & \dot\alpha_t I
     \end{pmatrix}
     P^{-1}
    \begin{pmatrix}0 \\ \frac{H^T}{\sigma_y^2}y - \frac{H^TH}{\sigma_y^2} \hat{x}_1\end{pmatrix}.
\end{align}
Usually, $\begin{pmatrix}\dot\beta_t I & \dot\alpha_t I\end{pmatrix}P^{-1}$ is difficult to obtain:
\begin{equation}
    g_t^{\text{sim-inv}} = (\dot\beta_t P^{-1}_{01} + \dot\alpha_t P^{-1}_{11})\left(\frac{H^T}{\sigma_y^2}y - \frac{H^TH}{\sigma_y^2} \hat{x}_1\right), 
\end{equation}
where $P_{01}^{-1}$ and $P_{11}^{-1}$ requires computing the inversion of $P$ and thus in general intractable. 
Using block matrix inversion, we have
\begin{align}
    g_t^{\text{sim-inv}} = \Bigg{(}-\dot\beta_t \Xi_{11}^{-1}\Xi_{01} \Big{(}\Xi_{11} + \frac{H^TH}{\sigma_y^2}-\Xi_{10}\Xi_{11}^{-1}\Xi_{01}\Big{)}^{-1}
    +\dot\alpha_t \Big{(}\Xi_{11} + \frac{H^TH}{\sigma_y^2}-\Xi_{10}\Xi_{11}^{-1}\Xi_{01}\Big{)}^{-1}\Bigg{)}
    \left(\frac{H^T}{\sigma_y^2}y - \frac{H^TH}{\sigma_y^2} \hat{x}_1\right).
\end{align}

For general (possibly coupled) affine path flow matching, we can make approximations and set the \emph{blocks} in $\Sigma_t^{-1}$ to scalars. It should be noted that this Gaussian assumption can still capture some coupling between $x_0$ and $x_1$ since the off-diagonal blocks $\Xi_{01}$ and $\Xi_{10}$ are not set to zero.
Specifically, we have
\begin{equation}
    g_t^{\text{sim-inv-A}} = -\lambda_t \Big{(}\frac{\sigma_y^2}{r_t^2} + H^TH\Big{)}^{-1}
    H^T\left(y - {H} \hat{x}_1\right),
\end{equation}
where $\lambda_t$ and $r_t^2$ are hyperparameters. $\lambda_t$ approximates $\dot\alpha_t-\dot\beta_t \Xi_{11}^{-1}\Xi_{01}$, absorbing the flow schedule.

\textbf{Special case: the non-coupled affine Gaussian path}

Next, we prove that $g_t^{\text{sim-inv}}$ covers $\Pi$GDM \citep{song_pseudoinverse-guided_2022} and OT-ODE \citep{pokle_training-free_2024} as special cases.
Under the uncoupled affine Gaussian path assumption (Assumption \ref{assumption:uncoupled_affine_gaussian_path}), one may think that the covariance matrix is block diagonal, but it is false: $x_0$ and $x_1$ are still dependent on each other in the distribution $p(z|x_t) = p(x_0,x_1|x_t)$ even if the coupling is independent. In the uncoupled case, the probability graph is $x_0 \rightarrow x_t \leftarrow x_1$, so although $x_0$ and $x_1$ are marginally independent ($\pi(x_0,x_1)= p(x_0)p(x_1)$), their conditional can be dependent $p(x_0,x_1|x_t)\neq p(x_0|x_t)p(x_1|x_t)$.
Then, we notice the uncoupled path is 
\begin{equation}
    x_t = \alpha_t x_1 + \beta_t x_0,
\end{equation}
so we actually should not have approximated the distribution $p_{z|x_t}$ as a Gaussian in the uncoupled case. Fortunately, there is a workaround to make $x_0$ almost entirely dependent on $x_1$. We can set $x_0 = -\frac{\alpha_t}{\beta_t}x_1 + \frac{1}{\beta_t}x_t + \xi \epsilon$, where $\epsilon \sim \mathcal{N}(0, I)$, and setting $\xi \rightarrow 0$ gives our desired uncoupled path results. The covariance matrix of $x_0$ and $x_1$ to:
\begin{equation}
    \Sigma_t = \begin{pmatrix}
        \frac{\alpha_t^2}{\beta_t^2}\Sigma_{x_1 x_1} + \xi^2 I & -\frac{\alpha_t}{\beta_t} \Sigma_{x_1 x_1} \\
        -\frac{\alpha_t}{\beta_t} \Sigma_{x_1 x_1} & \Sigma_{x_1 x_1}
    \end{pmatrix}
\end{equation}
Note that $\Sigma_{x_1 x_1}^{-1} \neq \Xi_{11}$ as $\Xi_{11}$ is a block in the inversion of the larger matrix $\Sigma_t$. Next we compute $\Sigma_t^{-1}$:
\begin{align}
    \nonumber \Sigma_t^{-1} =& \begin{pmatrix}
    \frac{1}{\xi^2} I & -\frac{a}{\xi^2} I \\
    -\frac{a}{\xi^2} I  & \Big{(} \Sigma_{x_1 x_1} - a^2\Sigma_{x_1 x_1} (a^2\Sigma_{x_1 x_1} +\xi^2 I)^{-1} \Sigma_{x_1 x_1} \Big{)}^{-1}
    \end{pmatrix} \\
    = &\begin{pmatrix}
    \frac{1}{\xi^2} I & \frac{\alpha_t}{\beta_t\xi^2} I  \\
    \frac{\alpha_t}{\beta_t\xi^2} I  & \Big{(} \Sigma_{x_1 x_1} - \Sigma_{x_1 x_1} (\Sigma_{x_1 x_1} +\frac{\beta_t^2}{\alpha_t^2}\xi^2 I)^{-1} \Sigma_{x_1 x_1} \Big{)}^{-1}
    \end{pmatrix}.
\end{align}
where $a = -\frac{\alpha_t}{\beta_t}$. Therefore, $\Xi_{11} \rightarrow \infty$, $\Xi_{01}=\Xi_{10}=\frac{\alpha_t}{\beta_t\xi^2} I \rightarrow \infty$, and $\Xi_{00} = \frac{1}{\xi^2} I\rightarrow \infty $. Thus, we need more detailed calculations to get the result:

\begin{align}
    \nonumber g_t^{\text{sim-inv-diffusion}} =& \Bigg{(}-\dot\beta_t \underbrace{\Xi_{11}^{-1}\Xi_{01}}_{\rightarrow \frac{\alpha_t}{\beta_t} I } \Big{(}\underbrace{\Xi_{11}}_{\rightarrow \infty} + \frac{H^TH}{\sigma_y^2}-\underbrace{\Xi_{10}\Xi_{11}^{-1}\Xi_{01}}_{\rightarrow \infty}\Big{)}^{-1}
    +\dot\alpha_t \Big{(}\underbrace{\Xi_{11}}_{\rightarrow \infty} + \frac{H^TH}{\sigma_y^2}-\underbrace{\Xi_{10}\Xi_{11}^{-1}\Xi_{01}}_{\rightarrow \infty} \Big{)}^{-1}\Bigg{)}\\
    &\left(\frac{H^T}{\sigma_y^2}y - \frac{H^TH}{\sigma_y^2} \hat{x}_1\right).
\end{align}
Obviously we want to find the finite term left in $\Xi_{11} - \Xi_{10}\Xi_{11}^{-1}\Xi_{01}$:
\begin{align}
    \nonumber &\lim_{\xi\rightarrow 0}\Xi_{11} - \Xi_{10}\Xi_{11}^{-1}\Xi_{01} \\
    \nonumber =&\lim_{\xi\rightarrow 0}\Big{(} \Sigma_{x_1 x_1} - \Sigma_{x_1 x_1} (\Sigma_{x_1 x_1} +\frac{\beta_t^2}{\alpha_t^2}\xi^2 I)^{-1} \Sigma_{x_1 x_1} \Big{)}^{-1} - \Xi_{10}\Xi_{11}^{-1}\Xi_{01}\\
    \nonumber =&\lim_{\xi\rightarrow 0}\Big{(} \Sigma_{x_1 x_1} - \Sigma_{x_1 x_1} (\Sigma_{x_1 x_1} +\frac{\beta_t^2}{\alpha_t^2}\xi^2 I)^{-1} \Sigma_{x_1 x_1} \Big{)}^{-1} - \frac{\alpha_t^2}{\beta_t^2\xi^2} \\
    \nonumber =&\lim_{\xi\rightarrow 0}\left( \Sigma_{x_1 x_1}\left(
    \Sigma_{x_1 x_1} +\frac{\beta_t^2}{\alpha_t^2}\xi^2 I\right)^{-1}
    \left(
    (\cancel{\Sigma_{x_1 x_1}} +\frac{\beta_t^2}{\alpha_t^2}\xi^2 I)
    - \cancel{\Sigma_{x_1 x_1}}
    \right)
    \right)^{-1} - \frac{\alpha_t^2}{\beta_t^2\xi^2}\\
    \nonumber =&\lim_{\xi\rightarrow 0}
    \frac{\alpha_t^2}{\beta_t^2\xi^2}
    \left(
    \Sigma_{x_1 x_1}\left(
    \Sigma_{x_1 x_1} +\frac{\beta_t^2}{\alpha_t^2}\xi^2 I\right)^{-1}
    -1
    \right)
    \\
    =&\Sigma_{x_1 x_1}^{-1}. 
\end{align}

Now we have 
\begin{equation}
    g_t^{\text{sim-inv-diffusion}} = \frac{\dot \alpha_t \beta_t - \dot\beta_t\alpha_t}{\beta_t} \left(
    \Sigma_{x_1 x_1}^{-1} + \frac{H^TH}{\sigma_y^2}
    \right)^{-1}
    \left(
    \frac{H^T}{\sigma_y^2}y - \frac{H^TH}{\sigma_y^2} \hat{x}_1
    \right).
\end{equation}

This is essentially the same formulation as in $\Pi$GDM \citep{song_pseudoinverse-guided_2022} and OT-ODE \citep{pokle_training-free_2024}. Next, we will make some trivial conversions to cover the formulations exactly.

In diffusion paths (Assumption \ref{proposition:jacobian_trick}) we proved that $\frac{\partial\hat{x}_t}{\partial x_t} = \frac{\alpha_t}{\beta_t^2}\Sigma_{1|t}$ where $\Sigma_{1|t}$ is just what we denote $\Sigma_{x_1 x_1}$ here. Equivalently, \begin{equation}
    \frac{\partial x_t}{\partial \hat{x}_1} = \frac{\beta_t^2}{\alpha_t}\Sigma_{x_1x_1}^{-1}.
\end{equation}

Thus,
\begin{align}
    \nonumber g_t^{\text{sim-inv-diffusion}} =& \frac{\dot \alpha_t \beta_t - \dot\beta_t\alpha_t}{\beta_t} \left(
    \Sigma_{x_1 x_1}^{-1} + \frac{H^TH}{\sigma_y^2}
    \right)^{-1}
    \left(
    \frac{H^T}{\sigma_y^2}y - \frac{H^TH}{\sigma_y^2} \hat{x}_1
    \right) \\
    \nonumber =&\frac{\dot \alpha_t \beta_t - \dot\beta_t\alpha_t}{\beta_t}
    \Sigma_{x_1 x_1}
    \left(
    {\sigma_y^2}I + \Sigma_{x_1 x_1} H^TH
    \right)^{-1}
    H^T
    \left(
    y - H \hat{x}_1
    \right) \\
    \nonumber =&\frac{\dot \alpha_t \beta_t - \dot\beta_t\alpha_t}{\beta_t}
    \frac{\beta_t^2}{\alpha_t}\frac{\partial \hat{x}_1}{\partial x_t}
    \left(
        {\sigma_y^2}I + \Sigma_{x_1 x_1}H^TH 
        \right)^{-1}
        \left(
        H^T
        y - H^TH \hat{x}_1
    \right) \\
    \nonumber =&\frac{\beta_t(\dot \alpha_t \beta_t - \dot\beta_t\alpha_t)}{\alpha_t}
    \left(
        \frac{\partial \hat{x}_1}{\partial x_t}
        \left(
        {\sigma_y^2}I + \Sigma_{x_1 x_1}H^TH 
        \right)^{-1}
        \left(
        H^T
        y - H^TH \hat{x}_1
        \right)
    \right)\\
    =&\frac{\beta_t(\dot \alpha_t \beta_t - \dot\beta_t\alpha_t)}{\alpha_t}
    \left(
        \left(
        y - H \hat{x}_1
        \right)^TH
        \left(
        {\sigma_y^2}I + \Sigma_{x_1 x_1} H^T H
        \right)^{-1}
        \frac{\partial \hat{x}_1}{\partial x_t}
    \right)^T.
\end{align}

Now we make the same approximation in $\Pi$GDM that $\Sigma_{x_1x_1} = r_t^2 I$. Then
by noticing that 
\begin{align}
    \nonumber \left(
        {\sigma_y^2}I + r_t^2  H H^T
    \right)H
    =&
    H\left(
        {\sigma_y^2}I + r_t^2  H^T H
    \right)
    \\
    H\left(
        {\sigma_y^2}I + r_t^2  H^T H
    \right)^{-1}
    =&
    \left(
        {\sigma_y^2}I + r_t^2  H H^T
    \right)^{-1}   
    H
\end{align}
We exactly cover 
\begin{equation}
    g_t^{\text{sim-inv-$\Pi$GDM}} = \frac{\beta_t(\dot \alpha_t \beta_t - \dot\beta_t\alpha_t)}{\alpha_t}
    \left(
        \left(
        y - H \hat{x}_1
        \right)^T
        \left(
        {\sigma_y^2}I + r_t^2 H^T H
        \right)^{-1}H
        \frac{\partial \hat{x}_1}{\partial x_t}
    \right)^T,
\end{equation}
and the scheduler $\frac{\beta_t(\dot \alpha_t \beta_t - \dot\beta_t\alpha_t)}{\alpha_t}$ in the path $\alpha_t=t,\beta_t=1-t$ becomes $\frac{1-t}{t}$, which exactly covers the schedule in OT-ODE which takes the same path.
In addition, we can also directly compute $\Sigma_{x_1x_1}$ using $\frac{\partial \hat{x}_1}{\partial x_t}$ instead of approximating it with $r_t$. This corresponds to the approach in \citet{boys_tweedie_2024}, which uses the Jacobian to acquire the covariance and then remove the approximation error in computing $\left(
{\sigma_y^2}I + \Sigma_{x_1 x_1} H^T H
\right)^{-1}$.

\begin{remark}
    Starting from the more general assumption of affine path flow matching, we derived the guidance compatible with dependent coupling flow matching, including OT-CFM. The fact that our guidance can exactly cover classical diffusion guidance like $\Pi$GDM and affine Gaussian path flow matching guidance like OT-ODE verifies the validity of our theory.
\end{remark}

\section{Experiment Details}
\label{app:exp}

\subsection{Synthetic Dataset Experiment Details}
\label{app:exp_toy}

The training of flow matching models involves sampling $x_0$ from a source distribution of \{Circle,8 Gaussians,Uniform,Gaussian\} and sampling $x_1$ from the target distribution of \{S-Curve,Moons,8 Guassians\}. The model backbone is an MLP of 4 layers with a hidden dimension of 256. The models are trained 1e5 steps.

\begin{figure}[!htb]
    \centering
    \includegraphics[width=0.8\linewidth]{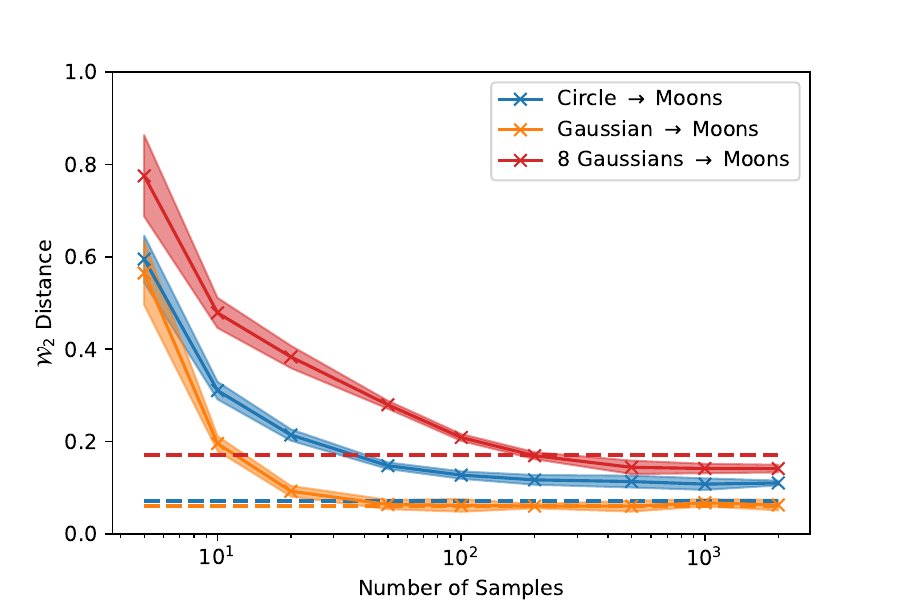}
    \caption{Error scaling with Monte Carlo sample number. In the synthetic dataset, the guidance performance ($\mathcal{W}_2$ distance between the generated distribution and the ground truth energy weighted distribution $p(x_1)e^{-J(x_1)} / Z$) decreases as the number of Monte Carlo samples increases. The dashed lines denote the $\mathcal{W}_2$ distance between the learned unguided distribution and the original ground truth distribution $p(x_1)$. The reason why the guided generation errors (crosses) do not converge to the dashed lines is that they measure the $\mathcal{W}_2$ distance of $p(x_1)$ and $p(x_1)e^{-J(x_1)} / Z$, respectively.}
    \label{fig:asymptotic}
\end{figure}

To evaluate the asymptotic exactness of $g^{\text{MC}}$, we compute the Wasserstain-2 ($\mathcal{W}_2$) distance of the samples generated under guidance, with the ground truth energy-weighted distribution $p(x_1)e^{-J(x_1)} / Z$. Since the source distribution $p(x_1)$ is learned, the flow matching model itself has a small error $w$, which can also be quantified using the $\mathcal{W}_2$ distance. In principle, this error $w$ characterizes the performance upper bound of the guided distribution: the $\mathcal{W}_2$ distance of the guided distribution will, in principle, not be significantly lower than $w$. The result is shown in Figure \ref{fig:asymptotic}, where $w$ is demonstrated using the dashed line.

\subsection{Planning Experiment Details}
\label{app:exp_rl}
\paragraph{Settings.}
The experiment leverages the D4RL (Datasets for Deep Data-Driven Reinforcement Learning) dataset \citep{d4rl}, specifically the Locomotion datasets, which is a common choice to evaluate offline reinforcement learning methods, as well as offline planning methods \citep{janner_planning_2022,dou_diffusion_2023}. The datasets contain non-expert behaviors from which the model is required to learn the optimal policy, such as a mixture of expert and medium-level experts, or the training replay buffer of an RL agent.

To evaluate the performance of different guidance methods, we conduct experiments on offline RL tasks where generative models have been used as planners \citep{janner_planning_2022,chen_flow_2024}. Our setting is based on that of a classical generative planner called Diffuser \citep{janner_planning_2022}, where a generative model generates a state-action pair sequence of multiple future steps, and another critic model that predicts the future reward\footnote{Formally, it predicts the discounted return-to-go.} of the generated plans. The generative model then optimizes its plans using guidance for higher future rewards. Following the formulation in \citet{levine_reinforcement_2018,janner_planning_2022}, the optimization is realized through sampling from $\frac{1}{Z}p(x_1)e^{R(x_1)}$, where $R$ is the critic model. This aligns with the goal of guidance which we discussed in this paper, so we chose this experiment to evaluate different guidance methods.

\paragraph{Baseline Methods.}
The results of the two baselines are collected from the literature \citep{janner_planning_2022}. 
Behavior cloning refers to using a Gaussian distribution to fit the offline behavior distribution. Diffuser refers to using a diffusion model to learn the offline behavior and then guiding the model to generate plans with higher expected future returns.
Note that the Diffuser also adopts training-based guidance, which requires re-training the guidance model when switching to a new objective function. On the contrary, the training-free guidances $g^{\text{cov-A}}$, $g^{\text{cov-G}}$, $g^{\text{sim-MC}}$, and $g^{\text{MC}}$ has zero-shot generalization ability for \citep{zhou_diffusion_2024} the objective function. In the experiment results, we do not highlight the results of baselines as we focus on the comparison between different guidance methods.

\paragraph{Hyperparameters.}
As we mentioned, a generative model is first trained on the offline dataset as a behavior-cloning method but captures the actual action distribution rather than approximating it with a Gaussian. The generative model we consider here is the CFM or mini-batch optimal transport CFM with affine paths $\alpha_t= t, \beta_t = 1 - t$, and whose backbones are an 8-layer Transformer with a hidden dimension of 256. The models are trained with 1e5 steps, a batch size of 32, a learning rate of 2e-4, and the cosine annealing learning rate scheduler. As for the critic model, it is trained with the same backbone model using the last token as the value output and trained 1e4 steps, batch size of 64, and a learning rate of 2e-4. The value discount factor is set to 0.99 for all 3 datasets. We use a planning horizon of 20 steps and the planning stride 1. We exclude tricks such
as using the inverse dynamics model, planning with stride,
and using sample-and-select methods \citep{What_Making}.
During evaluation, the same base model is utilized for different guidance methods to ensure a fair comparison. We report the normalized score \citep{d4rl} where $100$ is the expert RL agent's return.

For different guidance methods, different hyperparameter combinations are tuned. We elaborate on them here.
\begin{itemize}
    \item $g^{\text{cov-A}}$: We tune $\lambda_t$ in \{constant, cosine decay, exponential decay, linear decay\} with a scaler \{0.01, 0.1, 1.0, 10.0\}, where the schedule functions are normalized to [0, 1].
    \item $g^{\text{cov-G}}$: The same hyperparameters are tuned as in $g^{\text{cov-A}}$. 
    \item $g^{\text{MC}}$: We tune the scale of $J$ in \{0.2, 1, 2, 3, 5\}. The Monte Carlo sample number is limited to be smaller than \{128\}. We also include a small number $\epsilon$ to enhance numerical stability. We conduct an ablation study to show the performance is insensitive to $\epsilon$, as shown in Table \ref{tab:app:rl_epsilon}.
    \item $g^{\text{sim-MC}}$: We tune scale before $J$ in \{0.1, 1, 10\} and the assumed standard deviation of $p(x_1|x_t)$ in \{0.1, 0.5, 1, 10\}, and do extra schedule and scale of the estimated guidance with schedule tuned in \{linear decay, constant\} and the scale tuned in \{0.1, 1, 10\}. It is worth noting that if the objective function $J$ is properly normalized, the scale does not require extensive tuning.
    \item $g_\phi$: Training-based methods have many hyperparameters involving model architecture and the training settings. We switch between different model depth and hidden dimensions, and different training losses. The best results for each loss are provided in Table \ref{tab:rl_complete_results_with_std_ot} and \ref{tab:rl_complete_results_with_std_cfm}.
\end{itemize}
Finally, for each pre-trained VF (CFM or OT-CFM), we try adding different guidance VFs (also CFM or OT-CFM), and then the best result is reported.

\paragraph{Estimation of the ground truth distribution of $J$.}
Suppose the unguided model generates $x\sim p(x)$, and $J(x)$ follows the distribution $p_J(J)$. Then, if $x'\sim \frac{1}{Z}p(x)e^{-J(x)}$, then $J'$ follows 
\begin{equation}
    p'(J) = p'(x)\text{det}\left(\frac{\partial x}{\partial J(x)}\right) = \frac{1}{Z}p(x)e^{-J(x)}\text{det}\left(\frac{\partial x}{\partial J(x)}\right).
\end{equation}
Since for the original distribution 
\begin{equation}
    p(J) = p(x) \text{det}\left(\frac{\partial x}{\partial J(x)}\right),
\end{equation}
so
\begin{equation}
    p'(J) = \frac{1}{Z}e^{-J}p(J).
\end{equation}
Therefore, by sampling from the unguided model and then reweighting the distribution of $J$, we can compute the ground truth distribution $p'(J)$ for the $J$ under ideal guidance. Note that in the planning experiment, $J=-R$.

It should be noted that although gradient-based guidances $g^{\text{cov-A}}$ and $g^{\text{cov-G}}$ result in distributions where the estimated return $R$ is higher, it does not necessarily mean that their performance is better: first, the goal of the guidance is the gray line, what we assume here is that the methods produce a distribution close to the gray line is better; second, practically speaking, the high return is predicted by the critic model, but gradient methods may produce plans that the critic has not seen during training, which is called distribution shift, thus cheating the critic. 
On the contrary, the target guided distribution $p(x_1)e^{R(x_1)} / Z$ regularizes the guided distribution on the support of $p(x_1)$, alleviating the problem of distribution shift.

\begin{figure}[!htb]
    \centering
    \begin{minipage}{0.3\textwidth}
        \centering
        \includegraphics[width=\textwidth]{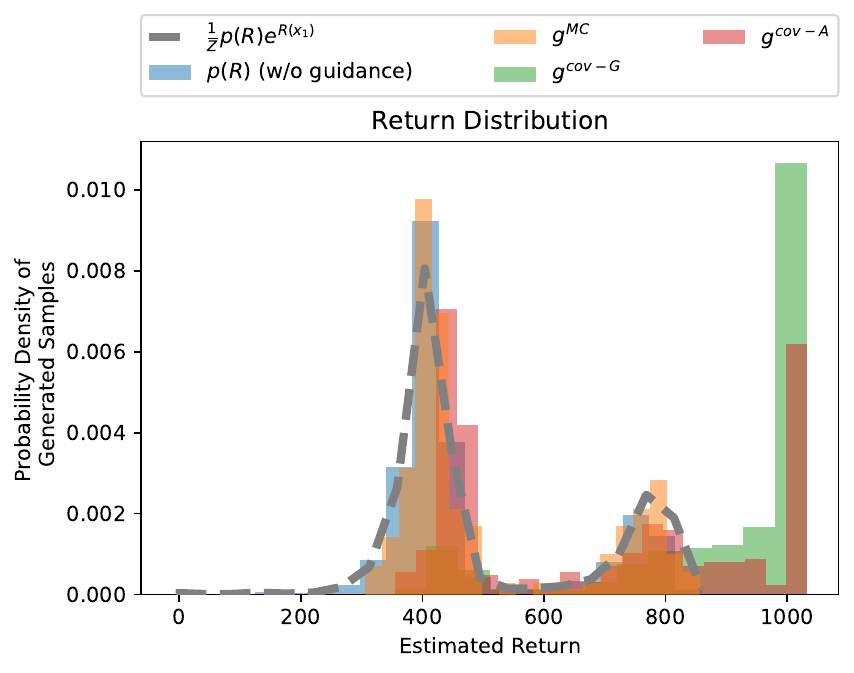}
        \captionof{subfigure}{HalfCheetah, scale $0.001$.}
    \end{minipage}
    \hfill
    \begin{minipage}{0.3\textwidth}
        \centering
        \includegraphics[width=\textwidth]{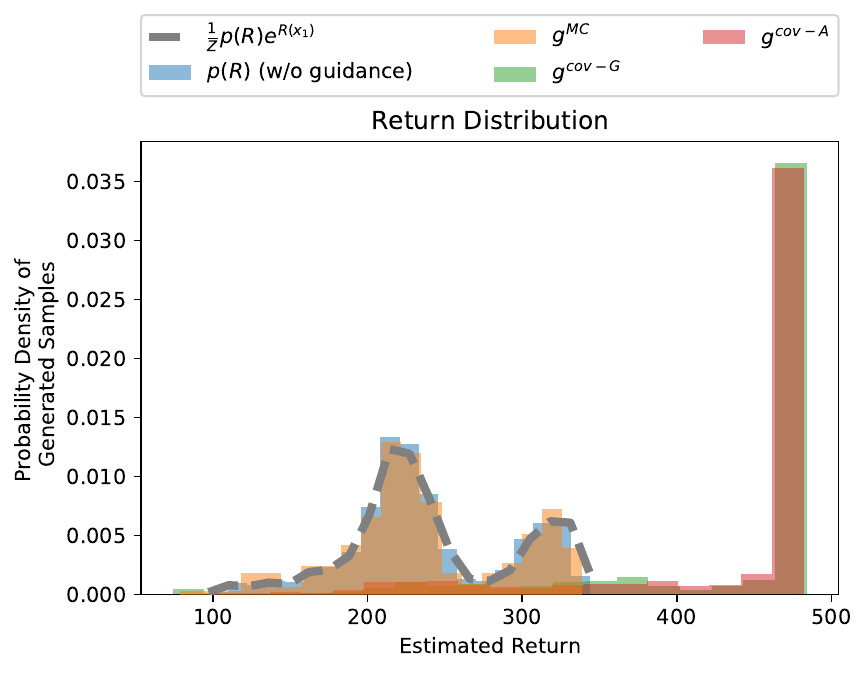}
        \captionof{subfigure}{Walker2d, scale $0.001$.}
    \end{minipage}
    \hfill
    \begin{minipage}{0.3\textwidth}
        \centering
        \includegraphics[width=\textwidth]{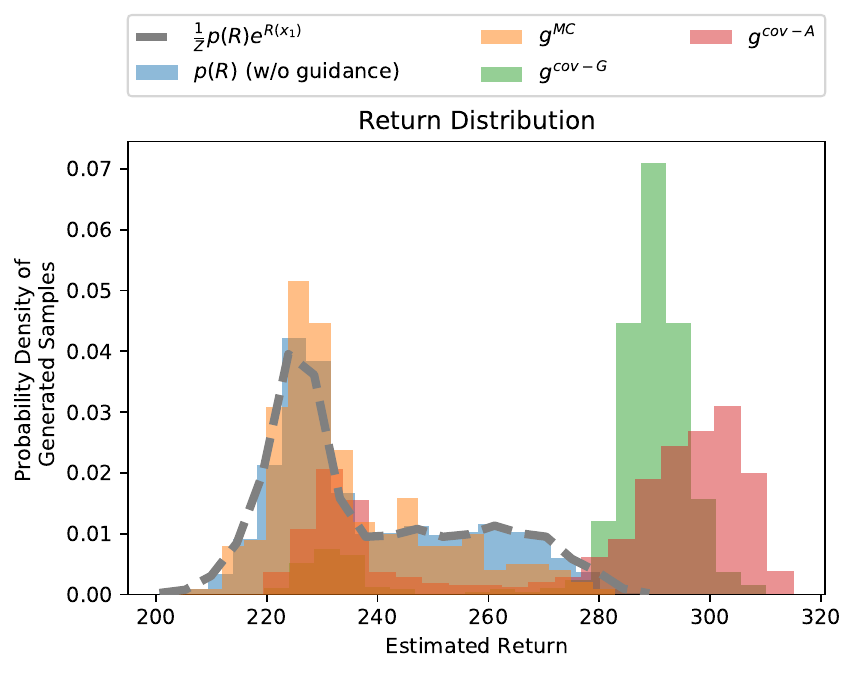}
        \captionof{subfigure}{Hopper, scale $0.001$.}
    \end{minipage}
    
    \vskip\baselineskip %

    \begin{minipage}{0.3\textwidth}
        \centering
        \includegraphics[width=\textwidth]{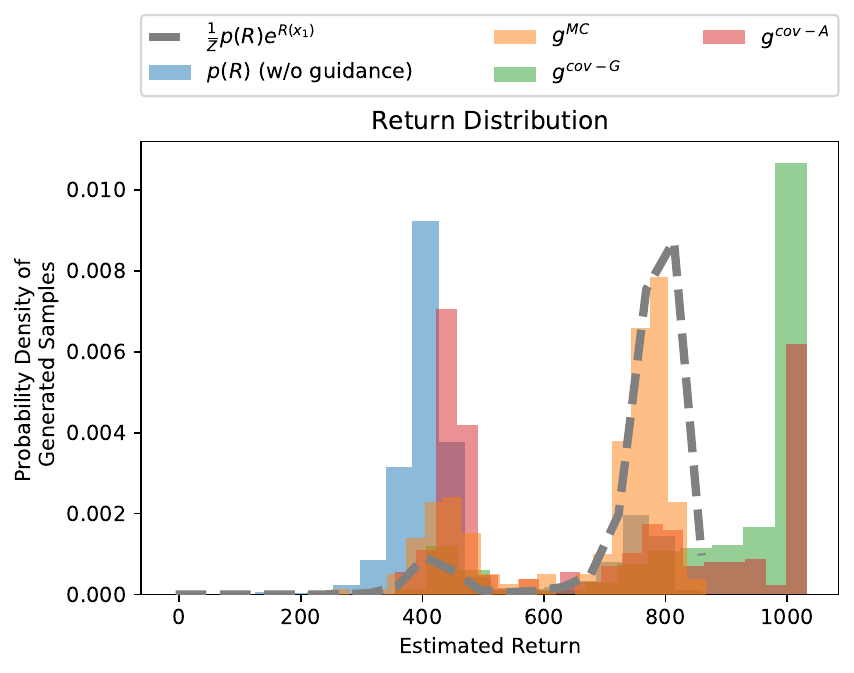}
        \captionof{subfigure}{HalfCheetah, scale $0.01$.}
    \end{minipage}
    \hfill
    \begin{minipage}{0.3\textwidth}
        \centering
        \includegraphics[width=\textwidth]{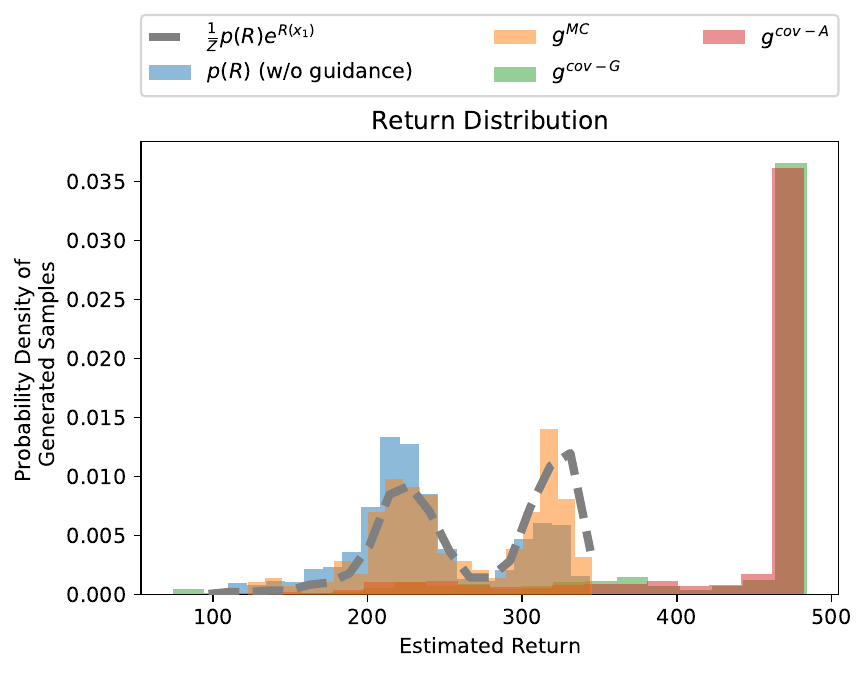}
        \captionof{subfigure}{Walker2d, scale $0.01$.}
    \end{minipage}
    \hfill
    \begin{minipage}{0.3\textwidth}
        \centering
        \includegraphics[width=\textwidth]{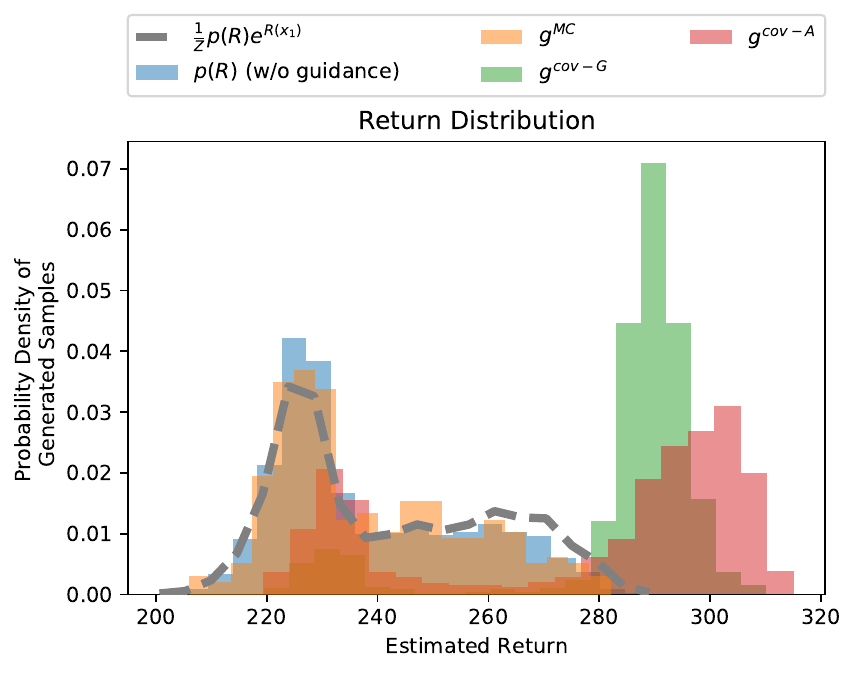}
        \captionof{subfigure}{Hopper, scale $0.01$.}
    \end{minipage}
    
    \vskip\baselineskip

    \begin{minipage}{0.3\textwidth}
        \centering
        \includegraphics[width=\textwidth]{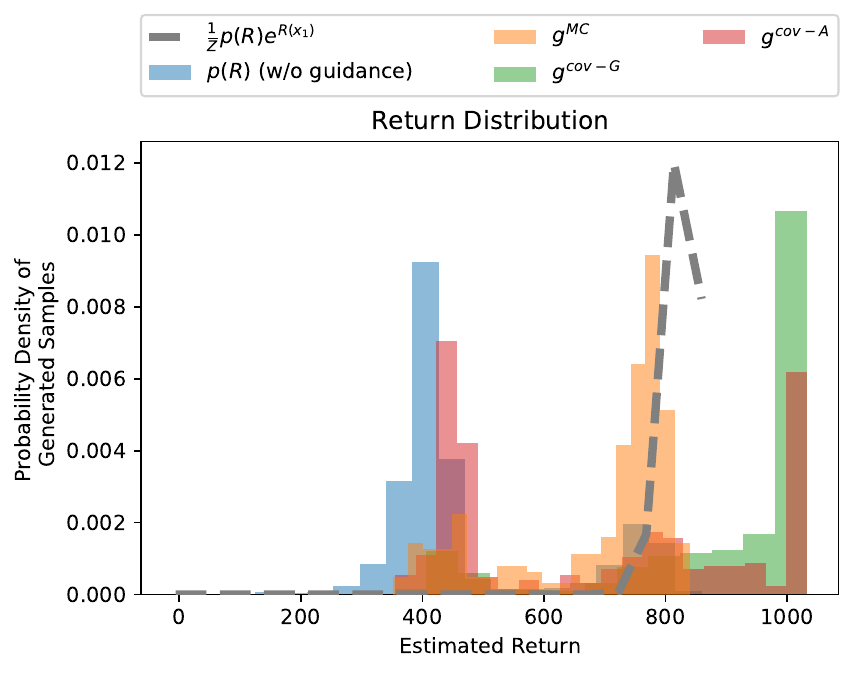}
        \captionof{subfigure}{HalfCheetah, scale $0.05$.}
    \end{minipage}
    \hfill
    \begin{minipage}{0.3\textwidth}
        \centering
        \includegraphics[width=\textwidth]{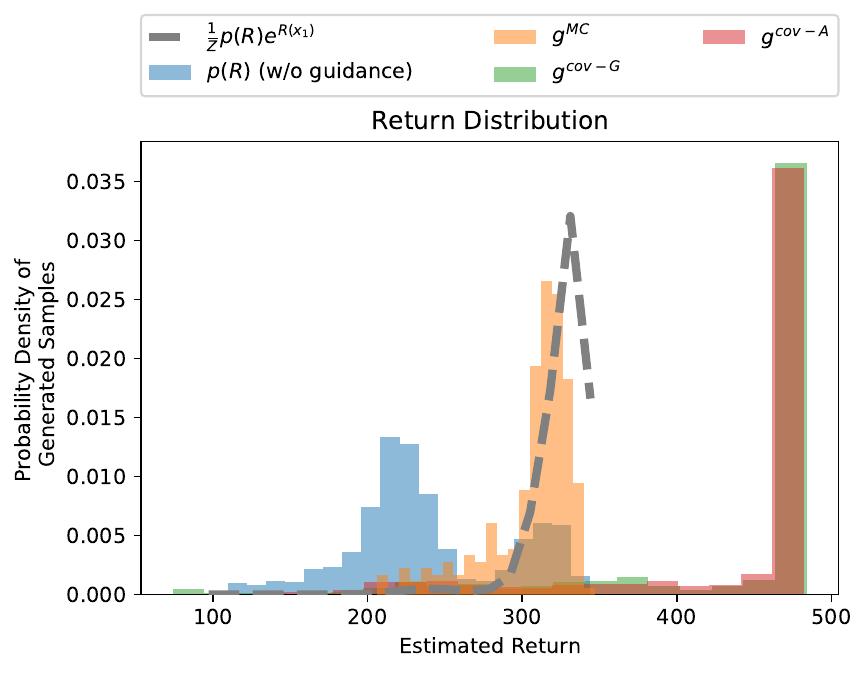}
        \captionof{subfigure}{Walker2d, scale $0.05$.}
    \end{minipage}
    \hfill
    \begin{minipage}{0.3\textwidth}
        \centering
        \includegraphics[width=\textwidth]{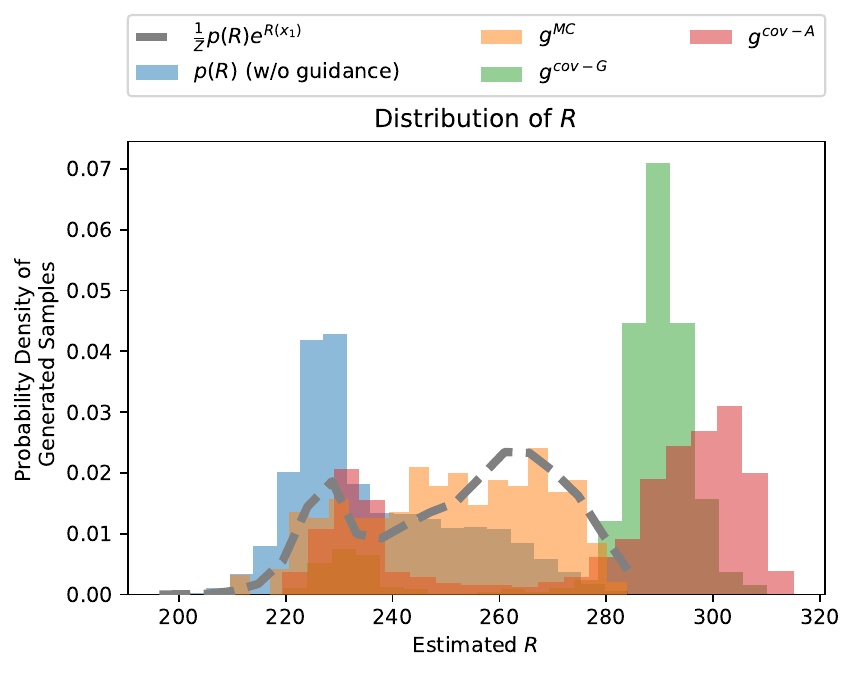}
        \captionof{subfigure}{Hopper, scale $0.05$.}
    \end{minipage}
    
    \caption{The complete results of the distribution of $R$. The distribution of $J$ under $g^{\text{MC}}$ matches the ground truth value (gray dashed line) very well.}
    \label{fig:appendix_rl_complete_distribution_of_R}
\end{figure}

\paragraph{Additional Results on the Distribution of Generated $R$.}
The additional results of the distribution of $R$ in different environments with different guidance scales (the $\alpha$ in $p'(R)=p(R)e^{-\alpha R} / Z$) are shown in Figure \ref{fig:appendix_rl_complete_distribution_of_R}.

\paragraph{Additional Results.}
The complete results, including standard deviations, are provided in Table \ref{tab:rl_complete_results_with_std_ot} and \ref{tab:rl_complete_results_with_std_cfm}.

\begin{table*}[h!]
\caption{Full experiment results on D4RL Locomotion datasets. The base model is mini-batch optimal transport conditional flow matching. Entries with $\ge 95\%$ score than the best per task are highlighted in bold. Baselines are excluded from the ranking.}
\label{tab:rl_complete_results_with_std_ot}
\resizebox{\textwidth}{!}{ %
\centering
\begin{tabular}{llccccccccc}
\toprule
 & & w/o $g$ & $g^{\text{cov-A}}$ & $g^{\text{cov-G}}$ & $g^{\text{sim-MC}}$ & $g^{\text{MC}}$ & $g_\phi$ GM & $g_\phi$ VGM & $g_\phi$ RGM & $g_\phi$ MRGM \\
\midrule
Medium-Expert & HalfCheetah & 61.9 ± 13.3 & 64.8 ± 12.7 & 73.2 ± 9.5 & 78.1 ± 3.2 & \textbf{86.4 ± 0.8} & 59.5 ± 18.4 & 57.7 ± 14.1 & 57.5 ± 13.1 & 70.2 ± 18.1 \\
 & Hopper & 95.2 ± 20.4 & 101.8 ± 22.2 & \textbf{112.3 ± 1.8} & \textbf{112.3 ± 0.8} & \textbf{112.7 ± 0.9} & 85.2 ± 23.3 & 98.1 ± 16.3 & 90.3 ± 24.1 & 89.3 ± 18.7 \\
 & Walker2d & 79.1 ± 35.2 & 97.3 ± 9.4 & \textbf{107.2 ± 1.4} & 101.0 ± 10.2 & \textbf{107.5 ± 1.0} & 87.0 ± 16.7 & 90.5 ± 10.0 & 88.0 ± 17.2 & 91.3 ± 11.7 \\
\midrule
Medium & HalfCheetah & 34.7 ± 9.6 & \textbf{42.2 ± 0.8} & \textbf{42.9 ± 0.9} & \textbf{43.1 ± 1.7} & \textbf{43.1 ± 0.4} & \textbf{42.7 ± 1.4} & \textbf{43.0 ± 1.2} & \textbf{42.7 ± 0.8} & \textbf{43.4 ± 0.9} \\
 & Hopper & 63.3 ± 4.6 & 75.1 ± 14.9 & \textbf{89.8 ± 13.6} & 76.2 ± 13.2 & 79.8 ± 14.8 & 79.7 ± 12.4 & 71.6 ± 9.2 & 77.4 ± 6.6 & 72.5 ± 6.0 \\
 & Walker2d & 72.4 ± 13.3 & \textbf{82.7 ± 5.3} & \textbf{81.3 ± 2.0} & \textbf{83.4 ± 1.9} & \textbf{83.0 ± 3.4} & \textbf{80.6 ± 2.2} & \textbf{80.2 ± 2.0} & 78.4 ± 3.9 & 76.6 ± 6.1 \\
\midrule
Medium-Replay & HalfCheetah & 25.6 ± 13.0 & 31.7 ± 3.4 & 36.1 ± 5.1 & 36.8 ± 1.8 & \textbf{40.0 ± 1.6} & 33.4 ± 2.6 & 35.5 ± 1.8 & 32.9 ± 1.8 & 34.7 ± 3.1 \\
 & Hopper & 40.1 ± 3.7 & 57.7 ± 15.4 & 74.1 ± 5.1 & 60.9 ± 13.1 & \textbf{88.6 ± 11.6} & 54.6 ± 14.8 & 48.1 ± 15.2 & 46.6 ± 11.9 & 55.3 ± 19.4 \\
 & Walker2d & 31.2 ± 6.0 & 62.5 ± 16.8 & 82.5 ± 10.8 & 64.4 ± 9.7 & \textbf{88.1 ± 2.1} & 45.6 ± 17.2 & 37.8 ± 14.5 & 44.3 ± 23.1 & 52.4 ± 20.6 \\
\bottomrule
\end{tabular}
}
\end{table*}

\begin{table*}[h!]
\caption{Full experiment results on D4RL Locomotion datasets. The base model is conditional flow matching. Entries with $\ge 95\%$ score than the best per task are highlighted in bold. Baselines are excluded from the ranking.}
\label{tab:rl_complete_results_with_std_cfm}
\resizebox{\textwidth}{!}{ %
\centering
\begin{tabular}{llcccccccccc}
\toprule
 & & w/o $g$ & $g^{\text{cov-A}}$ & $g^{\text{cov-G}}$ & $g^{\text{sim-MC}}$ & $g^{\text{MC}}$ & $g_\phi$ GM & $g_\phi$ VGM & $g_\phi$ RGM & $g_\phi$ MRGM \\
\midrule
Medium-Expert & HalfCheetah & 46.4 ± 10.1 & 63.4 ± 17.8 & 68.5 ± 6.2 & \textbf{83.5 ± 4.2} & \textbf{87.7 ± 1.8} & 61.2 ± 17.1 & 81.5 ± 7.7 & 64.2 ± 20.7 & 66.4 ± 19.2 \\
 & Hopper & 83.4 ± 19.2 & 93.9 ± 22.5 & \textbf{113.3 ± 1.8} & 88.5 ± 28.0 & \textbf{113.3 ± 0.9} & 91.4 ± 16.1 & 84.2 ± 22.2 & 86.2 ± 18.7 & 86.5 ± 22.4 \\
 & Walker2d & 65.7 ± 12.1 & 100.4 ± 10.7 & \textbf{106.9 ± 0.8} & \textbf{107.0 ± 0.8} & \textbf{107.1 ± 0.4} & 96.2 ± 13.7 & 102.3 ± 9.2 & 100.9 ± 9.4 & 98.2 ± 12.3 \\
\midrule
Medium & HalfCheetah & \textbf{41.8 ± 0.9} & \textbf{43.6 ± 0.6} & \textbf{43.3 ± 1.2} & \textbf{43.8 ± 1.0} & \textbf{43.8 ± 1.0} & \textbf{43.7 ± 0.7} & \textbf{43.7 ± 1.0} & \textbf{42.9 ± 0.7} & \textbf{43.8 ± 1.1} \\
 & Hopper & 73.2 ± 5.4 & 79.1 ± 9.8 & 82.7 ± 11.9 & 82.1 ± 9.2 & \textbf{88.0 ± 11.3} & 81.4 ± 14.0 & \textbf{85.2 ± 15.5} & 74.0 ± 16.3 & 71.6 ± 14.7 \\
 & Walker2d & 72.2 ± 5.9 & \textbf{80.7 ± 1.1} & \textbf{82.5 ± 2.8} & \textbf{81.9 ± 5.4} & \textbf{81.9 ± 1.4} & 67.5 ± 23.7 & 72.9 ± 2.1 & 63.3 ± 21.4 & 55.9 ± 28.6 \\
\midrule
Medium-Replay & HalfCheetah & 22.2 ± 14.9 & 33.4 ± 7.1 & \textbf{39.3 ± 2.0} & 37.9 ± 1.4 & \textbf{40.6 ± 2.0} & 36.9 ± 2.3 & 39.1 ± 1.7 & 37.3 ± 2.1 & 36.8 ± 4.4 \\
 & Hopper & 55.1 ± 17.2 & 63.0 ± 17.5 & 69.3 ± 16.4 & 61.0 ± 22.6 & \textbf{80.9 ± 15.7} & 51.4 ± 22.6 & 63.5 ± 12.7 & 48.6 ± 10.9 & 57.3 ± 24.0 \\
 & Walker2d & 28.3 ± 7.3 & 64.9 ± 20.2 & \textbf{76.6 ± 10.7} & 58.9 ± 18.9 & 70.9 ± 21.7 & 57.5 ± 20.6 & 70.3 ± 4.9 & 53.1 ± 22.0 & 54.6 ± 18.7 \\
\bottomrule
\end{tabular}
}
\end{table*}

\begin{table*}[h!]
\caption{Ablation study of the impact of the epsilon to the performance of $g^{\text{MC}}$. The best results and the second best per task are highlighted in bold and underlined.}
\label{tab:app:rl_epsilon} %
\begin{center}
\resizebox{0.8\textwidth}{!}{ %
\begin{tabular}{cccccccc}
\toprule
 & & $1$ & $1e^{-3}$ & $5e^{-3}$ & $1e^{-2}$ & $5e^{-2}$ \\
\midrule
\textbf{HalfCheetah} & & & & & & \\
\quad Medium & & 42.5 ± 1.6 & 43.1 ± 0.6 & 39.8 ± 8.4 & 41.7 ± 2.9 & \textbf{43.2 ± 1.5} \\
\quad Medium-Expert & & 68.2 ± 15.5 & 66.9 ± 17.1 & \textbf{75.6 ± 12.6} & \textit{74.6 ± 12.1} & 72.7 ± 16.6 \\
\quad Medium-Replay & & 33.5 ± 9.9 & 31.7 ± 12.6 & \textbf{39.7 ± 2.0} & 34.7 ± 10.9 & 37.3 ± 6.3 \\
\midrule
\textbf{Hopper} & & & & & & \\
\quad Medium & & 69.0 ± 10.6 & 67.6 ± 4.9 & \textbf{73.1 ± 9.4} & 72.2 ± 10.0 & \textit{72.6 ± 12.3} \\
\quad Medium-Expert & & 95.4 ± 19.5 & 103.2 ± 19.2 & \textbf{108.0 ± 12.3} & \textit{107.1 ± 14.3} & 101.5 ± 20.8 \\
\quad Medium-Replay & & 53.8 ± 18.7 & 59.1 ± 17.0 & 68.2 ± 18.7 & 68.1 ± 18.1 & \textbf{74.3 ± 18.8} \\
\midrule
\textbf{Walker2d} & & & & & & \\
\quad Medium & & 74.6 ± 6.5 & \textbf{75.1 ± 15.2} & 74.0 ± 10.6 & 71.1 ± 13.8 & \textit{74.4 ± 11.0} \\
\quad Medium-Expert & & 79.0 ± 26.4 & 95.7 ± 17.8 & 94.2 ± 22.0 & 100.9 ± 17.6 & \textbf{103.0 ± 7.9} \\
\quad Medium-Replay & & 48.7 ± 21.5 & 49.5 ± 16.7 & 54.7 ± 20.6 & \textit{57.6 ± 22.3} & \textbf{56.7 ± 18.6} \\
\bottomrule
\end{tabular}
}
\end{center}
\end{table*}

\subsection{Image Experiment Details}\label{app:exp_image}
We pre-trained a CFM and mini-batch optimal transport CFM model with affine path $\alpha_t=t,\beta_t=1-t$ on CelebA-HQ 256$\times$256 dataset. The flow matching model utilizes the backbone of a U-Net following \citep{pokle_training-free_2024}. The pretraining was conducted with a learning rate of 1e-4 and a batch size of 128 for 500 epochs. The run time was roughly 3 days on two H800 GPUs. For the CelebA dataset, we employed a train-validation-test split of 8:1:1. The test data was subsequently used for three downstream tasks: central box inpainting, superresolution by four times, and Gaussian deblurring which is all common benchmarks.

\textbf{Settings for the experiments.} We evaluated the guidance methods using 3,000 images randomly sampled from the test set across the three inverse problems. Specifically, for deblurring, we apply a 61$\times$61 Gaussian kernel with a standard deviation of $\sigma_b = 1.0$. For super-resolution, we perform 4 $\times$ downsampling on the CelebA images. In the case of box-inpainting, we use a centered 40$\times$40 mask. Furthermore, for all three tasks, we add Gaussian noise after the degradation operation with a standard deviation of $\sigma = 0.05$ to the images.

\textbf{Metrics.} In this paper, we use four commonly adopted metrics for image quality assessment: FID (Fréchet Inception Distance), which measures the distance between generated and real image distributions; LPIPS (Learned Perceptual Image Patch Similarity), which evaluates the perceptual similarity between images; and PSNR (Peak Signal-to-Noise Ratio) and SSIM (Structural Similarity Index) to quantify image quality in terms of signal preservation and structural consistency, respectively.

\begin{figure}[!htb]
    \centering
    \includegraphics[width=1.0\linewidth]{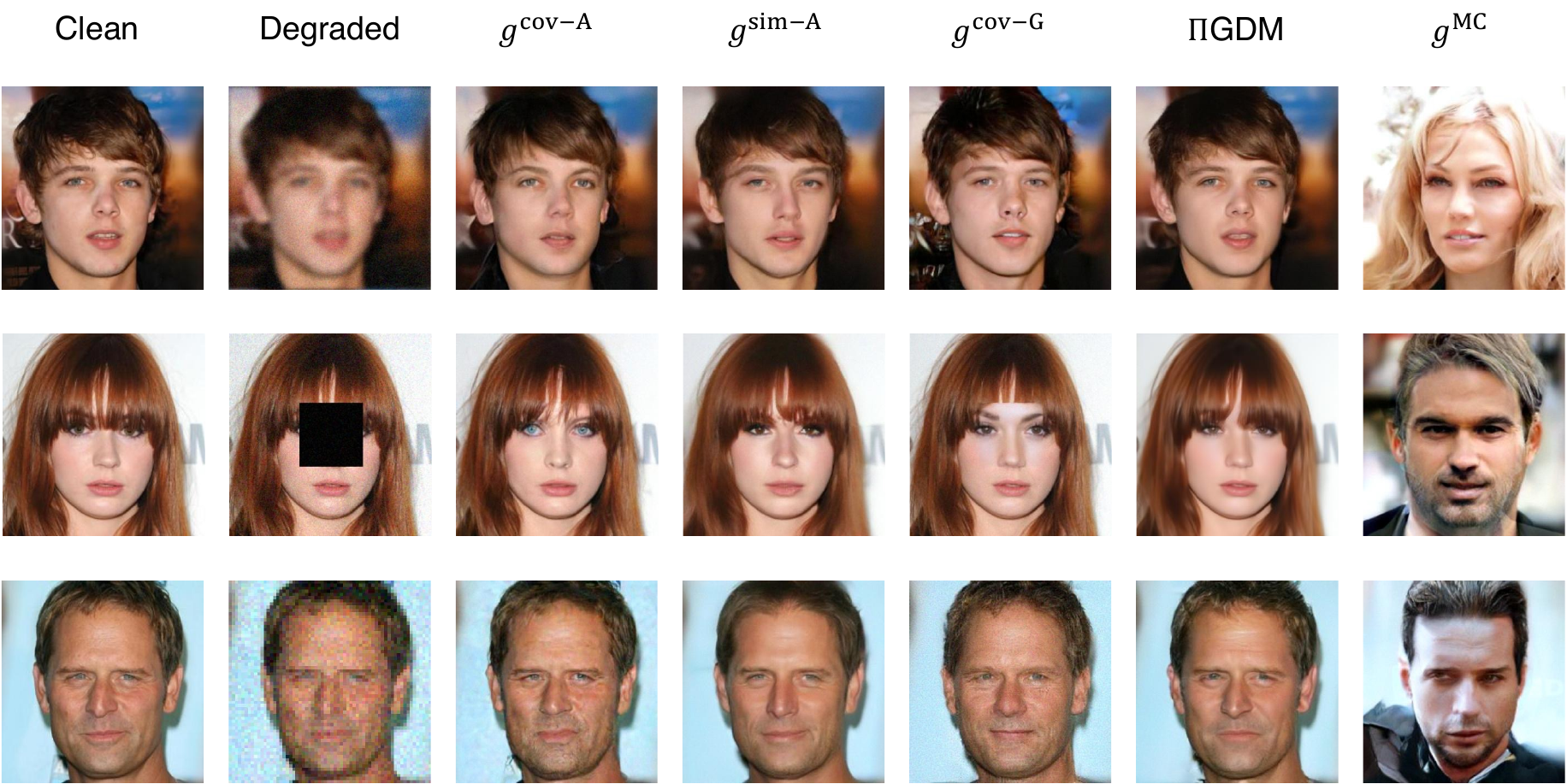}
    \caption{The visualization of the image inverse problems with the base flow matching model of mini-batch optimal transport conditional flow matching (OT-CFM). Three rows show the results of Gaussian deblurring, box-inpainting, and super-resolution from top to bottom.}
    \label{fig:app_image_visualization_ot}
\end{figure}

\begin{figure}[!htb]
    \centering
    \includegraphics[width=1.0\linewidth]{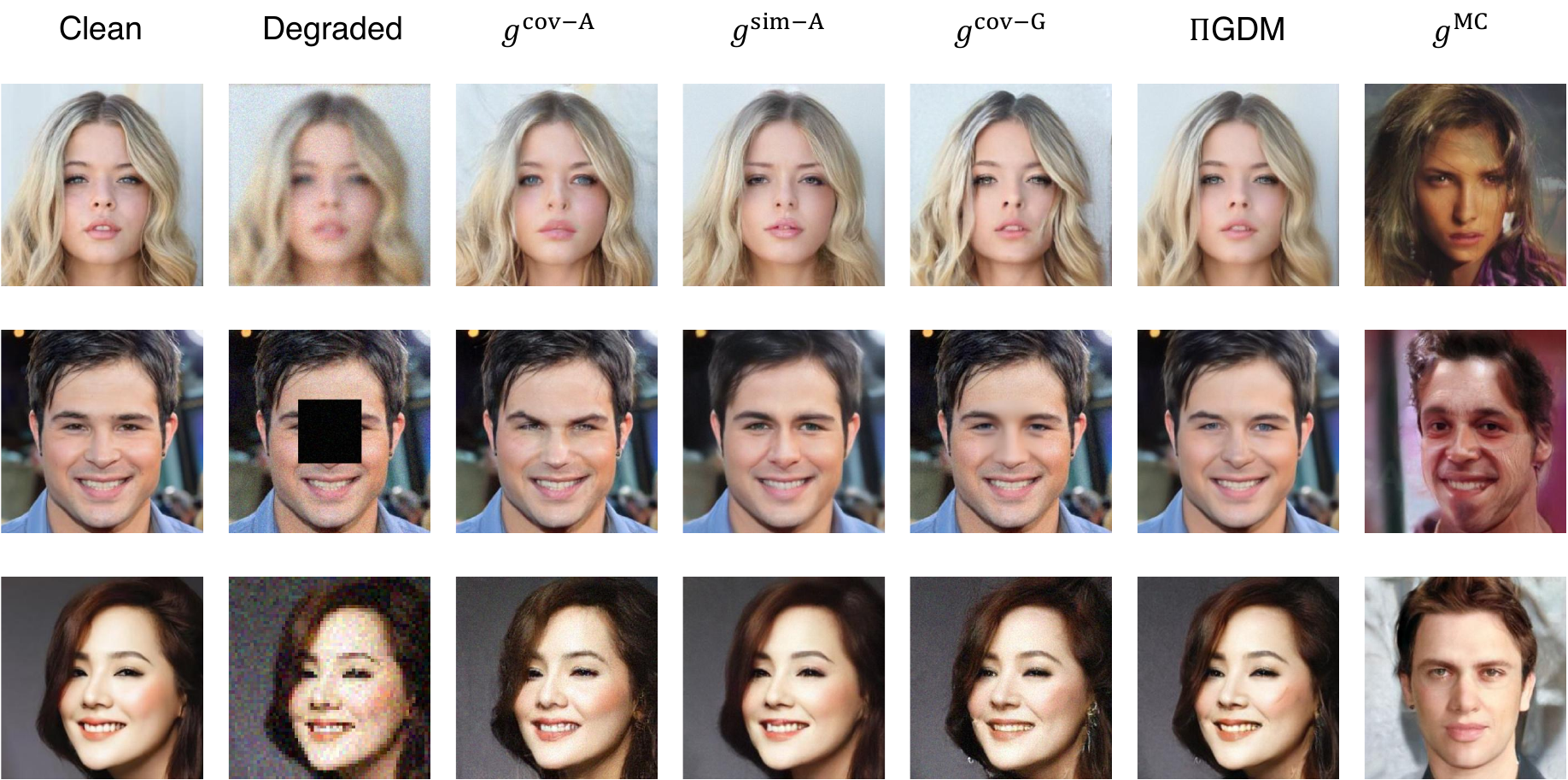}
    \caption{The visualization of the image inverse problems with the base flow matching model of conditional flow matching (CFM). Three rows show the results of Gaussian deblurring, box-inpainting, and super-resolution from top to bottom.}
    \label{fig:app_image_visualization_cfm}
\end{figure}

\textbf{Why is $g^{\text{MC}}$ bad at image inverse problems?}
As can be seen from Figure  \ref{fig:app_image_visualization_ot} and Figure \ref{fig:app_image_visualization_cfm}, the images generated by $g^{\text{MC}}$ do not respect the reference degraded image. This is largely due to the variance of the MC estimation being too high given the limited number of samples. Specifically, to estimate $g_t$, one needs to obtain samples from regions where $e^{-J}$ is significantly higher than average, which corresponds to the images that already look like the degraded image. Sampling such images requires an infeasibly large number of samples. More advanced MC sampling techniques may help address this shortcoming, such as the control variable method \citep{mcbook}. Combining $g^{\text{MC}}$ and methods that are biased but with lower variance, such as $g^{\text{local}}$ or $g^{\text{sim}}$, may also boost the performance.

However, on tasks such as conditional generation, as long as the condition often appears in the dataset, it will be easier to obtain an accurate estimation of $g_t$ using $g^{\text{MC}}_t$. Such scenarios include property-conditioned molecular structure generation, label-conditioned image generation, and decision-making tasks, which are included in our experiments.

\textbf{Visualizations.}
We provide visualizations of the results of the inverse problems in Figure \ref{fig:app_image_visualization_ot} and \ref{fig:app_image_visualization_cfm}.

\end{document}